\documentclass[11pt]{article}

\makeatletter
    \renewcommand\section{\@startsection {section}{1}{\z@}%
                                       {-3.5ex \@plus -1ex \@minus -.2ex}%
                                       {2.3ex \@plus.2ex}%
                                       {\normalfont\fontfamily{phv}\fontsize{16}{19}\bfseries}}
    \renewcommand\subsection{\@startsection{subsection}{2}{\z@}%
                                         {-3.25ex\@plus -1ex \@minus -.2ex}%
                                         {1.5ex \@plus .2ex}%
                                         {\normalfont\fontfamily{phv}\fontsize{14}{17}\bfseries}}
    \renewcommand\subsubsection{\@startsection{subsubsection}{3}{\z@}%
                                        {-3.25ex\@plus -1ex \@minus -.2ex}%
                                         {1.5ex \@plus .2ex}%
                                         {\normalfont\normalsize\fontfamily{phv}\fontsize{14}{17}\selectfont}}
    \makeatother
\usepackage{amsmath}
\usepackage{amsthm} % Added for theorem environments
\usepackage{graphicx}
\usepackage{enumerate}
\usepackage{url}
\usepackage{textcomp}
\usepackage{xcolor}
\usepackage{float}
\usepackage{wrapfig}
\usepackage{multirow}
\usepackage{amsfonts}
\usepackage{mathrsfs}
\usepackage{amssymb,latexsym}
\usepackage{commath}
\usepackage{caption}
\usepackage{subcaption}
\usepackage{booktabs}
\usepackage{mathtools}
\usepackage{todonotes}
\usepackage{soul}
\usepackage{multicol}
\usepackage{amsbsy}
\usepackage{comment}
\usepackage[T1]{fontenc}
\usepackage[cmintegrals]{newtxmath}
\usepackage{bm}
\usepackage{array}
\usepackage{longtable}
\usepackage{setspace}
\usepackage{mdframed} % Added for boxed environments
\usepackage{hyperref}
\usepackage{braket}

\usepackage{algorithm}
\usepackage{algpseudocode}
\usepackage[round,authoryear]{natbib}
\let\cite\citep

\hypersetup{
	linkcolor=cyan,
	filecolor=magenta,
	urlcolor=blue,
}

\theoremstyle{definition} % Move this command to the top

\newmdtheoremenv{theorem}{Theorem}
\newmdtheoremenv{lemma}{Lemma}
\newmdtheoremenv{proposition}{Proposition}
\newmdtheoremenv{definition}{Definition}
\newmdtheoremenv{property}{Property}
\newmdtheoremenv{assumption}{Assumption}

\graphicspath{{Figs/}}

\begin{document}

  \def\spacingset#1{\renewcommand{\baselinestretch}{#1}\small\normalsize} \spacingset{1}
          
  \title{Inverse ECG Reconstruction via a Preconditioned Variational Quantum Linear Solver}

  \author{Xizhuo (Cici) Zhang$^a$, Chao Lu$^b$, and Bing Yao$^a$\footnote{Corresponding author: Bing Yao, byao3@utk.edu}  \\  $^a$Department of Industrial \& Systems Engineering, \\The University of Tennessee, Knoxville, TN,  USA, 37996 \\
  $^b$National Center for Computational Sciences, \\Oak Ridge National Laboratory, Oak Ridge, TN,  USA, 37830 \\ }

  \maketitle

\begin{abstract}
Inverse electrocardiography (ECG) reconstructs heart-surface potentials from body-surface measurements, enabling noninvasive mapping of cardiac electrical activity. However, the reconstruction requires solving dense and ill-conditioned linear systems, whose computational cost grows rapidly with the refined heart mesh. Quantum computing offers new opportunities for solving such systems. This paper presents a hybrid
quantum-classical framework for inverse ECG reconstruction, which, to the best of our knowledge, is the first to apply a quantum linear solver to this problem. We first propose a Laplacian-regularized pre-encoding procedure to reduce the dimension of the system while improving its conditioning. Second, the resulting reduced system is encoded and solved using a coherent variational quantum linear solver (VQLS). Finally, we provide a rigorous theoretical proof establishing that the proposed framework preserves the exact solution of the original reconstruction problem throughout all stages of the computational process. Experimental results on a realistic heart-torso geometry demonstrate that our approach holds significant ability for accurate inverse ECG reconstruction. These findings highlight the potential of quantum solvers for noninvasive cardiac imaging, providing a foundation for further research into quantum-enhanced cardiac assessment and other healthcare applications. 
\footnote{\textbf{Notice:} This manuscript has been authored by UT-Battelle, LLC, under contract DE-AC05-00OR22725 with the US Department of Energy (DOE). The US government retains and the publisher, by accepting the article for publication, acknowledges that the US government retains a nonexclusive, paid-up, irrevocable, worldwide license to publish or reproduce the published form of this manuscript, or allow others to do so, for US government purposes. DOE will provide public access to these results of federally sponsored research in accordance with the DOE Public Access Plan (\href{http://energy.gov/downloads/doe-public-access-plan}{http://energy.gov/downloads/doe-public-access-plan}).}
  \end{abstract}

  \noindent%

  \spacingset{1.5}

  % \section*{Note to Practitioners}
  % This article proposes an effective framework for predictive modeling of dynamic systems that involve complex 3D geometries. The key innovation of our framework lies in incorporating geometric awareness in spatiotemporal predictive modeling and further leveraging adaptive active learning to enable automatic identification of the most informative locations for sequential data collection. While our implementation focuses on cardiac modeling applications, this approach can be valuable in various industrial settings involving spatiotemporal predictive modeling, such as structural health monitoring, environmental sensing, and manufacturing process control. By addressing the challenges of geometric complexity and data efficiency, our framework has the potential to advance predictive modeling of complex dynamic systems and open new avenues for diverse practical applications.

  \section{Introduction} \label{s:intro}

%{\color{blue}(\textbf{P1 – Clinical Background:} Definition of inverse ECG problem: reconstructing heart-surface potentials from body-surface measurements $\rightarrow$ The importance of recovering heart-surface potentials)}

Cardiac electrical activity is routinely monitored by electrocardiography (ECG), which records the electric potentials generated by the heart on the body surface \citep{xie2024automated,wang2023hierarchical}. Body surface potential (BSP) mapping extends the spatial coverage of conventional 12-lead ECG through dense arrays of torso electrodes, providing a more comprehensive picture of the cardiac electrodynamics \citep{yao2021spatiotemporal}. %However, BSPs are spatially blurred due to signal attenuation within the torso. The inverse ECG problem, also known as electrocardiographic imaging, aims to noninvasively reconstruct the heart-surface potentials (HSPs) from BSP measurements \citep{rudy1999noninvasive}. Recovering the HSPs reveals the electrical activity directly on the heart surface, which enables the localization of electrocardiac events \citep{oster1997noninvasive}, the characterization of the location and extent of myocardial infarctions \citep{dawoud2008using, yao2018characterizing}, and the investigation of cardiac repolarization \citep{rudy2009cardiac}, thereby supporting the diagnosis and treatment of heart diseases \citep{yang2023sensing}.
However, volume conduction through the torso attenuates and spatially smooths the cardiac electrical signals, limiting the localization of their sources from BSP measurements alone. Electrocardiographic imaging addresses this limitation by solving an inverse problem to noninvasively reconstruct cardiac electrical activity represented by heart-surface potentials (HSPs) from BSPs \citep{rudy1999noninvasive}. These reconstructions provide spatially resolved estimates of cardiac electrical activity to support the localization of electrophysiological events \citep{oster1997noninvasive}, characterization of the location and extent of myocardial infarction \citep{dawoud2008using, wang2010noninvasive, yao2018characterizing}, and assessment of cardiac repolarization \citep{rudy2009cardiac}, thereby informing the diagnosis and treatment of cardiac disease \citep{yang2023sensing}.

  Under a boundary element method (BEM) discretization, the forward relationship between HSPs and BSPs can be expressed as a linear system, in which a transfer matrix maps the potentials at the heart-surface mesh nodes to those at the body-surface measurement locations \citep{barr1977relating}. However, the inverse problem is severely ill-conditioned because volume conduction through the torso attenuates spatial variations in cardiac electrical activity, resulting in substantial information loss and amplifying the effects of measurement noise and modeling uncertainties \citep{ghosh2009application, yao2016physics}. As such, additional prior information, introduced through regularization and physiological constraints, is essential for obtaining stable and physiologically plausible reconstructions \citep{yao2021spatiotemporal,zhang2026physics}. A commonly adopted approach is Laplacian-based Tikhonov regularization, which formulates the reconstruction as a quadratic optimization problem that balances data fidelity against a penalty on the spatial roughness of the reconstructed HSPs \citep{tikhonov1977solutions, huiskamp1991difference}. Despite its quadratic structure, the computational burden can be significant, particularly when reconstructing cardiac electrical activity over high-resolution meshes and long-duration recordings. %These considerations motivate the development of computationally efficient solution methods that can support high-resolution reconstruction of cardiac electrical activity.

  Classical numerical methods provide established tools for regularized inverse ECG reconstruction \citep{golub2013matrix, trefethen2022numerical}. Direct methods compute solutions through matrix factorizations or decompositions, while iterative methods progressively refine approximate solutions through successive matrix operations \citep{hansen1998rank, saad2003iterative}. %Both approaches have been investigated for inverse ECG modeling. For example, truncated singular value decomposition (TSVD) provides a spectral regularization approach that suppresses contributions associated with small singular values \citep{yao2021spatiotemporal}, whereas the generalized minimal residual (GMRES) method has been applied to reconstruct epicardial potentials from experimental measurements \citep{ramanathan2003noninvasive}. 
  However, computational demands increase as reconstruction problems become larger. Finer heart meshes introduce more unknown potentials \citep{yao2016mesh}, while tracking cardiac activity over time requires repeated reconstructions \citep{oster1997noninvasive, ramanathan2003noninvasive}. These demands are particularly significant for dense formulations: an explicitly stored \(n\times n\) system requires \(O(n^2)\) memory, and each conventional matrix–vector multiplication requires \(O(n^2)\) arithmetic operations \citep{golub2013matrix, trefethen2022numerical}. These scaling demands motivate the investigation of alternative computational approaches that can reduce the cost of solving the underlying linear systems while maintaining the accuracy and stability required for inverse ECG reconstruction.

Quantum computing offers an alternative computational paradigm that exploits quantum-mechanical principles to represent and process information. Quantum computers use qubits that can exist in superpositions of computational basis states, providing a foundation for encoding high-dimensional data into quantum states. %In particular, amplitude encoding represents a normalized vector of dimension $2^q$ using the probability amplitudes of a quantum state defined by $q$ qubits \citep{nielsen2010quantum}. Such a representation provides a compact way to encode high-dimensional vectors and enables quantum operations to transform their encoded information through unitary transformations. 
Building on this capability, quantum linear system algorithms have demonstrated the potential to achieve computational speedups over classical approaches for certain structured linear algebra problems, including matrix inversion, under appropriate assumptions on matrix properties, data access, and solution extraction \citep{ciliberto2018quantum}. These potential computational advantages have motivated growing interest in applying quantum computing to scientific and engineering problems, including hydrological modeling and incompressible flow simulations \citep{golden2022quantum, gopalakrishnan2024towards, ye2024hybrid, bosco2024demonstration}, as well as healthcare applications such as ECG-based arrhythmia detection and heart disease prediction \citep{ozpolat2023performance, pitchal2024heart, burns2025quantum}. These developments motivate the investigation of quantum algorithms as an alternative computational approach to inverse ECG reconstruction. However, translating their theoretical advantages into practical computational improvements requires overcoming several challenges:

%{\color{blue}(\textbf{P5 – Three challenges of applying Quantum Linear Solvers for inverse ECG problem:} (1) The original system is severe ill-conditioning. (2) Large, dense matrix is expensive to handle in quantum circuits. (3) Need recover a complete heart map with meaningful physical values.)}

\textbf{(1) Constructing compact and well-conditioned systems from the ill-conditioned inverse model.}
%Realistic heart--torso discretizations produce large, dense transfer matrices, while ill-conditioning amplifies the effects of noise on reconstruction \citep{barr1977relating, yao2016mesh, yao2016physics}. Quantum linear solvers are sensitive to both system dimension and conditioning \citep{harrow2009quantum, costa2022optimal}. Therefore, effectively applying quantum computing to the inverse ECG problem necessitates reformulating the system to improve its conditioning while preserving the integrity of the original solution.
The dense, ill-conditioned transfer matrix underlying the inverse ECG problem poses substantial challenges for quantum linear algorithms, whose computational complexity depends on the system dimension and the condition number of the coefficient matrix \citep{harrow2009quantum, costa2022optimal}. %Directly applying quantum linear solvers to the original formulation may limit their potential computational advantages. 
Effective application of quantum computing requires converting the original inverse problem into a more compact and better-conditioned linear system while preserving the essential characteristics of the original solution.

% \textbf{(2) Efficient quantum implementation of large, dense systems.}
% The boundary element formulation of inverse ECG produces a dense transfer matrix, and finer heart meshes increase the number of unknown potentials \citep{barr1977relating, yao2016mesh}. Although amplitude encoding represents these unknowns using a compact qubit register, preparing the input state and implementing the matrix can still require substantial circuit resources. For example, in variational quantum algorithms, the matrix is often decomposed into a linear combination of unitaries (LCU), leading to up to $4^q$ Pauli terms for $q$ qubits and a rapidly growing evaluation workload \citep{lu2026distributed}. Pruning terms can reduce this burden, but the resulting approximation error must be controlled to retain reconstruction accuracy. Thus, practical quantum reconstruction requires a suitable problem dimension and matrix encoding that control circuit complexity and evaluation workload while preserving the information needed for recovery.

\textbf{(2) Feasible quantum implementation of the inverse ECG system.}  
%Reducing the system dimension does not by itself ensure an inexpensive quantum implementation. For example, a variational quantum linear solver (VQLS) prepares a trial solution state and uses a classical optimizer to minimize a cost measuring its consistency with the linear system. This requires repeated quantum evaluations of a cost determined by the system matrix. 
A compact linear system can still incur substantial quantum implementation costs.
In Pauli-decomposition-based implementations of the variational quantum linear solver (VQLS), a dense $q$-qubit matrix may be expressed as a linear combination of unitaries (LCU) containing up to $4^q$ Pauli terms. During the optimization process, evaluating the cost of each trial solution on quantum circuits can then require circuit evaluations that grow quadratically with the term count \citep{bravo2019variational, lu2026distributed}. %Therefore, the quantum implementation must control both the gate count for representing the matrix action and the number of circuit evaluations needed during optimization.

%{\color{orange}\textbf{(3) Physically faithful reconstruction from quantum solution states.} Quantum linear solvers return normalized solution states, so the physical scale of the solution is not directly encoded \citep{nielsen2010quantum, aaronson2015read}. Inverse ECG requires spatially resolved potential values over the entire heart surface, including their relative signs and physical magnitude \citep{oster1997noninvasive, rudy2009cardiac}. The {\color{red}reduced solution \textbf{(reduced solution comes from nowhere...)}} must therefore be recovered, rescaled, and mapped to the full spatial domain while satisfying the prescribed physical bounds. This requires an explicit decoding mechanism that connects the compact quantum representation to a complete, physically interpretable HSP map. {\color{red}\textbf{(This challenge may not be necessary)}}}

%{\color{blue}(\textbf{P6 – Proposed Framework and Contributions:} Propose the integrated quantum–classical framework: (1) Classical pre-encoding within mesh-Laplacia regularizer reduces dimension and improves conditioning. (2) Coherent VQLS solves the encoded system. (3) Decoding restores the full heart-surface potential map and its physical scale. (4) Theoretical Guarantees for Solution Preservation and Error Control)}

To address these challenges, we propose an integrated quantum--classical framework comprising Laplacian-regularized pre-encoding, coherent variational quantum linear solver, and physically consistent decoding. The specific contributions are as follows:

\textbf{(1) Laplacian-Regularized Pre-Encoding:} %To render the inverse ECG problem tractable for quantum linear solvers, we construct a classical pre-encoding stage that reduces the dimension and improves the conditioning of the inverse ECG system. A coarse estimate regularized by the mesh Laplacian guides the construction of a compact spatial subspace that resolves the electrically active region. The compressed system is then expressed in Krylov coordinates and conditioned by block-Jacobi preconditioning and a spectral tail correction. These steps produce a system that fits a small quantum register and whose condition number can be brought close to one under the stated conditions.
We develop a classical pre-encoding procedure to reduce the system dimensionality and improve numerical conditioning prior to quantum computation. First, a Laplacian-regularized coarse reconstruction is used to identify the spatial region of cardiac electrical activation, which guides the construction of a compact reconstruction subspace. The inverse problem is then projected onto an orthonormal Krylov basis to obtain a reduced-order system. We further apply block-Jacobi preconditioning followed by spectral correction targeting the extreme singular values. This pre-encoding procedure yields a well-conditioned reduced system with a dimension tailored to the available quantum register, achieving a condition number close to one.

\textbf{(2) Coherent VQLS for the Inverse ECG Problem:} To the best of our knowledge, this work is the first to apply a quantum linear solver to the inverse ECG problem. We propose to solve the compact system produced by the pre-encoding stage using a coherent VQLS \citep{bravo2019variational, mari2019coherent}. Specifically, the trial solution state is prepared using a M\"{o}tt\"{o}nen ansatz, and the dense system matrix is incorporated into a quantum circuit through a fast approximate block-encoding (FABLE) scheme. The global cost function is then evaluated using two measurement probabilities obtained from a coherent circuit construction. This implementation provides a unified quantum-circuit framework for representing the linear system, preparing the variational solution, and evaluating the optimization objective. Following optimization and scale recovery, the decoding stage further maps the estimated reduced coefficients to potentials across the full heart surface.
%Specifically, a M\"{o}tt\"{o}nen ansatz prepares the trial solution state, a FABLE block encoding embeds the dense system matrix into a unitary operator, and the global cost is evaluated from two measurement probabilities obtained using a single circuit design. 

%{\color{orange}\textbf{(3) Physically Consistent Decoding for Complete Heart-Surface Reconstruction:} To recover a complete and physically meaningful heart map, we recover the physical scale of the reduced solution and map its coefficients to the full heart surface through an isometric decoder constructed during pre-encoding. Projection onto the admissible potential range then enforces the prescribed physical bounds. This procedure converts the normalized quantum representation into a complete HSP map with the physical values required for cardiac interpretation.{\color{red} \textbf{(This may not be necessary).}}}

\textbf{(3) Theoretical Guarantees for Solution Preservation and Error Control:} 
%To provide a rigorous mathematical foundation, each stage of the framework is supported by theoretical guarantees. At the preparation stage, a forward-error certificate bounds the error of the mesh-Laplacian coarse estimate under the stated noise assumptions, supporting selection of the regularization weight used to construct one reduced space. Once the spatial subspace is fixed, the conditioning transformations preserve the system solution in exact arithmetic and yield a computable condition number. Within this processed system, the M\"{o}tt\"{o}nen ansatz can represent the normalized solution state. Errors in the recovered coefficients then pass through the fixed isometric decoder without changing their Euclidean norm, while the subsequent physical projection is non-expansive. Together, these results guarantee that the solution is preserved through the conditioning steps and that the error of the quantum solver is not amplified in the final HSP map.
We establish theoretical guarantees to support the mathematical validity and numerical reliability of the proposed framework. Specifically, we characterize the reconstruction error under prescribed noise assumptions, establish conditions under which the solution is preserved by the conditioning transformations, and derive a computable characterization of the condition number of the resulting system. In addition, we demonstrate the representational capability of the adopted M\"{o}tt\"{o}nen ansatz for encoding the solution state and establish that errors in the recovered quantum state are not amplified by the subsequent isometric decoding.

% {\color{orange}(short version)
% \textbf{(4) Theoretical Guarantees for Solution Preservation and Error Control:} 
% To provide rigorous guarantees for each stage, a forward-error certificate bounds coarse reconstruction error using observable quantities, guiding the regularization weight. Conditioning transformations preserve the compressed-system solution exactly and provide a computable condition number. The M\"{o}tt\"{o}nen ansatz spans all real solution states, the decoder is an isometry, and the physical prior is non-expansive, so decoding does not amplify quantum-solver residuals. Together, these properties control error propagation through spatial decoding and physical projection.
% }

%{\color{blue}(\textbf{P7 – Validation:} Reconstruction accuracy. The future potential ability of quantum computing for healthcare applications within the evolution of quantum hardwares.)}

We evaluate the proposed framework on reconstructing HSPs from BSPs 
across a realistic heart-torso geometry. Experimental results confirm that integrating classical pre-encoding with the quantum linear solver enables accurate inverse ECG reconstruction. 
% Although a quantum speedup over classical methods is not yet attainable, our findings establish the practicality of quantum linear solvers for this application.
As quantum hardware continues to advance, the framework provides a promising foundation for future quantum-enabled cardiac imaging and healthcare innovations.

% The remainder of this paper is organized as follows. Section~\ref{s:re-review} reviews related work on inverse ECG modeling, quantum linear system algorithms, and quantum computing for cardiac applications. Section~\ref{s:method} presents the proposed methodology. Section~\ref{s:results} describes the experimental design and discusses the results. The last section concludes the paper, and the theoretical properties and implementation details of the framework are provided in the Supplementary Material.

  \section{Research Background} \label{s:re-review}

\subsection{Inverse ECG Problem}

The inverse ECG formulation treats the torso as a passive volume conductor bounded by the heart and torso surfaces. Combining the volume conductor model with the BEM discretization yields a transfer matrix $\mathbf{R} \in \mathbb{R}^{m \times n}$, which relates the HSPs at $n$ heart-surface nodes to the BSPs at $m$ torso electrodes \citep{barr1977relating, yao2021spatiotemporal}. Accordingly, the measurement model at time $t$ is expressed as $\tilde{\boldsymbol{y}}(t) = \mathbf{R}\boldsymbol{x}(t) + \boldsymbol{\xi}(t)$, where $\boldsymbol{x}(t) \in \mathbb{R}^{n}$ represents the unknown HSPs, $\tilde{\boldsymbol{y}}(t) \in \mathbb{R}^{m}$ denotes the measured BSPs, and $\boldsymbol{\xi}(t) \in \mathbb{R}^{m}$ is measurement noise. As such, the inverse ECG problem involves recovering the HSP vector by solving a linear inverse problem. However, the transfer matrix $\mathbf{R}$ is generally dense because each body-surface measurement depends on the potentials across the entire heart surface. More importantly, it is severely ill-conditioned due to the smoothing effect of volume conduction, which attenuates spatial variations in cardiac electrical activity. 

%{\color{blue}(\textbf{P2 - Regularization or physical constraints stabilize reconstruction} (introduce mesh-Laplacian spatial regularizer))}

A robust solution of the inverse ECG problem requires regularization, which incorporates prior knowledge of the HSPs to stabilize the reconstruction \citep{yao2021spatiotemporal}. A representative spatially regularized formulation is
$
\min_{\boldsymbol{x}\in\mathcal{C}}
    \left\{\norm{\mathbf{R}\boldsymbol{x}-\tilde{\boldsymbol{y}}}^2
    +\lambda^2\norm{\mathbf{D}\boldsymbol{x}}^2\right\},
    \label{eq:bg_regularized}
$
where regularization parameter $\lambda>0$ controls the penalty strength, $\mathbf{D}$ specifies the spatial prior, and $\mathcal{C}$ denotes an admissible set, such as prescribed potential bounds. Choosing $\mathbf{D}=\mathbf{I}_n$ yields zero-order Tikhonov regularization, whereas a discrete mesh Laplacian penalizes spatial variation according to the connectivity and geometry of the triangulated heart surface \citep{hansen1998rank, huiskamp1991difference, yao2016physics,zhang2026geometry}. %The latter favors spatially smooth reconstructions on an irregular anatomical domain.

Alternative regularization strategies can incorporate additional structural and temporal information into the reconstruction. For example, $L_1$-norm regularization can promote sparse solutions and preserve sharper spatial features, while temporal regularization couples reconstructions across successive measurements to exploit the temporal continuity of cardiac electrical activity \citep{ghosh2009application, brooks1999inverse, yao2021spatiotemporal}. Beyond these conventional regularization approaches, model-based methods incorporate cardiac electrophysiology directly into the inverse problem by imposing physiological constraints derived from mathematical models of cardiac electrical activity, such as stochastic state-space representations \citep{wang2010physiological}. More recently, physics-constrained deep learning approaches have incorporated physical consistency into data-driven inverse ECG reconstruction \citep{xie2023physics, xie2022physics,zhu2025physics}. 

%Alternative penalties, including the $L_1$ norm, can retain sharper spatial features, and temporal constraints can couple reconstructions across successive measurements \citep{ghosh2009application, brooks1999inverse, yao2021spatiotemporal}. More advanced regularization techniques further incorporate the physics of cardiac electrodynamics into the reconstruction, for example, by constraining the solution with a physiological model of cardiac electrophysiology formulated as a stochastic state-space system \citep{wang2010physiological}. Additionally, physics-constrained deep learning methods combine the misfit of the BSPs with the residual of the physics-based cardiac electrodynamics model in the training loss to improve the robustness of inverse ECG reconstruction \citep{xie2023physics, zhu2025physics}. %Although such techniques capture richer prior knowledge, they lead to nonlinear reconstruction problems. By contrast, the regularized formulations above can be solved more efficiently and reliably, because they rely only on the solution of linear systems. The accuracy and efficiency of inverse ECG reconstruction therefore depend on how these linear systems are solved.

% These choices define the reconstruction model. Unconstrained quadratic formulations lead to linear least-squares problems, while nonquadratic penalties or additional constraints may require repeated linear solves within an optimization procedure. This distinction separates the stabilization of the inverse problem from the numerical algorithms used to solve it.

\subsection{Classical Linear System Solvers}

%{\color{blue}(\textbf{P1 – Direct Solution Approaches:} Direct factorization and decomposition methods used for the relevant linear or least-squares formulations $\rightarrow$ The effectiveness and numerical properties $\rightarrow$ Computational and memory requirements, including opportunities to reuse a factorization across repeated reconstructions.)}

 %{\color{red}\textbf{(Please make better connection with the previous subsection...->Revised)}}{\color{orange}With a quadratic penalty, the regularized inverse ECG problem at each time instance is a linear least-squares problem, equivalent to a symmetric positive definite system of normal equations.}

%Although advanced physics-constrained approaches capture more informative prior knowledge, they typically yield nonlinear reconstruction problems. In contrast, regularized formulations are generally more computationally efficient and numerically robust, because they often reduce to the solution of linear systems. As such, the accuracy and computational performance of inverse ECG reconstruction depend critically on the methods used to solve these linear systems.

Regularization methods incorporate prior information into the inverse ECG reconstruction problem through appropriately designed objective functions or constraints, resulting in mathematical formulations with different computational structures. Among these approaches, quadratic regularization methods are particularly amenable to numerical solution because their formulations can be expressed as linear least-squares problems. Classical solvers for such systems fall into direct and iterative methods. Direct methods compute solutions through matrix factorizations \citep{golub2013matrix, trefethen2022numerical}. Specifically, the Cholesky factorization can be applied to the normal equations, whereas QR factorization and singular value decomposition (SVD) can be applied directly to the least-squares formulation \citep{golub2013matrix,hansen1998rank}. Despite their numerical accuracy and predictable computational behavior, direct methods can become computationally demanding for large-scale problems \citep{golub2013matrix}. %For a dense $m \times n$ transfer matrix, standard factorizations generally require $\mathcal{O}(mn\min\{m,n\})$ operations and $\mathcal{O}(mn)$ storage, making their computational and memory requirements potentially prohibitive as the problem size increases \citep{golub2013matrix}.

Iterative methods compute approximate solutions through successive updates based primarily on matrix--vector products \citep{saad2003iterative}. The choice of solver depends on the structure of the problem. For example, conjugate gradient is suited to symmetric positive definite systems, LSQR addresses least-squares problems without explicitly forming the normal equations, and generalized minimal residual (GMRES) accommodates nonsymmetric systems \citep{hestenes1952methods, paige1982lsqr, saad1986gmres}. In inverse ECG applications, \citet{ramanathan2003noninvasive} used GMRES to reconstruct HSPs, and \citet{yao2021spatiotemporal} applied block-Jacobi iteration to the linear subproblems arising from their $L_1$-norm spatiotemporal model. 
Although these iterative methods can avoid full matrix factorizations, repeated solves across time points or within spatiotemporal optimization can remain computationally demanding as the heart mesh is refined.

\subsection{Quantum Linear System Algorithms (QLSAs)}

Quantum linear system algorithms (QLSAs) provide an alternative computational framework for solving the linear systems arising in inverse ECG reconstruction. The Harrow--Hassidim--Lloyd (HHL) algorithm was the first QLSA, which prepares a quantum state encoding the normalized solution through quantum eigenvalue inversion \citep{harrow2009quantum}. Specifically, the algorithm first encodes the normalized input vector as a quantum state. For a Hermitian system matrix, quantum phase estimation is then applied to the matrix-generated unitary evolution to encode eigenvalue estimates into an ancillary register, thereby associating each eigenvector component with its corresponding eigenvalue. Controlled rotations are then performed on an additional ancilla qubit to apply reciprocal-eigenvalue weighting to the corresponding eigenvector amplitudes. By reversing the phase estimation procedure and postselecting the ancilla qubit in the desired state, the algorithm produces a quantum state proportional to the inverse-matrix transformation of the input state, thereby approximating the normalized solution. %For sparse and efficiently accessible system matrices, HHL achieves a complexity that scales logarithmically with the system dimension, quadratically with the condition number, and inversely with the target solution-state error \citep{harrow2009quantum}.

%Later algorithms improved these dependences. \citet{childs2017quantum} replaced QPE with Fourier and Chebyshev expansions of the inverse, implemented as linear combinations of unitaries (LCUs), improving the precision dependence exponentially. Adiabatic-inspired algorithms \citep{subasi2019quantum, an2022quantum} and eigenstate filtering \citep{lin2020optimal} reduced the condition-number dependence to the asymptotically optimal linear scaling \citep{costa2022optimal}. The quantum singular value transformation (QSVT) unifies these developments \citep{gilyen2019quantum}. Specifically, given a block encoding of the system matrix, it applies a polynomial transformation to the singular values, realizing inversion through a polynomial approximation of the reciprocal function. However, these algorithms require deep coherent circuits, such as high-precision QPE, long Hamiltonian simulations, or high-degree polynomial transformations, which demand fault-tolerant hardware beyond current noisy intermediate-scale quantum (NISQ) devices \citep{preskill2018quantum}. Moreover, state preparation and block encoding become costly for unstructured vectors and dense matrices \citep{mottonen2004transformation, shende2006synthesis, camps2022fable}, and even the optimal algorithm scales linearly with the condition number \citep{costa2022optimal}.

Subsequent developments have further improved the computational complexity of QLSAs. \citet{childs2017quantum} developed an approach based on Fourier and Chebyshev expansions of the inverse function, implemented through LCUs, achieving an exponential improvement in precision dependence over HHL. Other approaches, including adiabatic-inspired algorithms \citep{subasi2019quantum, an2022quantum} and eigenstate filtering \citep{lin2020optimal}, have improved the dependence on the condition number from quadratic to linear scaling, which is asymptotically optimal under the relevant algorithmic assumptions \citep{costa2022optimal}. Quantum singular value transformation (QSVT) further provides a general framework for implementing matrix functions through polynomial transformations of singular values \citep{gilyen2019quantum}. %Given a block encoding that embeds a scaled system matrix within a larger unitary operator, QSVT implements approximate inversion using a polynomial approximation to the reciprocal function.
Despite the theoretical advances, implementing such algorithms can require deep coherent quantum circuits, involving high-precision quantum phase estimation, extensive Hamiltonian simulation, or high-degree polynomial transformations. These requirements pose substantial challenges for implementation on current noisy intermediate-scale quantum (NISQ) hardware \citep{preskill2018quantum}. Furthermore, efficient quantum implementations depend critically on the cost of state preparation and matrix access. Preparing arbitrary input states and constructing block encodings of dense matrices can introduce substantial circuit and resource overheads \citep{mottonen2004transformation, shende2006synthesis, camps2022fable}. Moreover, even theoretically optimal QLSAs retain a linear dependence on the condition number, which can remain a significant computational bottleneck for ill-conditioned inverse problems \citep{costa2022optimal}.

To address the hardware limitations of current NISQ devices, \citet{bravo2019variational} introduced the VQLS, a hybrid quantum--classical algorithm that combines parameterized quantum circuits with classical optimization \citep{cerezo2021variational}. Specifically, a parameterized quantum circuit prepares a trial quantum state, and quantum measurements are used to estimate a cost function that quantifies the extent to which the trial state satisfies the target linear system. A classical optimizer then iteratively updates the circuit parameters to minimize this cost and identify an approximate solution state. In the original VQLS formulation, the system matrix is represented as an LCU, and the cost function is evaluated using Hadamard tests to estimate the relevant pairwise interference terms \citep{bravo2019variational}.
VQLS has been investigated in solving the advection-diffusion equation \citep{demirdjian2022variational}, preconditioned hydrological systems \citep{golden2022quantum}, the incompressible Navier-Stokes equations \citep{gopalakrishnan2024towards}, and computational fluid dynamics \citep{ye2024hybrid, bosco2024demonstration}. These studies demonstrate the applicability of variational quantum approaches to structured linear systems. Nevertheless, applying VQLS to inverse ECG reconstruction presents distinct challenges, involving a dense and severely ill-conditioned transfer matrix that complicates both quantum data encoding and the optimization of the variational solution.

In practice, encoding dense system matrices presents a significant challenge for the original LCU-based VQLS formulation, in which the cost function is evaluated using Hadamard tests to estimate contributions from pairs of unitary terms \citep{bravo2019variational}. The number of such pairwise contributions scales quadratically with the number of terms in the LCU decomposition. This scaling motivates alternative strategies for cost evaluation. Coherent evaluation addresses this bottleneck by incorporating the block-encoded system matrix directly into a single quantum circuit, rather than evaluating the individual LCU contributions separately. Specifically, the coherent cost-evaluation circuit combines trial-state preparation, application of the block-encoded matrix, and inverse preparation of the right-hand-side state. The global cost can then be estimated from measurement probabilities, without evaluating every pairwise contribution in the LCU expansion \citep{mari2019coherent}. While this approach avoids the quadratic growth in the number of separately evaluated Hadamard-test contributions, its overall efficiency remains dependent on the resources required to implement the block encoding. Thus, an efficient block-encoding strategy is essential for practical coherent evaluation \citep{camps2024explicit, sunderhauf2024block}.

To facilitate block encoding of dense matrices, \citet{camps2022fable} introduced fast approximate block encodings (FABLE), which construct a quantum circuit using uniformly controlled rotations whose angles are determined by the matrix entries, together with Hadamard layers and a register swap. %For a real \(2^q \times 2^q\) matrix with entries bounded in magnitude by one, the FABLE construction uses \(q+1\) ancilla qubits and requires \(4^q\) single-qubit rotations and \(4^q\) CNOT gates arranged according to a Gray-code sequence, with a subnormalization factor of \(2^q\). Unlike an LCU-based representation, FABLE provides a direct approach to block encoding dense matrices. 
Its implementation is also available in quantum software libraries, including PennyLane \citep{bergholm2018pennylane}. Moreover, the circuits can be compressed by discarding small rotation angles with a controlled approximation error, which substantially reduces the gate counts for structured matrices such as Heisenberg and Hubbard Hamiltonians and discretized Laplacian operators \citep{camps2022fable}. As such, coherent cost evaluation and FABLE offer an effective framework for incorporating dense system matrices, including those arising in inverse ECG reconstruction, into a VQLS-based approach.

  \section{Research Methodology}
\label{s:method}

\begin{figure}[!ht]
  \centering
 \includegraphics[width=1\linewidth]{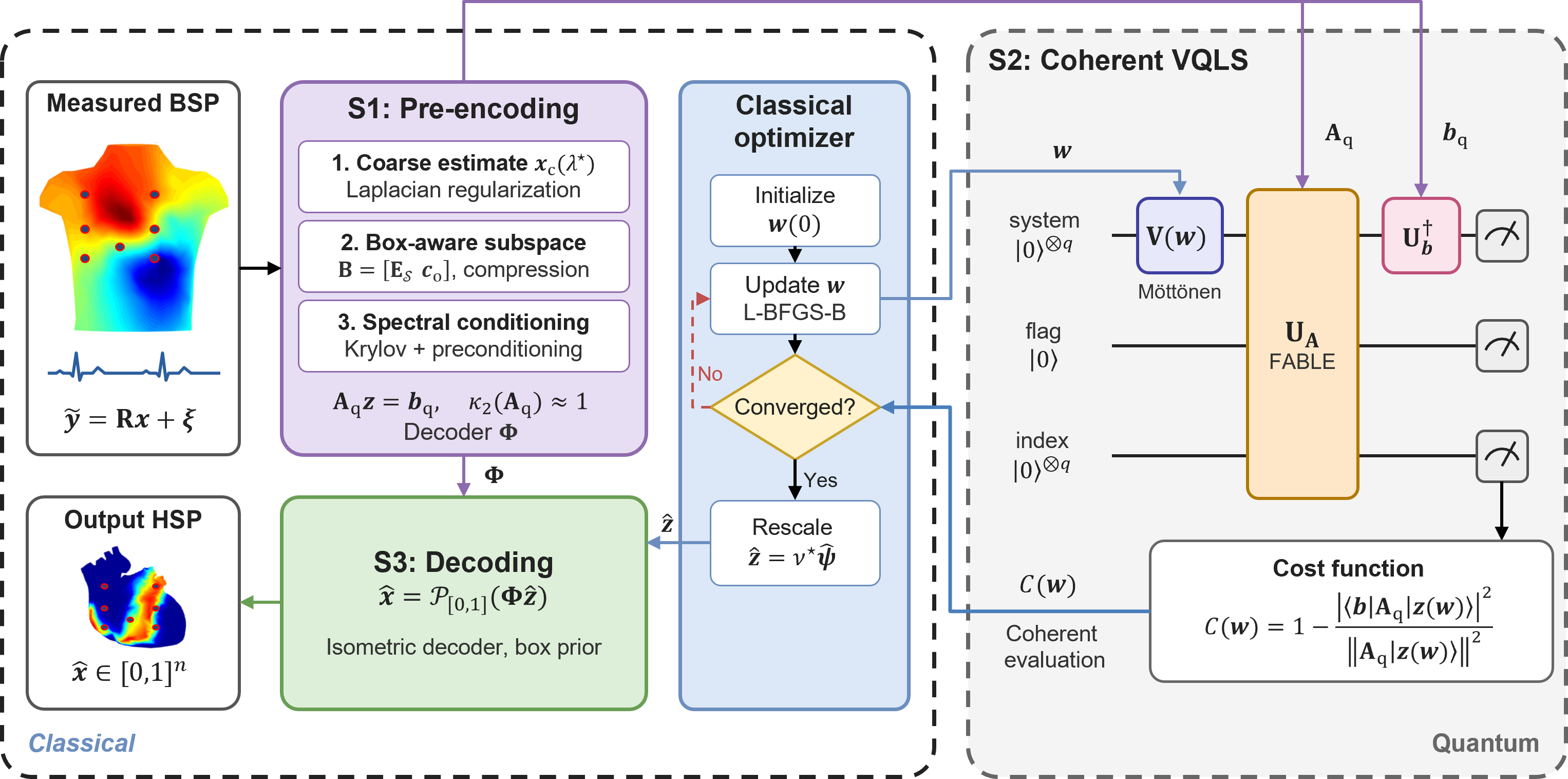} 
  % \fbox{\parbox[c][0.28\linewidth][c]{0.9\linewidth}{\centering Flowchart placeholder}}
  \caption{Flowchart of the proposed methodology. The pre-encoding stage transforms the ill-conditioned inverse system into a compact, well-conditioned reduced system and constructs an  isometric decoder. The coherent VQLS then solves the reduced system on a quantum register, where a M\"{o}tt\"{o}nen ansatz prepares the trial quantum state, a FABLE block encoding implements the system matrix, and two projector-based measurements evaluate the cost function for parameter optimization. The decoding stage further maps the reconstructed solution from the reduced representation back to the full set of cardiac nodes.}
  \label{Fig:flowchart}
\end{figure}

Fig.~\ref{Fig:flowchart} presents the proposed hybrid quantum-classical framework for reconstructing HSPs from BSPs. Let $\boldsymbol{x}(t) \in \mathbb{R}^{n}$ denote the HSP at the $n$ nodes of a triangulated heart surface at time $t$, and let $\boldsymbol{y}(t) \in \mathbb{R}^{m}$ denote the BSP at $m$ torso electrodes. In inverse ECG modeling, the HSP and the BSP are related through a transfer matrix $\mathbf{R} \in \mathbb{R}^{m \times n}$ determined by the heart-torso geometry, and the measured BSP is given by
\begin{equation}
    \tilde{\boldsymbol{y}}(t) = \mathbf{R}\boldsymbol{x}(t) + \boldsymbol{\xi}(t),
    \label{eq:forward}
\end{equation}
where $\boldsymbol{\xi}(t) \sim \mathcal{N}(\boldsymbol{0}, \sigma_{\boldsymbol{\xi}}^2\sigma^2\mathbf{I}_m)$ denotes the measurement noise, which follows a zero-mean Gaussian distribution. Here, $\sigma_{\boldsymbol{\xi}}$ is a dimensionless noise level, $\sigma$ is a reference variance scale of the BSP recording and $\mathbf{I}_m$ is an $m \times m$ identity matrix. Our objective is to estimate $\boldsymbol{x}(t)$ from $\tilde{\boldsymbol{y}}(t)$ via a VQLS \citep{bravo2019variational, lu2026distributed}.
% Since the reconstruction is carried out independently for each time instance, the time index $t$ is omitted hereafter for brevity. Throughout this paper, $\norm{\cdot}$ denotes the Euclidean norm for vectors and the induced spectral norm for matrices.

The proposed framework consists of three stages as shown in Fig.~\ref{Fig:flowchart}. First, in the pre-encoding stage (S1), the original inverse problem is reformulated into a compact linear system $\mathbf{A}_\text{q}\boldsymbol{z} = \boldsymbol{b}_\text{q}$ of dimension $N = 2^{q}$, where the system is constructed to have a condition number close to one to facilitate stable quantum solution. A linear decoder $\boldsymbol{\Phi}\in\mathbb{R}^{n\times N}$ is also constructed to map the reduced solution from the reduced representation back to the full set of cardiac nodes. Second, in the quantum solution stage (S2), the coherent VQLS solves the encoded linear system using a register of $q$ system qubits. Finally, in the decoding stage (S3), the resulting quantum solution is mapped through $\boldsymbol{\Phi}$ to recover the HSPs at all $n$ heart-surface nodes. For conciseness, the proofs of the theoretical properties of each stage are deferred to Supplementary Material~\ref{app:theory}, and the implementation details are provided in Supplementary Material~\ref{app:implementation}.

\subsection{S1: Pre-encoding to Construct a Well-Conditioned System}

% \textbf{\color{red}(Please learn from my revision in Section 1.1, and revise the relevant Supplementary Material section accordingly)}

\label{s:s1}
%The inverse ECG problem in Eq.~\eqref{eq:forward} is severely ill-posed. 
For a full-column-rank matrix $\mathbf{M}$ with singular values $\omega_1 \ge \dots \ge \omega_n > 0$, the spectral condition number $\kappa_2(\mathbf{M}) = \omega_1/\omega_n$ bounds the amplification of a relative data perturbation into a relative solution error \citep{golub2013matrix}. Specifically, for a noise-free BSP $\boldsymbol{y} = \mathbf{M}\boldsymbol{x}$ and the least-squares solution $\hat{\boldsymbol{x}} = \mathbf{M}^{+}(\boldsymbol{y} + \boldsymbol{\xi})$ of the perturbed system, the relative solution error is bounded as
\begin{equation}
    \frac{\norm{\hat{\boldsymbol{x}} - \boldsymbol{x}}}{\norm{\boldsymbol{x}}} \le \kappa_2(\mathbf{M}) \, \frac{\norm{\boldsymbol{\xi}}}{\norm{\boldsymbol{y}}},
    \label{eq:cond_bound}
\end{equation}
where $\mathbf{M}^{+}$ is the Moore-Penrose pseudoinverse of $\mathbf{M}$. Note that the transfer matrix $\mathbf{R}$ in Eq.~\eqref{eq:forward} is severely ill-conditioned, so that a small measurement perturbation can be amplified into a reconstruction error that is several orders of magnitude larger. %Thus, regularization is indispensable for reliable inverse ECG modeling \citep{tikhonov1977solutions, hansen1998rank, yao2021spatiotemporal}.

%Beyond this intrinsic ill-posedness, three properties of the physical system prevent it from being passed to a VQLS directly. First, a VQLS operates on a square system whose dimension is a power of two, and the natural square reformulation of Eq.~\eqref{eq:forward} is the normal equations $\mathbf{R}^\top\mathbf{R}\boldsymbol{x} = \mathbf{R}^\top\tilde{\boldsymbol{y}}$ of dimension $n$. A dense $n \times n$ system matrix requires $\log_2 n$ system qubits in addition to the ancilla qubits of its block encoding, and the number of gates in the block encoding grows as $n^{2}$ \citep{camps2022fable}. Second, the normal equations square the condition number, i.e., $\kappa_2(\mathbf{R}^\top\mathbf{R}) = \kappa_2(\mathbf{R})^{2}$.
% which reaches the order of $10^{10}$. 
%This is critical because the solution accuracy that a VQLS cost value can certify deteriorates in proportion to the condition number of the system matrix \citep{bravo2019variational} (see Eq.~\eqref{eq:vqls_bound} in Section~\ref{s:cost}). 
% Third, amplitude encoding represents only the direction of a vector, so that the magnitude of the solution is lost in the quantum state and must be restored classically.

We propose a pre-encoding stage that reformulates the ill-conditioned physical system into a compact and well-conditioned surrogate system before it is passed to the VQLS. Specifically, the pre-encoding constructs a triple $(\mathbf{A}_\text{q}, \boldsymbol{b}_\text{q}, \boldsymbol{\Phi})$ that satisfies three requirements: (i) $\mathbf{A}_\text{q} \in \mathbb{R}^{N \times N}$ with $N = 2^{q}$, such that the reduced solution is exactly the amplitude vector of a $q$-qubit register; (ii) $\kappa_2(\mathbf{A}_\text{q}) \approx 1$, such that the VQLS cost landscape is well shaped and a small cost value certifies an accurate solution; and (iii) the decoder $\boldsymbol{\Phi}$ maps the reduced solution to all the $n$ cardiac nodes.
%  (iv) every quantity in the construction depends only on $\mathbf{R}$, $\tilde{\boldsymbol{y}}$, and a spatial regularization matrix, and never on the reference HSP, such that the reported reconstruction accuracy is free of ground-truth leakage.
To construct such a triple, the pre-encoding proceeds in the successive steps, which are detailed in the following subsections. %In Section~\ref{s:coarse}, a Laplacian-regularized coarse estimate of the HSP is computed to locate the spatial support of the cardiac electrical activity. In Section~\ref{s:subspace}, this coarse estimate guides the construction of a low-dimensional orthonormal subspace, onto which the physical system is compressed so as to satisfy requirements (i) and (iii). In Section~\ref{s:krylov}, the compressed system is expressed in Krylov coordinates and conditioned by two invertible left multiplications, which preserve its solution while achieving requirement (ii).

\subsubsection{Laplacian-Regularized Coarse Estimation}
\label{s:coarse}

The coarse estimate of the HSP is obtained by solving the Tikhonov-regularized least-squares problem \citep{tikhonov1977solutions}
\begin{equation}
    \boldsymbol{x}_\text{c}(\lambda) = \arg\min_{\boldsymbol{x} \in \mathbb{R}^{n}} \; \norm{\mathbf{R}\boldsymbol{x} - \tilde{\boldsymbol{y}}}^2 + \lambda^2 \norm{\mathbf{D}\boldsymbol{x}}^2,
    \label{eq:tikhonov}
\end{equation}
where $\lambda > 0$ is the regularization weight, and $\mathbf{D} \in \mathbb{R}^{n \times n}$ is the surface Laplacian of the heart mesh, whose penalty $\norm{\mathbf{D}\boldsymbol{x}}^2$ suppresses spatial oscillations between neighboring nodes \citep{yao2021spatiotemporal}. 
%The solution of Eq.~\eqref{eq:tikhonov} provides the intermediate coarse estimate that is used to choose a low-dimensional subspace that can be handled by the quantum solver.
%{\color{orange}Solving Eq.~\eqref{eq:tikhonov} through its normal equations would, however, square the condition number of the problem. We therefore compute $\boldsymbol{x}_\text{c}(\lambda)$ as the solution of the equivalent stacked least-squares problem
%\begin{equation}
%    \boldsymbol{x}_\text{c}(\lambda) = \arg\min_{\boldsymbol{x} \in \mathbb{R}^{n}} \; \norm{ \begin{bmatrix} \mathbf{R} \\ \lambda\mathbf{D} \end{bmatrix} \boldsymbol{x} - \begin{bmatrix} \tilde{\boldsymbol{y}} \\ \boldsymbol{0} \end{bmatrix} }^2,
%    \label{eq:augmented}
%\end{equation}
%whose numerical sensitivity is governed by the condition number of the stacked matrix rather than by its square \citep{golub2013matrix}. {\color{red}\textbf{(Why you need this part? -> I want to explain the calculation details for numerical safeguard. It's not necessary)}}}
Note that the coarse estimate $\boldsymbol{x}_\text{c}(\lambda)$ is used to localize cardiac activation on the heart surface and construct the subspace for the inverse reconstruction, rather than to resolve fine activation patterns. %Accordingly, we select $\lambda$ to balance noise suppression against spatial over-smoothing, using an admissibility criterion based solely on {\color{red}observable quantities \textbf{(?)}}. Specifically, we choose the largest admissible value of $\lambda$, corresponding to the strongest regularization {\color{red}consistent with the measurements $\tilde{\boldsymbol{y}}$ \textbf{(?)}}. {\color{red}This choice reflects the asymmetric consequences of insufficient and excessive regularization for the intended purpose \textbf{(?)}}. 
Because the data residual $\norm{\mathbf{R}\boldsymbol{x}_{\mathrm{c}}(\lambda)-\tilde{\boldsymbol{y}}}_2$ is nondecreasing with $\lambda$, admissibility criteria that impose an upper bound on this residual define a range of acceptable values. Within this range, insufficient regularization can substantially amplify measurement noise, whereas stronger regularization progressively smooths the estimate. Because accurate localization of the active region is more important than resolving fine spatial details for subspace selection, we select the upper admissibility threshold for $\lambda$ to limit noise amplification while retaining consistency with the measurements. 

%The regularization weight governs the trade-off between noise amplification and over-smoothing.  We select $\lambda$ as the {\color{red}largest value} that satisfies an admissibility criterion computed from observable quantities only, which corresponds to the strongest regularization supported by the data . The largest value is taken for the following concern. As the data residual $\norm{\mathbf{R}\boldsymbol{x}_\text{c}(\lambda) - \tilde{\boldsymbol{y}}}$ increases monotonically with $\lambda$, each criterion introduced below is satisfied by every weight up to a threshold, so that the criterion alone does not single out one weight. Among these admissible weights, the consequences of a poor choice are not symmetric: a weight that is too small amplifies the measurement noise by a factor of up to $\kappa_2(\mathbf{R})$, whereas a weight that is too large only smooths the estimate gradually. Since $\boldsymbol{x}_\text{c}$ is required to locate the active region rather than to resolve its fine structure, we take the admissible weight that is farthest from the noise-amplifying regime, i.e., the threshold itself.

For noise-free data, $\lambda$ is selected based on a forward-error certificate. Specifically, let $\rho(\boldsymbol{x}) = \norm{\mathbf{R}\boldsymbol{x} - \tilde{\boldsymbol{y}}}/\norm{\tilde{\boldsymbol{y}}}$ denote the relative forward residual of a candidate HSP $\boldsymbol{x}$, and $\boldsymbol{x}^\star$ is the true HSP. If $\norm{\boldsymbol{\xi}} \le \eta\norm{\tilde{\boldsymbol{y}}}$ for some $\eta \in [0,1)$, then every candidate $\boldsymbol{x}$ satisfies (Proposition~\ref{prop:certificate} in Supplementary Material~\ref{app:certificate})
\begin{equation}
    \frac{\norm{\boldsymbol{x} - \boldsymbol{x}^\star}}{\norm{\boldsymbol{x}^\star}} \le \beta(\boldsymbol{x}, \eta) := \kappa_2(\mathbf{R}) \, \frac{\rho(\boldsymbol{x}) + \eta}{1 - \eta}.
    \label{eq:certificate}
\end{equation}
% where every quantity on the right-hand side is available without $\boldsymbol{x}^\star$. 
For noise-free data ($\eta = 0$), the regularization weight is selected as
\begin{equation}
    \lambda^\star = \max\left\{ \lambda > 0 : \beta\bigl(\boldsymbol{x}_\text{c}(\lambda), 0\bigr) \le \tau \right\},
    \label{eq:lambda_cert}
\end{equation}
where $\tau$ is a prescribed level of the certified relative error. %{\color{red}(\textbf{How to specify $\tau$ in our study? -> It's a hyper-parameter. In the experiment, we set $\tau = 0.10$, i.e., the coarse estimate is required to carry a certified relative error of at most 10\%. Since Eq.~\eqref{eq:certificate} is a worst-case bound, and the actual error of the coarse estimate is several times smaller than $\tau$ in our experiments. This level is sufficient because the coarse estimate is only used to locate the active region.})}. 
%Under measurement noise, however, the certificate cannot be used, because its noise-dependent term $\kappa_2(\mathbf{R})\,\eta/(1-\eta)$ is independent of $\lambda$ and, for a severely ill-conditioned $\mathbf{R}$, exceeds any practical $\tau$, so that no weight would be admissible.
% with $\kappa_2(\mathbf{R})$ of the order of $10^{5}$, this term alone exceeds $\tau = 0.10$ once $\eta$ exceeds approximately $10^{-6}$, so that no weight would be admissible. 
For noisy data, we adopt Morozov's discrepancy principle \citep{morozov1966solution}, which selects the smoothest estimate that explains the measurements up to the noise level:
\begin{equation}
    \lambda^\star = \max\left\{ \lambda > 0 : \norm{\mathbf{R}\boldsymbol{x}_\text{c}(\lambda) - \tilde{\boldsymbol{y}}} \le \norm{\boldsymbol{\xi}} \right\},
    \label{eq:lambda_morozov}
\end{equation}
where $\norm{\boldsymbol{\xi}}$ is the norm of the measurement noise. The coarse estimate is then fixed as $\boldsymbol{x}_\text{c} = \boldsymbol{x}_\text{c}(\lambda^\star)$. Please refer to Supplementary Material~\ref{app:lambda_search} for more details on the selection of $\lambda$.
% In the simulation study, $\norm{\boldsymbol{\xi}}$ is the norm of the realized noise; in clinical practice, it is replaced by an estimate of the noise level.
%As the residual norm $\norm{\mathbf{R}\boldsymbol{x}_\text{c}(\lambda) - \tilde{\boldsymbol{y}}}$ increases monotonically with $\lambda$ \citep{hansen1998rank}, both criteria define a threshold on $\lambda$, which is located by a descending logarithmic scan followed by bisection on the logarithmic scale (Supplementary Material~\ref{app:lambda_search}). 

\subsubsection{Box-Aware Subspace Construction and Compression}
\label{s:subspace}

The coarse estimate $\boldsymbol{x}_\text{c}$ is used to select a subset of $s$ ($s<n$) heart-surface nodes, whose potentials are retained as separate unknowns in the reduced system; the remaining unselected nodes share a single unknown potential. To allocate these individual degrees of freedom to the active region, we select the nodes with the largest coarse potential values after enforcing the physiological range of the adopted model. Note that the HSP is modeled as a normalized transmembrane potential bounded in $[0,1]$, whereas the coarse estimate often contains negative or overshooting artifacts. Ranking the nodes by $|x_{\text{c},i}|$ would promote such artifacts. We instead score each node by its box-projected coarse value
\begin{equation}
    p_i = \min\bigl(\max(x_{\text{c},i}, 0), 1\bigr), \qquad i = 1, \dots, n,
    \label{eq:box_score}
\end{equation}
where $x_{\text{c},i}$ is the $i$-th entry of $\boldsymbol{x}_\text{c}$. The active support $\mathcal{S} \subset \{1, \dots, n\}$ is then defined as the set of the $s$ nodes with the largest scores $p_i$. %Because cardiac electrical activity propagates across the heart surface over time, the support is selected independently at each time instant, allowing the support to adapt to the evolving activation region. %Note that the score is the clipped coarse value rather than its magnitude, a node with a large negative artifact receives a zero score and therefore cannot displace a physically admissible activation. And the score is evaluated separately for each time instance rather than as a temporal norm of the estimated signal at each node. Since the activated region propagates across the heart surface during the cardiac cycle, a support that is fixed over time cannot follow it.

Let $\mathbf{E}_{\mathcal{S}} \in \mathbb{R}^{n \times s}$ denote the selection matrix whose columns are the canonical basis vectors $\boldsymbol{e}_i$, $i \in \mathcal{S}$. A subspace spanned by $\mathbf{E}_{\mathcal{S}}$ alone forces the potential at every node outside $\mathcal{S}$ to zero. To represent the remaining nodes with one additional degree of freedom, we augment this basis with the normalized off-support component of the coarse estimate:
\begin{equation}
    \boldsymbol{c}_\text{o} = \frac{(\mathbf{I}_n - \mathbf{E}_{\mathcal{S}}\mathbf{E}_{\mathcal{S}}^\top)\,\boldsymbol{x}_\text{c}}{\norm{(\mathbf{I}_n - \mathbf{E}_{\mathcal{S}}\mathbf{E}_{\mathcal{S}}^\top)\,\boldsymbol{x}_\text{c}}}, \qquad
    \mathbf{B} = \bigl[\, \mathbf{E}_{\mathcal{S}} \;\; \boldsymbol{c}_\text{o} \,\bigr] \in \mathbb{R}^{n \times k}, \qquad k = s + 1.
    \label{eq:basis}
\end{equation}
Because $\boldsymbol{c}_\text{o}$ vanishes on $\mathcal{S}$, it is orthogonal to every column of $\mathbf{E}_{\mathcal{S}}$ and $\mathbf{B}^\top\mathbf{B} = \mathbf{I}_k$ holds without explicit orthogonalization. By construction, $\boldsymbol{x}_\text{c} = \mathbf{E}_{\mathcal{S}}\mathbf{E}_{\mathcal{S}}^\top\boldsymbol{x}_\text{c} + \norm{(\mathbf{I}_n - \mathbf{E}_{\mathcal{S}}\mathbf{E}_{\mathcal{S}}^\top)\,\boldsymbol{x}_\text{c}}\,\boldsymbol{c}_\text{o}$, so that the coarse estimate lies exactly in $\operatorname{range}(\mathbf{B})$. Note that $\mathbf{B}$ stores directions rather than the values of the coarse estimate. In other words, the coarse estimate serves only to construct the subspace used by the compressed quantum solver in the next step, and the coefficients of the reconstruction will be determined by solving the compressed system. %The columns of $\mathbf{E}_{\mathcal{S}}$ depend on $\boldsymbol{x}_\text{c}$ only through the selected indices, and $\boldsymbol{c}_\text{o}$ retains only the normalized off-support profile. The coefficients along the $k$ directions are subsequently determined by the measurements through the compressed system below, so that the reconstruction is, in general, an element of $\operatorname{range}(\mathbf{B})$ different from $\boldsymbol{x}_\text{c}$. In other words, $\mathbf{B}$ only fixes the form of the reconstruction, $\boldsymbol{x} = \mathbf{B}\boldsymbol{u}$, which combines the $s$ selected nodes and the background profile $\boldsymbol{c}_\text{o}$ with $k$ weights collected in $\boldsymbol{u} \in \mathbb{R}^{k}$. However, these weights are unknowns, which are not copied from $\boldsymbol{x}_\text{c}$, but are computed from the measurements $\tilde{\boldsymbol{y}}$ in the next step. The resulting reconstruction therefore generally differs from $\boldsymbol{x}_\text{c}$, although both lie in $\operatorname{range}(\mathbf{B})$. 
Under the construction of $s = N - 1$, the subspace dimension $k$ equals the dimension $N$ of the quantum register.

We restrict the reconstruction to the subspace spanned by $\mathbf{B}$ through $\boldsymbol{x}=\mathbf{B}\boldsymbol{u}$, where $\boldsymbol{u}\in\mathbb{R}^{k}$ contains the unknown coefficients, which yields the compressed system $\mathbf{G}\boldsymbol{u} = \boldsymbol{g}$ with
\begin{equation}
    \mathbf{G} = (\mathbf{R}\mathbf{B})^\top(\mathbf{R}\mathbf{B}) \in \mathbb{R}^{k \times k}, \qquad
    \boldsymbol{g} = (\mathbf{R}\mathbf{B})^\top\tilde{\boldsymbol{y}} \in \mathbb{R}^{k}.
    \label{eq:compress}
\end{equation}
Note that the compressed system carries no explicit penalty term. Instead the Laplacian regularization enters the reduced problem through the construction of the subspace $\mathbf{B}$, which determines the admissible spatial pattern. This subspace restriction, $\boldsymbol{x} \in \operatorname{range}(\mathbf{B})$, is the only approximation introduced by the pre-encoding stage, and all subsequent transformations preserve the solution of the compressed system exactly (Proposition~\ref{prop:invariance}).

\subsubsection{Krylov Coordinates and Spectral Conditioning}
\label{s:krylov}

Although compression reduces the system dimension, it does not guarantee good conditioning. $\mathbf{G}$ inherits a large condition number from $\mathbf{R}$. We apply three transformations to make the surrogate system well-conditioned for the quantum solver. First, the system is expressed in the coordinates of the Krylov subspace generated by $\mathbf{G}$ and $\boldsymbol{g}$ \citep{saad2003iterative}:
\begin{equation}
    \mathcal{K}_N(\mathbf{G}, \boldsymbol{g}) = \operatorname{span}\bigl\{ \boldsymbol{g}, \mathbf{G}\boldsymbol{g}, \mathbf{G}^2\boldsymbol{g}, \dots, \mathbf{G}^{N-1}\boldsymbol{g} \bigr\}.
\end{equation}
We construct an orthonormal basis $\mathbf{Q} \in \mathbb{R}^{k \times N}$ of $\mathcal{K}_N(\mathbf{G}, \boldsymbol{g})$ using the Lanczos process \citep{lanczos1950iteration} with full reorthogonalization. Writing $\boldsymbol{u} = \mathbf{Q}\boldsymbol{z}$, the compressed system is then projected as
\begin{equation}
\mathbf{T}\boldsymbol{z} = \boldsymbol{h}, \qquad
    \mathbf{T} = \mathbf{Q}^\top\mathbf{G}\mathbf{Q} \in \mathbb{R}^{N \times N}, \qquad
    \boldsymbol{h} = \mathbf{Q}^\top\boldsymbol{g} \in \mathbb{R}^{N}.
    \label{eq:krylov}
\end{equation}
%The Lanczos process generates the columns $\boldsymbol{q}_1, \dots, \boldsymbol{q}_N$ of $\mathbf{Q}$ one at a time, which starts from $\boldsymbol{q}_1 = \boldsymbol{g}/\norm{\boldsymbol{g}}$ and obtains each new vector $\boldsymbol{q}_{j+1}$ by applying $\mathbf{G}$ to $\boldsymbol{q}_j$ and removing the components along the previous vectors. This construction has two consequences. First, since $\mathbf{G}$ is symmetric, $\mathbf{G}\boldsymbol{q}_j$ has components only along $\boldsymbol{q}_{j-1}$, $\boldsymbol{q}_j$, and $\boldsymbol{q}_{j+1}$, so that the entries $\boldsymbol{q}_i^\top\mathbf{G}\boldsymbol{q}_j$ of $\mathbf{T}$ vanish whenever $|i - j| > 1$. That is, $\mathbf{T}$ is tridiagonal, and each unknown is coupled only to its two neighbors. Second, since $\boldsymbol{g}$ is parallel to $\boldsymbol{q}_1$ and orthogonal to all other columns of $\mathbf{Q}$, the right-hand side $\boldsymbol{h} = \mathbf{Q}^\top\boldsymbol{g} = \norm{\boldsymbol{g}}\boldsymbol{e}_1$ has a single nonzero entry. Both properties hold exactly in the absence of rounding errors, which are controlled by the reorthogonalization described in Supplementary Material~\ref{app:lanczos}.With $k = N$, $\mathbf{Q}$ is a square orthogonal matrix, so that this step is an exact change of coordinates that concentrates the coupling of the system near the diagonal of $\mathbf{T}$. %{\color{red}\textbf{(What's the resulting system after the transformation in Step 1?)}}
The Lanczos process transforms the compressed system into a tridiagonal form with a right-hand side containing a single nonzero entry, providing the structure used in the subsequent transformations. Specifically, starting from $\boldsymbol{q}_1=\boldsymbol{g}/\|\boldsymbol{g}\|_2$, with $\boldsymbol{g}\neq\boldsymbol{0}$, it constructs each new basis vector by applying $\mathbf{G}$ to the preceding vector, removing its components along the existing basis vectors, and normalizing the remainder. As such, $\mathbf{G}\boldsymbol{q}_j$ lies in the span of $\boldsymbol{q}_1,\dots,\boldsymbol{q}_{j+1}$. The symmetry of $\mathbf{G}$ further implies that its components along $\boldsymbol{q}_1,\dots,\boldsymbol{q}_{j-2}$ vanish: for $i<j-1$, $ \boldsymbol{q}_i^\top\mathbf{G}\boldsymbol{q}_j  =(\mathbf{G}\boldsymbol{q}_i)^\top\boldsymbol{q}_j=0,$ because $\mathbf{G}\boldsymbol{q}_i$ lies in the span of $\boldsymbol{q}_1,\dots,\boldsymbol{q}_{i+1}$, which is orthogonal to $\boldsymbol{q}_j$. Thus, $\mathbf{G}\boldsymbol{q}_j$ involves at most three consecutive basis vectors, leading to a tridiagonal form of $\mathbf{T}$. Additionally, the initialization with $\boldsymbol{g}$ also simplifies the transformed right-hand side as
   $\boldsymbol{h}
    =\mathbf{Q}^\top\boldsymbol{g}
    =\|\boldsymbol{g}\|_2\boldsymbol{e}_1$.
Thus, the transformed system combines a tridiagonal matrix with a right-hand side containing a single nonzero entry. %These properties hold in exact arithmetic (Supplementary Material~\ref{app:lanczos}).

Second, we apply block-Jacobi left preconditioning \citep{saad2003iterative} to improve the conditioning of the transformed system. We partition the index set $\{1, \dots, N\}$ into $n_\text{b}$ contiguous blocks, and let $\mathbf{P} = \operatorname{blockdiag}(\mathbf{T}_{11}, \dots, \mathbf{T}_{n_\text{b} n_\text{b}})$ collect the corresponding diagonal blocks of $\mathbf{T}$. Each block is symmetric, positive definite and admits a Cholesky factorization $\mathbf{T}_{jj} = \mathbf{L}_j\mathbf{L}_j^\top$. 
%{\color{blue}This property is inherited from the compressed system. Since $\mathbf{R}$ has full column rank and $\mathbf{B}$ has orthonormal columns, $\boldsymbol{u}^\top\mathbf{G}\boldsymbol{u} = \norm{\mathbf{R}\mathbf{B}\boldsymbol{u}}^2 > 0$ for every $\boldsymbol{u} \neq \boldsymbol{0}$, so that $\mathbf{G}$ is symmetric positive definite. The same holds for $\mathbf{T} = \mathbf{Q}^\top\mathbf{G}\mathbf{Q}$ because $\mathbf{Q}$ is orthogonal, and to each diagonal block of $\mathbf{T}$, since the quadratic form of a block corresponds to restricting $\mathbf{T}$ to vectors with support only on that block (see Proposition~\ref{prop:invariance}(a)).}
Specifically, the preconditioned system is
\begin{equation}
    \mathbf{A}_0 = \mathbf{P}^{-1}\mathbf{T}, \qquad \boldsymbol{b}_0 = \mathbf{P}^{-1}\boldsymbol{h}.
    \label{eq:precond}
\end{equation}
%where $\mathbf{P}^{-1}$ is applied by triangular solves with $\mathbf{L} = \operatorname{blockdiag}(\mathbf{L}_1, \dots, \mathbf{L}_{n_\text{b}})$ and $\mathbf{L}^\top$ directly. 
The block partition is selected by minimizing $\kappa_2(\mathbf{P}^{-1}\mathbf{T})$ over the admissible contiguous partitions subject to the prescribed block-size bounds (Supplementary Material~\ref{app:partition}). %We adopt left rather than symmetric preconditioning deliberately. The symmetric alternative $\mathbf{L}^{-1}\mathbf{T}\mathbf{L}^{-\top}$ preserves symmetry but changes the solution coordinates, so that the decoder would become $\mathbf{B}\mathbf{Q}\mathbf{L}^{-\top}$, whose condition number equals $\kappa_2(\mathbf{L})$, and the residual error of the quantum solver would be amplified on its way back to the heart mesh (Supplementary Material~\ref{app:isometry}). 
% Since the VQLS does not require a symmetric system matrix, left preconditioning is preferred.
We use left preconditioning instead of symmetric preconditioning to preserve the solution coordinates and the isometric reconstruction map (see Supplementary Material~\ref{app:isometry}).

  Third, we apply a further left preconditioner to control the possible extreme singular values that determine the condition number of $\mathbf{A}_0$. This correction exploits the structure remaining after block-Jacobi preconditioning. Specifically, from $\mathbf{A}_0=\mathbf{P}^{-1}\mathbf{T}$, we have $
    \mathbf{A}_0-\mathbf{I}_N
    =\mathbf{P}^{-1}(\mathbf{T}-\mathbf{P})$.
Because $\mathbf{T}$ is tridiagonal and $\mathbf{P}$ contains its diagonal blocks, $\mathbf{T}-\mathbf{P}$ contains only the two symmetric coupling entries across each boundary between adjacent blocks. Each boundary contributes a matrix of rank at most two. Because there are $n_{\mathrm{b}}-1$ boundaries and multiplication by $\mathbf{P}^{-1}$ preserves rank, i.e., $
    \operatorname{rank}(\mathbf{A}_0-\mathbf{I}_N)
    =
    \operatorname{rank}(\mathbf{T}-\mathbf{P})
    \leq d,
    ~
    d=2(n_{\mathrm{b}}-1)$.
Hence, at most $d$ singular values of $\mathbf{A}_0$ exceed one and at most $d$ fall below one; the remaining singular values equal one in exact arithmetic (Supplementary Material~\ref{app:invariance}). This structure motivates a correction confined to the two ends of the singular-value spectrum.

Let $
    \mathbf{A}_0
    =\mathbf{U}_{\mathrm{L}}\boldsymbol{\Gamma}
    \mathbf{U}_{\mathrm{R}}^\top,~
    \gamma_1\geq\cdots\geq\gamma_N>0$,
and let $r$ denote the number of singular values corrected at each end, with $2r<N$. We retain the central singular values and move each extreme value to the nearest endpoint of the retained interval. Specifically, define
\begin{equation}
    f_i=
    \begin{cases}
        \gamma_{r+1}/\gamma_i,
        & 1\leq i\leq r,\\
        1,
        & r<i\leq N-r,\\
        \gamma_{N-r}/\gamma_i,
        & N-r<i\leq N,
    \end{cases}
\end{equation}
and construct
\begin{equation}
    \mathbf{F}_{\mathrm{tail}}
    =
    \mathbf{I}_N
    +
    \mathbf{U}_{\mathrm{L},\mathcal{J}}
    \operatorname{diag}
    \bigl(\boldsymbol{f}_{\mathcal{J}}-\boldsymbol{1}\bigr)
    \mathbf{U}_{\mathrm{L},\mathcal{J}}^\top,
    \qquad
    \mathcal{J}
    =
    \{1,\dots,r\}\cup\{N-r+1,\dots,N\}.
    \label{def-J}
\end{equation}
Here, $\mathbf{U}_{\mathrm{L},\mathcal{J}}$ contains the left singular vectors indexed by $\mathcal{J}$, and $\boldsymbol{f}_{\mathcal{J}}$ contains their rescaling factors. Multiplication by $\mathbf{F}_{\mathrm{tail}}$ reduces the $r$ largest singular values to $\gamma_{r+1}$ and raises the $r$ smallest to $\gamma_{N-r}$, leaving the central singular values unchanged. Choosing $r\geq d$ covers every singular value that can differ from one under the rank bound. %In this study, the partition contains at most five blocks, so we use $r=8$, requiring $N>16$.

We apply this correction to both sides of the preconditioned system, yielding the encoded system and reconstruction map as
\begin{equation}
    \mathbf{A}_{\mathrm{q}}
    =\mathbf{F}_{\mathrm{tail}}\mathbf{A}_0,
    \qquad
    \boldsymbol{b}_{\mathrm{q}}
    =\mathbf{F}_{\mathrm{tail}}\boldsymbol{b}_0,
    \qquad
    \boldsymbol{\Phi}=\mathbf{B}\mathbf{Q}.
    \label{eq:encoded}
\end{equation}
The quantum solver then operates on
$\mathbf{A}_{\mathrm{q}}\boldsymbol{z}=\boldsymbol{b}_{\mathrm{q}}$,
and the reconstructed potentials are recovered through
$\boldsymbol{x}=\boldsymbol{\Phi}\boldsymbol{z}$.
Because both left preconditioners are invertible, and are applied to the matrix and right-hand side together, they preserve the solution coordinates. Hence, the exact decoded solution remains the least-squares reconstruction restricted to $\operatorname{range}(\mathbf{B})$.
Note that this correction confines the singular values of $\mathbf{A}_{\mathrm{q}}$ to $[\gamma_{N-r},\gamma_{r+1}]$, giving $
    \kappa_2(\mathbf{A}_{\mathrm{q}})
    =\frac{\gamma_{r+1}}{\gamma_{N-r}}$.
Under the exact-rank bound and $r\geq d$, both endpoints equal one, so $\kappa_2(\mathbf{A}_{\mathrm{q}})=1$ in exact arithmetic (Proposition~\ref{prop:invariance}). Finally, because $\boldsymbol{\Phi}^\top\boldsymbol{\Phi}=\mathbf{I}_N$, decoding preserves the Euclidean norm of coefficient errors. As such, the spectral correction controls conditioning while preserving both the compressed-system solution and the isometric reconstruction map (Proposition~\ref{prop:isometry}).

\subsection{S2: Coherent Variational Quantum Linear Solver (VQLS)}
\label{s:vqls}

\begin{figure}[!ht]
	\begin{center}
		\includegraphics[width=\textwidth]{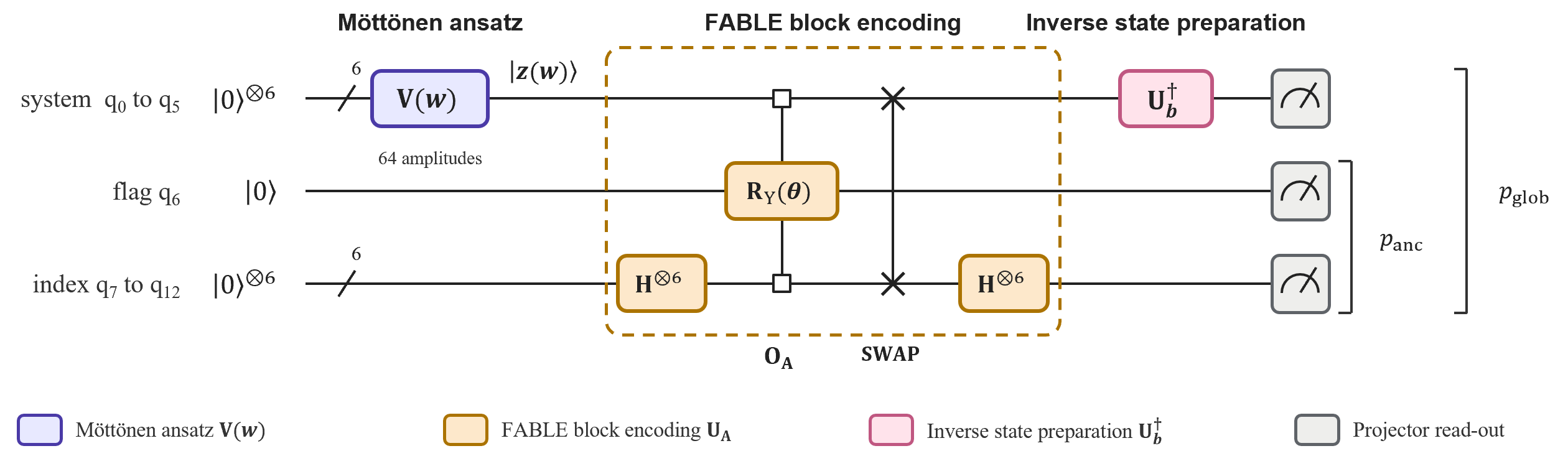}
		\caption{Coherent VQLS circuit of the complete model ($q = 6$ qubits, 13 total qubits), shown at register level: M\"{o}tt\"{o}nen state preparation $\mathbf{V}(\boldsymbol{w})$, FABLE block encoding $\mathbf{U}_{\mathbf{A}}$, inverse state preparation $\mathbf{U}_{\boldsymbol{b}}^\dagger$, and projector read-outs.
        %{\color{red}\textbf{(What is un-prepare?->Revised. "Un-prepare" is same with "inverse state preparation". The forward preparation $\mathbf{U}_{\boldsymbol{b}}$ maps $|0\rangle^{\otimes 6} \mapsto |\boldsymbol{b}\rangle$. Its inverse runs that map backwards: $\mathbf{U}_{\boldsymbol{b}}^\dagger|\boldsymbol{b}\rangle = |0\rangle^{\otimes 6}$. "Un-prepare" means turning $|\boldsymbol{b}\rangle$ back into the all-zero state.)}}
        }
		\label{Fig:circuit_register}
	\end{center}
\end{figure}

%The encoded system $\mathbf{A}_\text{q}\boldsymbol{z} = \boldsymbol{b}_\text{q}$ is solved by a VQLS \citep{bravo2019variational}. In contrast to fault-tolerant quantum linear system algorithms \citep{harrow2009quantum}, the VQLS is a hybrid quantum-classical algorithm, in which a parameterized quantum circuit prepares a trial solution state, circuit measurementsprovide the probabilities used to evaluate the cost function, and a classical optimizer updates the circuit parameters \citep{cerezo2021variational, lu2026distributed, lu2026memory}. 
%The pre-encoding stage yields the system $\mathbf{A}_{\mathrm{q}}\boldsymbol{z}=\boldsymbol{b}_{\mathrm{q}}$ and the reconstruction map $\boldsymbol{x}=\boldsymbol{\Phi}\boldsymbol{z}$. 
We propose to use a VQLS to prepare a quantum state approximating the normalized solution of encoded system from Section \ref{s:s1} \citep{bravo2019variational}. %VQLS combines quantum circuit evaluation with classical optimization: a parameterized circuit prepares a trial solution state, measurements provide estimates of the probabilities defining the cost function, and a classical optimizer updates the circuit parameters to minimize that cost \citep{cerezo2021variational, lu2026distributed}.
Fig.~\ref{Fig:circuit_register} shows the coherent VQLS circuit adopted in this study, which consists of a trial state preparation $\mathbf{V}(\boldsymbol{w})$, a block encoding $\mathbf{U}_{\mathbf{A}}$ of the system matrix, an inverse state preparation $\mathbf{U}_{\boldsymbol{b}}^\dagger$
for $\ket{\boldsymbol{b}} =
\boldsymbol{b}_\text{q}/\norm{\boldsymbol{b}_\text{q}}$,
the normalized right-hand-side vector of the encoded system, and two projector read-outs. Each component is detailed as follows.

\subsubsection{Amplitude Encoding and M\"{o}tt\"{o}nen Ansatz}
\label{s:ansatz}

%{\color{orange}To process the encoded system on a quantum computer, the candidate solution is represented by amplitude encoding, which uses a normalized classical vector as the probability amplitudes of a quantum state \citep{nielsen2010quantum, burns2025quantum}. Amplitude encoding preserves only the solution's direction, with its norm restored classically after optimization (see Section~\ref{s:optimization}). \textbf{\color{red}(Writing needs to be improved. Why use amplitude encoding? ->Amplitude encoding means using the numbers in a normalized vector as the coefficients of a quantum state. Those coefficients are called amplitudes. It help to design a compact quantum representation, because the qubit count alone doesn't establish a computational speedup. I attached a small example in Examples.tex for your understanding {\color{green}(I understand what the amplitude encoding is. You need to provide the readers the reasons why you choose amplitude encoding...->Revised)}.)} Specifically, the trial solution state is prepared by the parameterized circuit $\mathbf{V}(\boldsymbol{w})$ as
%\begin{equation}
%    \ket{\boldsymbol{z}(\boldsymbol{w})} = \mathbf{V}(\boldsymbol{w})\ket{0}^{\otimes q} = \frac{1}{\norm{\boldsymbol{w}}}\sum_{i=0}^{N-1} w_i \ket{i},
%    \label{eq:ansatz}
%\end{equation}}

To process the encoded system on a quantum computer, the candidate solution is represented by amplitude encoding, which is among the most widely used schemes for loading classical data into quantum states and employs a normalized classical vector as the probability amplitudes of a quantum state \citep{hur2022quantum, burns2025quantum}. Specifically, the trial solution is encoded into a $q$-qubit quantum state, which is prepared by a parameterized circuit $\mathbf{V}(\boldsymbol{w})$ as
\begin{equation}
    \ket{\boldsymbol{z}(\boldsymbol{w})} = \mathbf{V}(\boldsymbol{w})\ket{0}^{\otimes q} = \frac{1}{\norm{\boldsymbol{w}}}\sum_{i=0}^{N-1} w_i \ket{i},
    \label{eq:ansatz}
\end{equation}
where $\ket{0}^{\otimes q}$ denotes the all-zero state of the $q$ system qubits, $\ket{i}$ is the $i$-th computational basis state of the $q$-qubit register, and $\boldsymbol{w} = (w_0, \dots, w_{N-1})^\top \in \mathbb{R}^{N}$ is the trainable parameter vector. We construct $\mathbf{V}(\boldsymbol{w})$ by M\"{o}tt\"{o}nen state preparation \citep{mottonen2004transformation}, which synthesizes a cascade of uniformly controlled $\mathbf{R}_\text{Y}$ rotations that set the magnitudes of the amplitudes, followed by uniformly controlled $\mathbf{R}_\text{Z}$ rotations that set their signs. The rotation angles are computed from $\boldsymbol{w}$, and the rotation operations are defined as
\begin{equation}
    \mathbf{R}_\text{Y}(\theta) =
    \begin{bmatrix}
        \cos\frac{\theta}{2} & -\sin\frac{\theta}{2} \\
        \sin\frac{\theta}{2} & \cos\frac{\theta}{2}
    \end{bmatrix}, \qquad
    \mathbf{R}_\text{Z}(\theta) =
    \begin{bmatrix}
        e^{-\mathrm{i}\theta/2} & 0 \\
        0 & e^{\mathrm{i}\theta/2}
    \end{bmatrix},
    \label{eq:rotations}
\end{equation}
%In contrast to layered hardware-efficient ansatzes, whose reachable states form a subset of the state space determined by the circuit layout, the M\"{o}tt\"{o}nen ansatz reaches every real unit vector in $\mathbb{R}^{N}$ (Proposition~\ref{prop:expressivity}).
This construction can prepare any normalized real vector in \(\mathbb{R}^{N}\), up to a global phase, ensuring that the normalized solution of the encoded system is representable by the trial-state circuit (Proposition~\ref{prop:expressivity}).
% This choice therefore excludes representational error by construction: a failure of the VQLS can be attributed to the cost landscape of the encoded system rather than to an insufficient expressivity of the variational family, which is essential for isolating the contribution of the pre-encoding. It is worth noting that the number of parameters and the circuit depth of the M\"{o}tt\"{o}nen ansatz grow as $2^{q}$. The ansatz is thus adopted here as a methodological instrument for a controlled study rather than as a scalable hardware design.

\subsubsection{FABLE Block Encoding}
\label{s:fable}

%Because quantum circuits implement unitary operations whereas $\mathbf{A}_\text{q}$ is a general real matrix, $\mathbf{A}_\text{q}$ is embedded as the upper-left block of a larger unitary $\mathbf{U}_{\mathbf{A}}$ that acts on the system register and $n_\text{anc}$ ancilla qubits \citep{gilyen2019quantum}:
To evaluate the VQLS cost, we require a circuit representation of the action of \(\mathbf{A}_{\mathrm q}\) on the trial state. Because \(\mathbf{A}_{\mathrm q}\) is generally nonunitary, we use block encoding to embed \(\mathbf{A}_{\mathrm q}\) into a larger unitary operator \(\mathbf{U}_{\mathbf A}\) acting on the system register and \(n_{\mathrm{anc}}\) ancilla qubits \citep{gilyen2019quantum}:
\begin{equation}
    \bigl(\bra{0}^{\otimes n_\text{anc}} \otimes \mathbf{I}_N\bigr) \, \mathbf{U}_{\mathbf{A}} \, \bigl(\ket{0}^{\otimes n_\text{anc}} \otimes \mathbf{I}_N\bigr) = \frac{\mathbf{A}_\text{q}}{\alpha},
    \label{eq:block_encoding}
\end{equation}
where $\alpha \ge \norm{\mathbf{A}_\text{q}}$, ensuring that the encoded block has spectral norm at most one. Applying $\mathbf{U}_{\mathbf{A}}$ to $\ket{0}^{\otimes n_\text{anc}}\ket{\boldsymbol{z}}$ thus leaves $\mathbf{A}_\text{q}\ket{\boldsymbol{z}}/\alpha$ on the system register in the branch in which all ancilla qubits are in state $\ket{0}$. We realize $\mathbf{U}_{\mathbf{A}}$ by the FABLE construction
\citep{camps2022fable}, which adopts a \(q\)-qubit system register to hold the trial state and \(q+1\) ancilla qubits to construct the block encoding. These ancillas comprise a \(q\)-qubit index register, which labels matrix rows during the oracle operation, and a flag qubit, whose amplitudes encode the corresponding matrix entries. To encode the entries \((\mathbf A_{\mathrm q})_{ij}\) as amplitudes, we first rescale them to lie within \([-1,1]\): $c_\text{F} = \max\{1, \max_{i,j}|(\mathbf{A}_\text{q})_{ij}|\}$ and $\bar{a}_{ij} = (\mathbf{A}_\text{q})_{ij}/c_\text{F} \in [-1, 1]$. FABLE first constructs an oracle $\mathbf{O}_{\mathbf{A}}$ that loads every rescaled matrix entry into the amplitude of the flag qubit:
\begin{equation}
    \mathbf{O}_{\mathbf{A}} \ket{0}_\text{f}\ket{i}\ket{j} = \Bigl( \bar{a}_{ij}\ket{0}_\text{f} + \sqrt{1 - \bar{a}_{ij}^2}\,\ket{1}_\text{f} \Bigr)\ket{i}\ket{j},
    \label{eq:fable_oracle}
\end{equation}
where $\ket{\cdot}_\text{f}$ is the state of the flag qubit, and
the basis states $\ket{i}$ and $\ket{j}$ belong to the ancillary
index and system registers, respectively. This oracle operation is realized by a uniformly controlled $\mathbf{R}_\text{Y}(\theta_{ij})$  rotation on the flag qubit, with the index and system registers jointly selecting the angle %The oracle is a uniformly controlled rotation $\mathbf{R}_\text{Y}(\theta_{ij})$ on the flag qubit with angles 
$\theta_{ij} = 2\arccos(\bar{a}_{ij})$, which is implemented by $4^{q}$ single-qubit rotations and $4^{q}$ CNOT gates arranged in Gray-code order. %The block encoding is then given by
To convert the entry-encoding oracle into a block encoding of \(\mathbf A_{\mathrm q}\), we combine it with two Hadamard layers and a register swap:
\begin{equation}
    \mathbf{U}_{\mathbf{A}} = \bigl(\mathbf{I}_\text{f} \otimes \mathbf{H}^{\otimes q} \otimes \mathbf{I}_N\bigr) \bigl(\mathbf{I}_\text{f} \otimes \mathbf{SWAP}\bigr) \, \mathbf{O}_{\mathbf{A}} \, \bigl(\mathbf{I}_\text{f} \otimes \mathbf{H}^{\otimes q} \otimes \mathbf{I}_N\bigr),
    \label{eq:fable}
\end{equation}
where $\mathbf{H}$ denotes the Hadamard gate and $\mathbf{SWAP}$ exchanges the index and system registers qubit by qubit. Reading the operations from right to left, the first Hadamard layer prepares a uniform superposition of row indices in the ancillary index register. The oracle encodes the corresponding matrix entries in the flag amplitudes, and the SWAP transfers the row labels to the system register, where they identify the output coordinates. The final Hadamard layer maps each index-register basis state to a superposition with amplitude \(2^{-q/2}\) on \(|0^q\rangle\).
Each Hadamard layer contributes a factor $2^{-q/2}$ to the
selected block, resulting in an overall factor of $2^{-q}$.
Combining with the entry rescaling by $c_\text{F}$ yields
the block $\mathbf{A}_\text{q}/(2^q c_\text{F})$ and hence
$\alpha=2^q c_\text{F}$. %Consequently, the complete circuit acts on $2q + 1$ qubits.

\subsubsection{Global Cost Function and Coherent Evaluation}
\label{s:cost}

The VQLS seeks parameters $\boldsymbol{w}$ such that $\mathbf{A}_\text{q}\ket{\boldsymbol{z}(\boldsymbol{w})}$ is parallel to $\ket{\boldsymbol{b}} = \boldsymbol{b}_\text{q}/\norm{\boldsymbol{b}_\text{q}}$ to account for the unit-norm constraint on quantum states. We quantify this alignment using the normalized global cost function \citep{bravo2019variational}
\begin{equation}
    C(\boldsymbol{w}) = 1 - \frac{\bigl|\bra{\boldsymbol{b}}\mathbf{A}_\text{q}\ket{\boldsymbol{z}(\boldsymbol{w})}\bigr|^2}{\norm{\mathbf{A}_\text{q}\ket{\boldsymbol{z}(\boldsymbol{w})}}^2},
    \label{eq:cost_def}
\end{equation}
%which equals one minus the squared cosine of the angle between $\mathbf{A}_\text{q}\ket{\boldsymbol{z}(\boldsymbol{w})}$ and $\ket{\boldsymbol{b}}$. 
Hence, $C(\boldsymbol{w}) \in [0, 1]$, and $C(\boldsymbol{w}) = 0$ if and only if $\ket{\boldsymbol{z}(\boldsymbol{w})} = \pm\ket{\boldsymbol{z}^\star}$, where $\ket{\boldsymbol{z}^\star} = \boldsymbol{z}^\star/\norm{\boldsymbol{z}^\star}$ is the normalized solution of the encoded system. More importantly, the cost value certifies the quality of the trial state through \citep{bravo2019variational}
\begin{equation}
    d_\text{tr}\bigl(\ket{\boldsymbol{z}(\boldsymbol{w})}, \ket{\boldsymbol{z}^\star}\bigr) \le \kappa_2(\mathbf{A}_\text{q})\sqrt{C(\boldsymbol{w})},
    \label{eq:vqls_bound}
\end{equation}
% where $d_\text{tr}(\ket{\boldsymbol{\psi}_1}, \ket{\boldsymbol{\psi}_2}) = \sqrt{1 - |\langle \boldsymbol{\psi}_1 | \boldsymbol{\psi}_2 \rangle|^2}$ is the trace distance between two pure states.
where $d_\text{tr}\bigl(\ket{\boldsymbol{z}(\boldsymbol{w})}, \ket{\boldsymbol{z}^\star}\bigr) = \sqrt{1 - |\langle \boldsymbol{z}(\boldsymbol{w}) | \boldsymbol{z}^\star \rangle|^2}$ is the trace distance between the trial state and the solution state. Eq.~\eqref{eq:vqls_bound} quantifies the role of the pre-encoding: with $\kappa_2(\mathbf{A}_\text{q}) \approx 1$, a small cost value guarantees an accurate solution state, whereas for a system matrix with a condition number of the order of $10^{10}$, the same cost value provides no meaningful guarantee.
The cost function is evaluated coherently by the circuit \citep{mari2019coherent}:
\begin{equation}
    \mathbf{U}_\text{circ}(\boldsymbol{w}) = \bigl(\mathbf{I}_\text{anc} \otimes \mathbf{U}_{\boldsymbol{b}}^\dagger\bigr) \, \mathbf{U}_{\mathbf{A}} \, \bigl(\mathbf{I}_\text{anc} \otimes \mathbf{V}(\boldsymbol{w})\bigr),
    \label{eq:circuit}
\end{equation}
where $\mathbf{I}_\text{anc}$ is the identity on the ancilla qubits and $\mathbf{U}_{\boldsymbol{b}}$ is a state preparation satisfying $\mathbf{U}_{\boldsymbol{b}}\ket{0}^{\otimes q} = \ket{\boldsymbol{b}}$, so that $\mathbf{U}_{\boldsymbol{b}}^\dagger$ maps $\ket{\boldsymbol{b}}$ to the all-zero state. After applying $\mathbf{U}_\text{circ}(\boldsymbol{w})$ to the all-zero state of the $2q+1$ qubits, we measure two projectors, i.e., $\boldsymbol{\Pi}_\text{anc} = \ket{0}\bra{0}^{\otimes n_\text{anc}} \otimes \mathbf{I}_N$ on the ancilla qubits and $\boldsymbol{\Pi}_\text{glob} = \ket{0}\bra{0}^{\otimes (n_\text{anc}+q)}$ on all qubits, to calculate the global cost (Supplementary Material~\ref{app:cost_derivation}):
\begin{equation}
    p_\text{anc}(\boldsymbol{w}) = \frac{\norm{\mathbf{A}_\text{q}\ket{\boldsymbol{z}(\boldsymbol{w})}}^2}{\alpha^2}, \qquad
    p_\text{glob}(\boldsymbol{w}) = \frac{\bigl|\bra{\boldsymbol{b}}\mathbf{A}_\text{q}\ket{\boldsymbol{z}(\boldsymbol{w})}\bigr|^2}{\alpha^2}, \qquad
    C(\boldsymbol{w}) = 1 - \frac{p_\text{glob}(\boldsymbol{w})}{p_\text{anc}(\boldsymbol{w})}.
    \label{eq:cost}
\end{equation}
%The subnormalization factor $\alpha$ cancels in the ratio, so that the rescaling by $c_\text{F}$ in FABLE does not bias the objective. 
%Here, \(p_{\text{anc}}\) denotes the probability of measuring all ancilla qubits in the state \(\ket{0}^{\otimes n_{\text{anc}}}\), which is referred to as the success probability of the block-encoding procedure. $p_\text{glob}/p_\text{anc}$ is the conditional probability that the system register is found in $\ket{\boldsymbol{b}}$ given this success. The coherent evaluation thus requires a single circuit and two read-outs. %, without a Hadamard test or an additional interference register.
Here, \(p_{\mathrm{anc}}\) is the probability of measuring all ancilla qubits as zero, which is termed a block-encoding success because, immediately after \(\mathbf U_{\mathbf A}\), the system state conditioned on this outcome is proportional to \(\mathbf A_{\mathrm q}|\boldsymbol z\rangle\). The subsequent application of \(\mathbf U_{\boldsymbol b}^{\dagger}\) leaves the ancilla probability unchanged and converts the overlap with \(|\boldsymbol b\rangle\) into an all-zero system measurement probability. Specifically, \(p_{\mathrm{glob}}/p_{\mathrm{anc}}\) is the conditional probability of obtaining the all-zero system outcome after \(\mathbf U_{\boldsymbol b}^{\dagger}\), given the all-zero ancilla outcome. This ratio equals the squared overlap between \(|\boldsymbol b\rangle\) and the normalized matrix-transformed trial state \(\mathbf A_{\mathrm q}|\boldsymbol z\rangle/\|\mathbf A_{\mathrm q}|\boldsymbol z\rangle\|_2\). Thus, coherent cost evaluation uses a single circuit configuration, with both probabilities estimated from the same measurement samples collected over repeated executions.

\subsection{S3: Decoding with Physical Prior}
\label{s:decoding}

The estimated reduced solution \(\hat{\boldsymbol z}\) is mapped to the full set of heart-surface nodes using the decoder constructed during pre-encoding:
\begin{equation}
    \hat{\boldsymbol{x}}_\text{raw} = \boldsymbol{\Phi}\hat{\boldsymbol{z}} = \mathbf{B}\mathbf{Q}\hat{\boldsymbol{z}} \in \mathbb{R}^{n}.
    \label{eq:decode}
\end{equation}
Because $\boldsymbol{\Phi}$ is an isometry, an error in the reduced coordinates produces an error of identical norm on the original data space. Note that none of the preceding steps constrains the range of $\hat{\boldsymbol{x}}_\text{raw}$, which may lead to an estimate that contains out of range values. As the normalized HSP lies in $[0, 1]$, we impose this physical prior by the Euclidean projection onto the box $[0, 1]^{n}$:
\begin{equation}
    \hat{\boldsymbol{x}} = \mathcal{P}_{[0,1]}\bigl(\hat{\boldsymbol{x}}_\text{raw}\bigr), \qquad
    \bigl[\mathcal{P}_{[0,1]}(\boldsymbol{v})\bigr]_i = \min\bigl(\max(v_i, 0), 1\bigr), \quad i = 1, \dots, n.
    \label{eq:box}
\end{equation}
Because the Euclidean projection onto a closed convex set is non-expansive \citep{boyd2004convex} and the true HSP $\boldsymbol{x}^\star$ lies in the box, $\norm{\hat{\boldsymbol{x}} - \boldsymbol{x}^\star} \le \norm{\hat{\boldsymbol{x}}_\text{raw} - \boldsymbol{x}^\star}$ holds for every frame (Proposition~\ref{prop:box}), i.e., the prior never increases the reconstruction error. 
The complete procedure for one frame is summarized in Algorithm~\ref{alg:qecg}.

\begin{algorithm}[!ht]
\caption{Inverse ECG reconstruction via a preconditioned VQLS}
\label{alg:qecg}
\begin{algorithmic}[1]
\Require
    Transfer matrix $\mathbf{R}$, measured BSP $\tilde{\boldsymbol{y}}$, surface Laplacian $\mathbf{D}$,
    number of system qubits $q$ ($N = 2^{q}$), block-size bounds of the partition, tail rank $r$,
    certified error level $\tau$ (noise-free) or noise norm $\norm{\boldsymbol{\xi}}$ (noisy),
    number of restarts, and iteration budget of the optimizer.
\Ensure
    Reconstructed HSP $\hat{\boldsymbol{x}} \in [0,1]^{n}$

\State \textbf{S1: Pre-encoding}
\State Select $\lambda^\star$ by Eq.~\eqref{eq:lambda_cert} (noise-free) or Eq.~\eqref{eq:lambda_morozov} (noisy), solving Eq.~\eqref{eq:tikhonov} (Supplementary Material~\ref{app:lambda_search}); set $\boldsymbol{x}_\text{c} \gets \boldsymbol{x}_\text{c}(\lambda^\star)$ \Comment{Coarse estimate}
\State Score the nodes by Eq.~\eqref{eq:box_score} and select the support $\mathcal{S}$ of the $s = N - 1$ largest scores
\State Form $\boldsymbol{c}_\text{o}$ and $\mathbf{B} \gets [\,\mathbf{E}_{\mathcal{S}} \;\; \boldsymbol{c}_\text{o}\,]$ by Eq.~\eqref{eq:basis} \Comment{Box-aware subspace}
\State $\mathbf{G} \gets (\mathbf{R}\mathbf{B})^\top(\mathbf{R}\mathbf{B})$, \; $\boldsymbol{g} \gets (\mathbf{R}\mathbf{B})^\top\tilde{\boldsymbol{y}}$ by Eq.~\eqref{eq:compress} \Comment{Compression}
\State Compute the Lanczos basis $\mathbf{Q}$ of $\mathcal{K}_N(\mathbf{G}, \boldsymbol{g})$; \; $\mathbf{T} \gets \mathbf{Q}^\top\mathbf{G}\mathbf{Q}$, \; $\boldsymbol{h} \gets \mathbf{Q}^\top\boldsymbol{g}$ by Eq.~\eqref{eq:krylov} \Comment{Krylov coordinates}
\State Select the block partition (Supplementary Material~\ref{app:partition}); \; $\mathbf{A}_0 \gets \mathbf{P}^{-1}\mathbf{T}$, \; $\boldsymbol{b}_0 \gets \mathbf{P}^{-1}\boldsymbol{h}$ by Eq.~\eqref{eq:precond} \Comment{Block-Jacobi left preconditioning}
\State Compute the SVD $\mathbf{A}_0 = \mathbf{U}_\text{L}\boldsymbol{\Gamma}\mathbf{U}_\text{R}^\top$ and form $\mathbf{F}_\text{tail}$ by Eq.~\eqref{def-J} \Comment{Tail correction}
\State $\mathbf{A}_\text{q} \gets \mathbf{F}_\text{tail}\mathbf{A}_0$, \; $\boldsymbol{b}_\text{q} \gets \mathbf{F}_\text{tail}\boldsymbol{b}_0$, \; $\boldsymbol{\Phi} \gets \mathbf{B}\mathbf{Q}$ by Eq.~\eqref{eq:encoded}

\State \textbf{S2: Coherent VQLS}
\State Construct $\mathbf{U}_{\mathbf{A}}$ by Eq.~\eqref{eq:fable} and the state preparation $\mathbf{U}_{\boldsymbol{b}}$ of $\ket{\boldsymbol{b}} = \boldsymbol{b}_\text{q}/\norm{\boldsymbol{b}_\text{q}}$
\For{each restart}
    \State Draw a random unit-norm initial parameter vector $\boldsymbol{w}^{(0)} \in \mathbb{R}^{N}$
    \State Minimize $C(\boldsymbol{w})$ by L-BFGS-B from $\boldsymbol{w}^{(0)}$, with $p_\text{anc}$ and $p_\text{glob}$ read out from the circuit in Eq.~\eqref{eq:circuit} and combined by Eq.~\eqref{eq:cost}
\EndFor
\State Let $\hat{\boldsymbol{w}}$ be the parameters of the restart with the lowest final cost; \; $\hat{\boldsymbol{\psi}} \gets \ket{\boldsymbol{z}(\hat{\boldsymbol{w}})} = \hat{\boldsymbol{w}}/\norm{\hat{\boldsymbol{w}}}$ by Eq.~\eqref{eq:ansatz}

\State \textbf{S3: Decoding}
\State $\nu^\star \gets \langle \mathbf{A}_\text{q}\hat{\boldsymbol{\psi}}, \boldsymbol{b}_\text{q} \rangle / \norm{\mathbf{A}_\text{q}\hat{\boldsymbol{\psi}}}^2$, \; $\hat{\boldsymbol{z}} \gets \nu^\star\hat{\boldsymbol{\psi}}$ \Comment{Scale restoration}
\State $\hat{\boldsymbol{x}} \gets \mathcal{P}_{[0,1]}(\boldsymbol{\Phi}\hat{\boldsymbol{z}})$ by Eqs.~\eqref{eq:decode} and \eqref{eq:box}
\State \Return $\hat{\boldsymbol{x}}$
\end{algorithmic}
\end{algorithm}

  \section{Experimental Design and Results} \label{s:results}

% Values not yet measured are typeset in red by \pending{} and are to be replaced
% by the circuit-trained Ablation 1 run.  Delete this macro only after every
% \pending{} has been replaced.
% \providecommand{\pending}[1]{\textcolor{red}{#1}}
% \providecommand{\ket}[1]{\left\lvert #1 \right\rangle}

\subsection{Experimental Design} \label{s:design}

We evaluate the proposed framework on reconstructing HSPs from BSPs in a realistic heart--torso geometry. The ventricular surface is discretized into $n = 256$ nodes and the torso surface into $m = 352$ electrodes, sourced from the geometry dataset provided in the 2007 PhysioNet Computing in Cardiology Challenge \cite{goldberger2000physiobank}. The cardiac electrical activity is simulated on the ventricular mesh by the Aliev--Panfilov reaction--diffusion model \citep{aliev1996simple, zhang2026geometry, zhang2026pkgcn} with parameters $a_\text{AP} = 0.1$, $c_\text{AP} = 8$, $e_{0,\text{AP}} = 0.002$, $\mu_{1,\text{AP}} = \mu_{2,\text{AP}} = 0.3$, diffusion coefficient $\delta_\text{AP} = 10$, and time step $\Delta t = 0.1$, yielding 681 frames of the normalized transmembrane potential $\boldsymbol{x}(t) \in [0,1]^{256}$. The BSPs are generated by a torso-heart transfer matrix $\mathbf{R} \in \mathbb{R}^{352 \times 256}$ as $\boldsymbol{y}(t) = \mathbf{R}\boldsymbol{x}(t)$. $\mathbf{R}$ is severely ill-conditioned with $\kappa_2(\mathbf{R}) = 1.01 \times 10^{5}$, which is the defining difficulty of the inverse ECG problem \citep{yao2021spatiotemporal}.

As measurement noise is inevitable in clinical recordings, we corrupt the BSPs as $\tilde{\boldsymbol{y}}(t) = \boldsymbol{y}(t) + \boldsymbol{\xi}(t)$, where $\boldsymbol{\xi}(t) \sim \mathcal{N}(\boldsymbol{0}, \sigma_{\boldsymbol{\xi}}^2 \sigma^2 \mathbf{I}_m)$, $\sigma_{\boldsymbol{\xi}} \in \{0, 0.01, 0.02\}$ is the dimensionless noise level, and $\sigma$ is the reference standard deviation of the BSP recording.
Note that the physical system cannot be handed to a variational quantum linear solver directly. Encoding the 256-node HSP requires $\log_2 256 = 8$ system qubits, and the regularized normal equations inherit a condition number of order $\kappa_2(\mathbf{R})^2 \approx 10^{10}$, whereas the accuracy that a given VQLS cost can guarantee deteriorates in proportion to the condition number \citep{bravo2019variational}. Therefore, we implement the proposed pre-encoding and compress each frame into a $64 \times 64$ system $\mathbf{A}_\text{q} \boldsymbol{z} = \boldsymbol{b}_\text{q}$ with $\kappa_2(\mathbf{A}_\text{q}) = 1.00$, which is solved with $q = 6$ system qubits and decoded back to the full 256 heart nodes.

\begin{figure}[!ht]
	\begin{center}
		\includegraphics[width=1\textwidth]{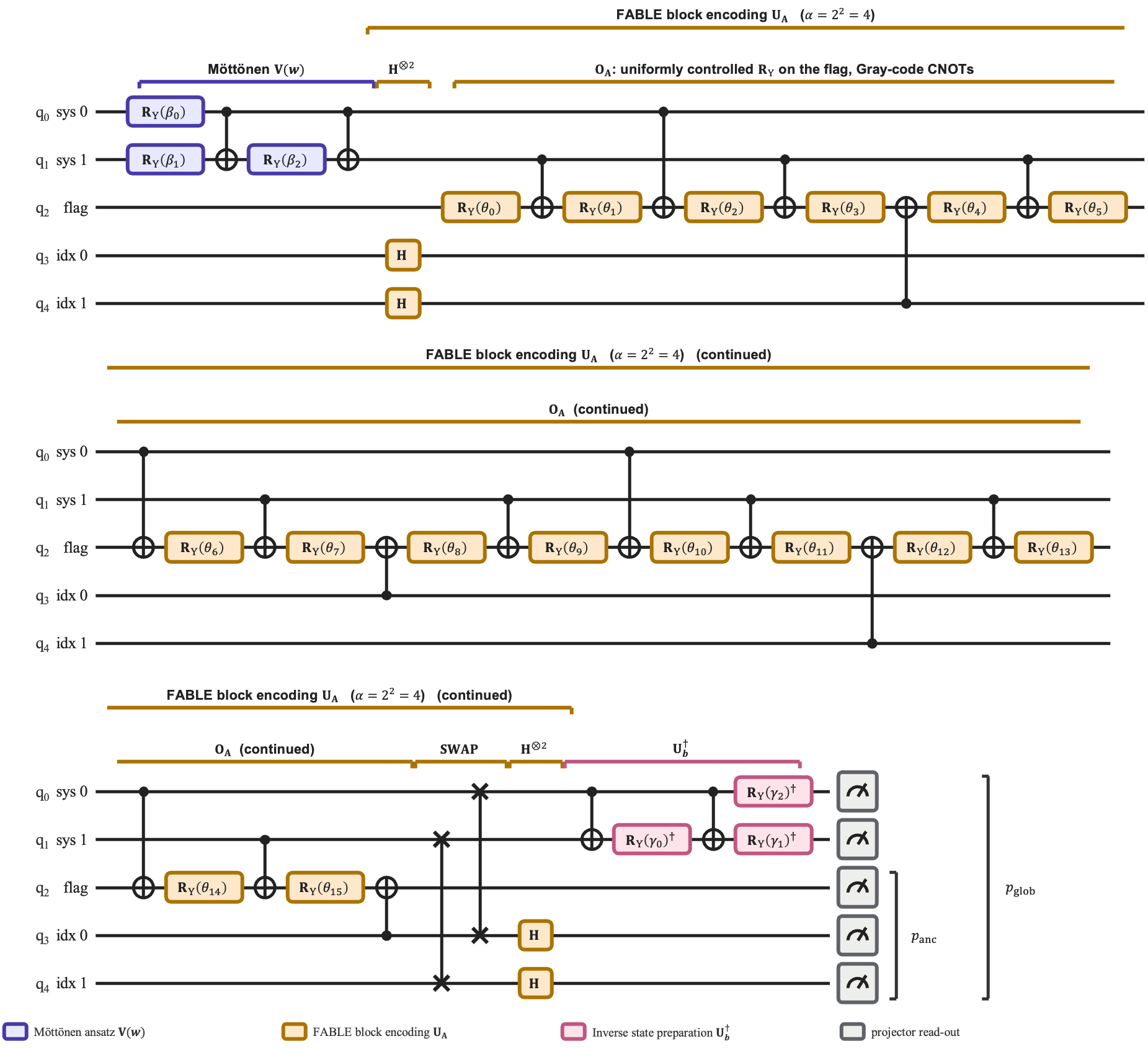}
    \caption{Gate-level expansion of the coherent VQLS circuit, shown for $q = 2$ on an illustrative $4 \times 4$ system. Rotation angles are data dependent and labeled symbolically, and the number of $\mathbf{R}_\text{Z}$ gates depends on the signs of the amplitudes. The two global-phase gates, one from $\mathbf{V}(\boldsymbol{w})$ and one from $\mathbf{U}_{\boldsymbol{b}}^\dagger$, are omitted because they do not affect either projector probability. All three models share this structure and differ only in $q$: $q = 6$ for the complete model and Ablation 2, and $q = 8$ for Ablation 1.}
		\label{Fig:circuit_gate}
	\end{center}
\end{figure}

We implement the coherent VQLS described in Section~\ref{s:vqls} in PennyLane \citep{bergholm2018pennylane}. % and simulate all circuits with its state-vector backend, \texttt{lightning.\allowbreak qubit}, using adjoint differentiation. 
For the complete model, the circuit in Fig.~\ref{Fig:circuit_register} acts on $2q + 1 = 13$ qubits, and the rescaling factor in Eq.~\eqref{eq:fable_oracle} is $c_\text{F} = 1.00$, so that the FABLE subnormalization factor is $\alpha = 2^{q} = 64$. Fig.~\ref{Fig:circuit_gate} expands the same circuit to individual gates for $q = 2$. The structure is identical at $q = 6$, where $\mathbf{O}_{\mathbf{A}}$ contains $4^6 = 4{,}096$ Gray-code steps. The parameters are optimized by L-BFGS-B \citep{byrd1995limited} with three random restarts, a budget of 500 iterations, and a threshold of $10^{-10}$ on the cost function in Eq.~\eqref{eq:cost}. 
% the restart with the lowest final cost is retained, and its physical scale is restored by Eq.~\eqref{eq:scale}. 
The remaining settings of the framework are summarized in Table~\ref{Tab:config} of Supplementary Material~\ref{app:config}, and Table~\ref{Tab:resources} lists the resulting quantum resources for each model.

\begin{table}[!ht]
	\centering
	\caption{Quantum resources of the coherent VQLS circuit for each model. %{\color{red}FABLE gate counts correspond to an uncompressed encoding (tolerance $0$). \textbf{(What do you mean by this? -> In FABLE, a tolerance hyperparameter prunes values with magnitude below the threshold. In our experiments, we set tolerance = 0 (no compression). We can omit this detail if it is confusing. )}}
    }
	\label{Tab:resources}
	\small
	\begin{tabular}{@{}lccc@{}}
		\toprule
		Quantity & Complete model & Ablation 1 & Ablation 2 \\
		\midrule
		System qubits $q$ & 6 & 8 & 6 \\
		Ancilla qubits (1 flag $+$ $q$ index) & 7 & 9 & 7 \\
		Total qubits $2q + 1$ & 13 & 17 & 13 \\
		Encoded dimension $N = 2^q$ & 64 & 256 & 64 \\
		M\"{o}tt\"{o}nen parameters & 64 & 256 & 64 \\
		FABLE subnormalization $\alpha = 2^q$ & 64 & 256 & 64 \\
		FABLE $\mathbf{R}_\text{Y}$ rotations & 4,096 & 65,536 & 4,096 \\
		FABLE CNOT gates & 4,096 & 65,536 & 4,096 \\
		FABLE Hadamard\,/\,SWAP gates & 12\,/\,6 & 16\,/\,8 & 12\,/\,6 \\
		FABLE gates per cost evaluation & 8,210 & 131,096 & 8,210 \\
		Condition number $\kappa_2(\mathbf{A})$ & 1.00 & $1.03 \times 10^{10}$ & 1.00 \\
		\bottomrule
	\end{tabular}
\end{table}

We benchmark the complete model against two ablations that each remove one component: \textbf{(i) Ablation 1 (without pre-encoding)}: We omit pre-encoding and apply the solver directly to the original normal equations, with \(\mathbf{A}=\mathbf{R}^{\top}\mathbf{R}\in\mathbb{R}^{256\times256}\) and \(\boldsymbol{b}=\mathbf{R}^{\top}\tilde{\boldsymbol{y}}\). This system requires \(q=8\) system qubits and has a condition number of \(\kappa_2(\mathbf{A})=1.03\times10^{10}\). \textbf{(ii) Ablation 2 (identity regularizer)}: The pre-encoding is retained, but the Laplacian $\mathbf{D}$ in the coarse estimation of Eq.~\eqref{eq:tikhonov} is replaced by an $n \times n$ identity matrix, $\mathbf{I}_{n}$. The resulting penalty controls the magnitude of the coarse estimate rather than its spatial roughness.
%  {\color{blue} To make the regularizer the only varying factor, Ablation 2 reuses, frame by frame, the regularization weight selected by the complete model.}

%For every model, frame, and noise level, the system handed to the VQLS is also solved by a classical linear solver (CLS, \texttt{numpy.\allowbreak linalg.\allowbreak solve}), and both solutions are decoded and projected through identical code paths. This matched protocol separates the two sources of error: the gap between the CLS solution and the reference measures the cost of the model, whereas the gap between the CLS and the VQLS solutions measures the cost of the quantum solver. 
For each model, time frame, and noise level, we solve the same linear system using both the VQLS and a classical linear solver (CLS, \texttt{numpy.linalg.solve}), then apply identical decoding procedure to both solutions. This matched comparison distinguishes reconstruction error present in the classical baseline from discrepancies introduced by the VQLS. We employ the relative error ($\text{RE}$) to assess reconstruction accuracy, and the fidelity ($F$) between the two solvers to assess how faithfully the VQLS reproduces the classical solution:
\begin{equation}
	\text{RE} = \frac{\norm{\hat{\boldsymbol{x}} - \boldsymbol{x}}}{\norm{\boldsymbol{x}}},
	\qquad
	F = \frac{\abs{\langle \hat{\boldsymbol{x}}_\text{CLS}, \hat{\boldsymbol{x}}_\text{VQLS} \rangle}^2}{\norm{\hat{\boldsymbol{x}}_\text{CLS}}^2 \, \norm{\hat{\boldsymbol{x}}_\text{VQLS}}^2},
	\label{eq:metrics}
\end{equation}
where $\boldsymbol{x}$ and $\hat{\boldsymbol{x}}$ denote the reference and reconstructed HSPs, respectively. %The RE is computed on the physical vector without removing sign or scale, while $F$ is scale invariant and reports agreement in shape only. 
% {\color{blue} We additionally record the number of L-BFGS-B iterations and the wall-clock time of each pipeline stage, which characterize the optimization effort and the computational cost of the framework.}

\subsection{Experimental Results} \label{s:results_analysis}

\begin{figure}[!ht]
	\begin{center}
		\includegraphics[width=1\textwidth]{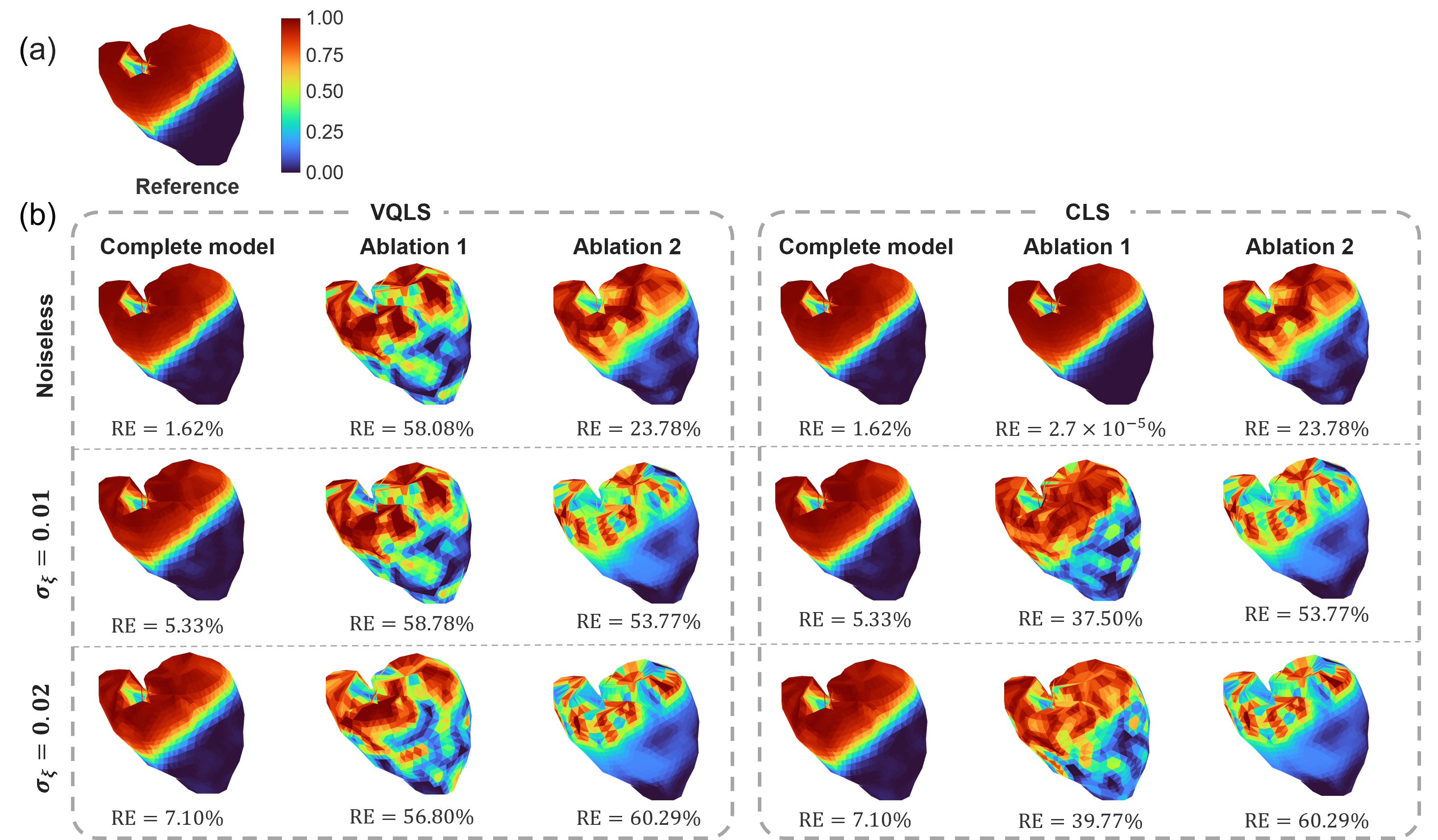}
		\caption{(a) Reference mapping of the heart-surface electrodynamics at frame $t = 475$. (b) Reconstructed mappings by the VQLS and CLS under varying noise conditions %($\sigma_{\boldsymbol{\xi}} = 0$, $0.01$, $0.02$) 
        and experimental settings %(complete model, Ablation 1 without pre-encoding, and Ablation 2 with the identity regularizer) 
        at frame $t = 475$.}
		\label{Fig:hsp_map}
	\end{center}
\end{figure}

Fig.~\ref{Fig:hsp_map} presents a visual comparison of the reconstructed HSP at frame $t = 475$ across the three experimental settings and three noise levels, with panel (a) showing the reference mapping and panel (b) the mappings recovered by the VQLS and CLS. 
From the visual comparison, it is evident that the complete VQLS model stays closest to the reference at every noise level, preserving both the position and smoothness of the wavefront, with $\text{RE} = 1.62\%$, $5.33\%$, and $7.10\%$, respectively, at $\sigma_{\boldsymbol{\xi}} = 0$, $0.01$, and $0.02$. Ablation 1 of VQLS fails to recover the front altogether: its mapping is fragmented into patches of alternating potential, and its error stays between $56.80\%$ and $58.78\%$ across the three noise levels, consistent with the severe ill-conditioning of the original normal equations without pre-encoding. The VQLS reconstruction in Ablation 2 captures the global structure of the front on noise-free data but mottles both plateaus with spurious extrema ($\text{RE} = 23.78\%$). Its accuracy deteriorates most sharply as noise increases, filling the recovered region with spurious mid-range potentials at $\sigma_{\boldsymbol{\xi}} = 0.01$ and $0.02$ ($53.77\%$ and $60.29\%$). These differences highlight the influence of coarse-estimate regularization on the subspace used for reconstruction. 
%As noted in Section~\ref{s:subspace}, the regularizer enters the reconstruction through the basis $\mathbf{B}$ in Eq.~\eqref{eq:basis}, i.e., through the active support $\mathcal{S}$ and the off-support profile $\boldsymbol{c}_\text{o}$.  
The mesh-Laplacian penalty yields a spatially smooth coarse estimate and a smooth off-support profile. %whereas the identity penalty controls only the magnitude of the coarse estimate and leaves its spatial irregularities, which are further aggravated by measurement noise, in the off-support profile and thus in the final reconstruction.
In contrast, the identity penalty controls the magnitude of the coarse estimate without explicitly penalizing spatial irregularities. These irregularities can become more pronounced under measurement noise and persist in the off-support profile, through which they can propagate into the final reconstruction.
%In addition, spurious extrema of the identity-regularized coarse estimate can receive large scores in Eq.~\eqref{eq:box_score} and displace nodes of the active region from $\mathcal{S}$. Because the final reconstruction is the least-squares solution restricted to $\operatorname{range}(\mathbf{B})$, a poorly selected subspace cannot be corrected by the subsequent solve, which explains the large difference between the two regularization models.

The two halves of panel (b) separate the contribution of the quantum solver from the error inherent in the system being solved, which is determined by the pre-encoding and the regularizer and is shared by the CLS solution of the same system. For the complete model and Ablation 2, the VQLS and the CLS mappings are visually indistinguishable and their REs agree to the reported precision at every noise level, so the residual error of these models is attributable to the encoding and the regularizer rather than to the variational solve. Ablation 1 behaves in the opposite way: on noise-free data the CLS recovers the reference almost exactly ($\text{RE} = 2.7 \times 10^{-5}\%$), despite the severe ill-conditioning of the normal equations. In this case, the reference satisfies the linear system up to numerical rounding, allowing the classical solver to recover it accurately. A large condition number indicates sensitivity to perturbations but does not necessarily lead to a large reconstruction error in their absence. The VQLS introduces additional approximation through variational optimization and circuit approximation or measurement sampling, and thus the VQLS solution without pre-encoding (Ablation 1) of the identical system remains fragmented at $58.08\%$. %{\color{orange}This failure originates in the conditioning rather than in the inverse problem, since the accuracy certified by the VQLS cost $C$ scales with $\kappa_2(\mathbf{A})\sqrt{C}$ \citep{bravo2019variational}: for Ablation 1, $\kappa_2(\mathbf{A}) = 1.03 \times 10^{10}$ and $C \approx 6.5 \times 10^{-8}$ give a bound of order $10^{6}$, so that a small cost carries no information about the solution, {\color{red}whereas $\kappa_2(\mathbf{A}_\text{q}) = 1.00$ and $C = 4.1 \times 10^{-9}$ bound the deviation of the complete model by $6.4 \times 10^{-5}$. \textbf{(?)}} The same ill-conditioning destabilizes the classical solution once noise is present, {\color{red}amplifying the perturbation to the scale of the signal itself \textbf{(?)}}, so that the CLS mappings of Ablation 1 break down to $\text{RE} = 37.50\%$ and $39.77\%$ at $\sigma_{\boldsymbol{\xi}} = 0.01$ and $0.02$. The per-frame errors in Fig.~\ref{Fig:RE} and the solver infidelity in Fig.~\ref{Fig:gap} quantify these observations over all frames.}
%This failure originates in the conditioning rather than in the inverse problem, because the accuracy certified by the VQLS cost $C$ scales with $\kappa_2(\mathbf{A})\sqrt{C}$ \citep{bravo2019variational}. For Ablation 1, $\kappa_2(\mathbf{A}) = 1.03 \times 10^{10}$ and $C \approx 6.5 \times 10^{-8}$ yield a bound of order $10^{6}$ on a trace distance that cannot exceed one, so that a small cost carries no information about the solution. In contrast, for the complete model, $\kappa_2(\mathbf{A}_\text{q}) = 1.00$ and $C = 4.1 \times 10^{-9}$, so that Eq.~\eqref{eq:vqls_bound} bounds the trace distance between the VQLS state and the exact solution state by $\kappa_2(\mathbf{A}_\text{q})\sqrt{C} = 6.4 \times 10^{-5}$, i.e., the small cost certifies an accurate solution. 
The same ill-conditioning destabilizes the classical solution once noise is present: %By Eq.~\eqref{eq:cond_bound}, the solution of the unregularized normal equations can amplify a relative perturbation of the BSPs by a factor of up to $\kappa_2(\mathbf{R}) = 1.01 \times 10^{5}$, so that 
the CLS mappings of Ablation 1 break down to $\text{RE} = 37.50\%$ and $39.77\%$ at $\sigma_{\boldsymbol{\xi}} = 0.01$ and $0.02$. %The per-frame errors in Fig.~\ref{Fig:RE} and the solver infidelity in Fig.~\ref{Fig:gap} quantify these observations over all frames.

\begin{figure}[!ht]
	\centering
	\begin{subfigure}{\textwidth}
		\centering
		\includegraphics[width=\textwidth]{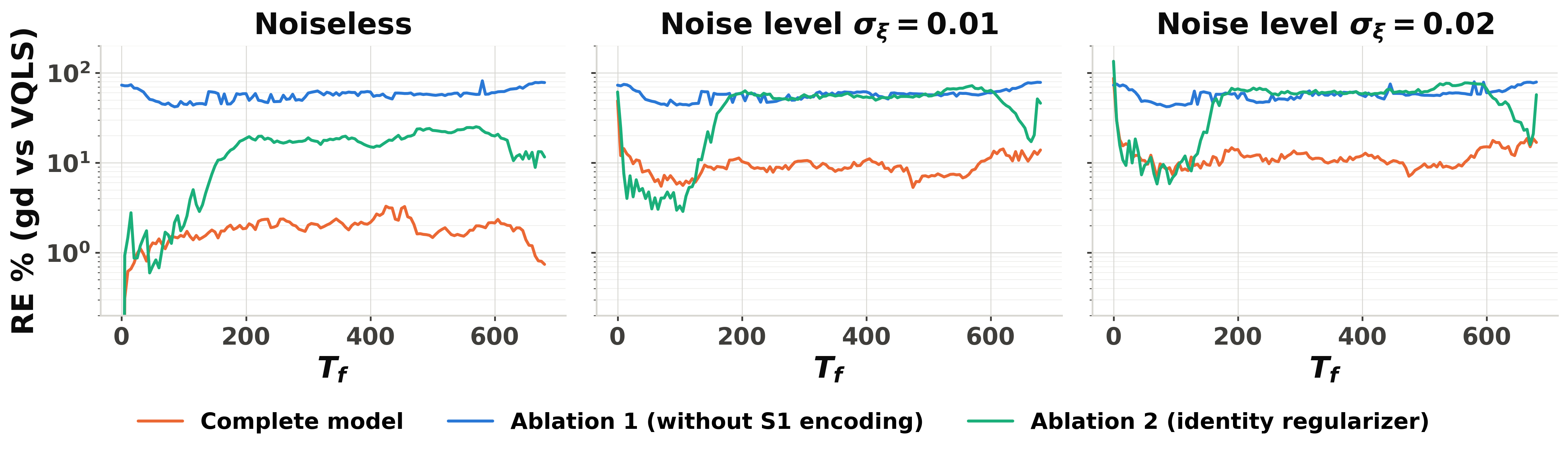}
		\caption{Groundtruth (gd) vs.\ VQLS}
	\end{subfigure}
	\vspace{0.6em}

	\begin{subfigure}{\textwidth}
		\centering
		\includegraphics[width=\textwidth]{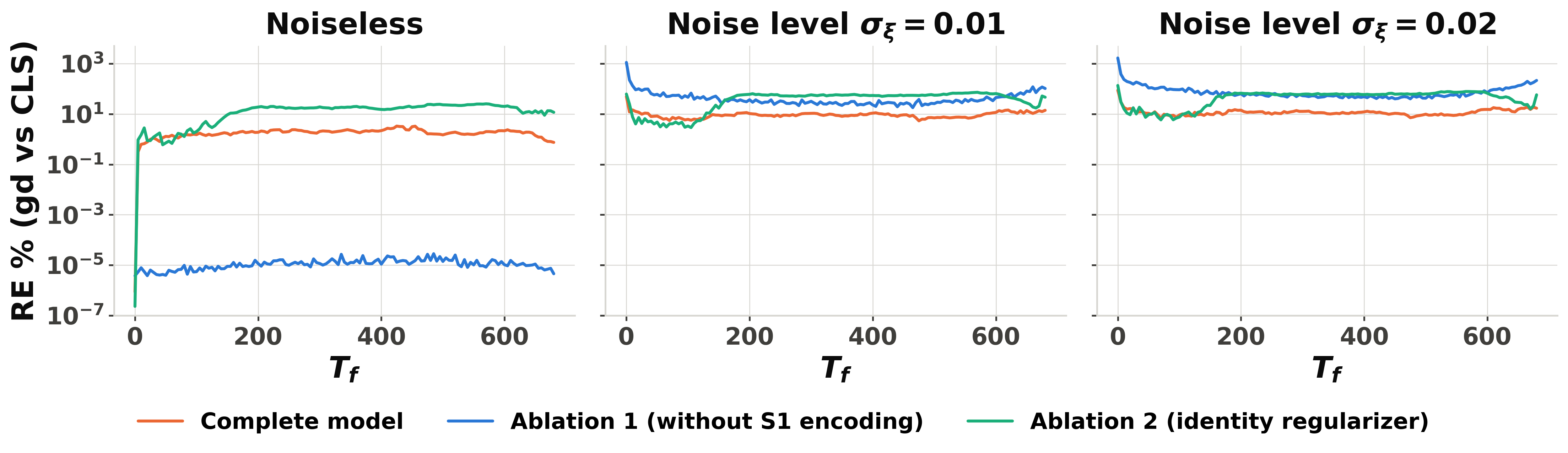}
		\caption{Groundtruth (gd) vs.\ CLS}
	\end{subfigure}
	\caption{Per-frame RE (log scale) of the complete model, Ablation 1, and Ablation 2 under noise levels $\sigma_{\boldsymbol{\xi}} = 0, 0.01, 0.02$, for (a) the VQLS and (b) CLS solutions.}
	\label{Fig:RE}
\end{figure}

Fig.~\ref{Fig:RE} extends the comparison to all frames, with the VQLS solutions in panel (a) and the CLS solutions in panel (b). The complete model achieves the lowest error at every noise level in the VQLS panel (a). Under noise-free conditions, it reconstructs the HSP with $\text{RE} = 1.81 \pm 0.54\%$, compared with $57.64 \pm 8.35\%$ for Ablation 1 and $14.60 \pm 7.40\%$ for Ablation 2, yielding error reductions of $96.9\%$ and $87.6\%$, respectively. The advantage is preserved as the measurement noise increases: the complete model yields $\text{RE} = 9.43 \pm 3.95\%$ against $57.72 \pm 7.84\%$ (Ablation 1) and $43.49 \pm 21.59\%$ (Ablation 2) when $\sigma_{\boldsymbol{\xi}} = 0.01$, and it yields $12.20 \pm 7.06\%$ against $57.94 \pm 8.63\%$ and $49.53 \pm 23.71\%$ when $\sigma_{\boldsymbol{\xi}} = 0.02$, corresponding to reductions of $78.9\%$ and $75.4\%$ at the highest noise level.

% {\color{blue} These statistics reproduce, over the whole beat, what the mappings show at a single frame. The error of Ablation 2 triples once noise is introduced ($14.60\% \rightarrow 43.49\%$), which is the frame-averaged counterpart of the mottling caused by the identity penalty, while that of the complete model remains an order of magnitude lower; for both models panels (a) and (b) are indistinguishable, so their error is the error of the model itself and not of the quantum solver. Ablation 1 again separates the two panels: averaged over frames, the CLS solves the raw normal equations almost exactly on noise-free data ($\text{RE} = 1.17 \times 10^{-5}\%$), whereas the VQLS stays near $58\%$ at every noise level, and the CLS error itself rises to $50.33 \pm 94.27\%$ at $\sigma_{\boldsymbol{\xi}} = 0.01$ and $89.47 \pm 142.37\%$ at $\sigma_{\boldsymbol{\xi}} = 0.02$, with isolated frames exceeding $10^{3}\%$.}

%{\color{orange}Fig.~\ref{Fig:gap} {\color{red}measures this separation of error sources directly \textbf{)(?->Revised)}}. For the complete model and Ablation 2 the infidelity lies between $1.12 \times 10^{-16}$ and $1.67 \times 10^{-16}$ at every noise level, at or below the double-precision limit of $2.2 \times 10^{-16}$: the VQLS reproduces the classical solution of the encoded system exactly. Ablation 1 instead exhibits a genuine solver gap that widens with noise, from $0.331 \pm 0.096$ to $0.493 \pm 0.142$, so that without the pre-encoding the quantum solution departs from the classical one by nearly half of its fidelity. Therefore, the pre-encoding is what renders the inverse ECG problem solvable by a variational quantum solver. {\color{red}\textbf{(The writing still needs polishing->Revised)}}}

\begin{figure}[!ht]
	\begin{center}
		\includegraphics[width=1\textwidth]{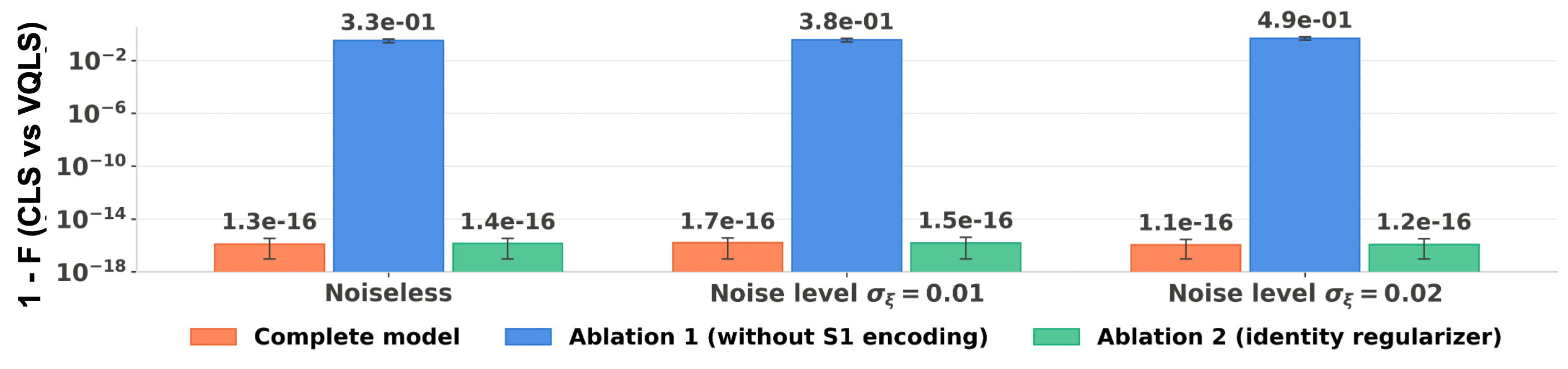}
		\caption{
        Infidelity $1 - F$ (log scale) between the CLS and VQLS reconstructions of the complete model, Ablation 1, and Ablation 2 under noise levels $\sigma_{\boldsymbol{\xi}} = 0, 0.01, 0.02$. %Bars denote the mean over all evaluated frames and error bars one standard deviation.}
    % Values near $10^{-16}$ lie at the double-precision limit.
    }
		\label{Fig:gap}
	\end{center}
\end{figure}

Fig.~\ref{Fig:gap} quantifies the discrepancy between the VQLS and CLS reconstructions through the infidelity $1 - F$ (see Eq.~\eqref{eq:metrics}), which isolates the error introduced by the quantum solver from the error inherent in the system being solved. For the complete model and Ablation 2, the mean infidelity lies between $1.12 \times 10^{-16}$ and $1.67 \times 10^{-16}$ at every noise level, which is below the double-precision limit of $2.2 \times 10^{-16}$. As such, the VQLS reproduces the classical solution of the encoded system to machine precision. In contrast, Ablation 1 exhibits a substantial solver gap that widens with the noise level, with the infidelity increasing from $0.331 \pm 0.096$ on noise-free data to $0.493 \pm 0.142$ at $\sigma_{\boldsymbol{\xi}} = 0.02$, i.e., the fidelity between the VQLS and CLS reconstructions drops to approximately $0.51$ at the highest noise level. These results demonstrate that the pre-encoding is essential for the VQLS to solve the inverse ECG system accurately.

%{\color{orange}Fig.~\ref{Fig:effort} reports the number of L-BFGS-B iterations per frame. For the complete model and Ablation 2, the optimizer converges within a median of six iterations at every noise level, and the three restarts reach the final cost to within $4 \times 10^{-15}$. This rapid and restart-independent convergence follows directly from $\kappa_2(\mathbf{A}_\text{q}) = 1.00$, which renders the cost landscape nearly isotropic for a quasi-Newton method. 
% Both models converge to an identical cost of $4.10 \times 10^{-9}$ across all frames and noise levels, yet their errors differ by a factor of eight in the noise-free case ($1.81\%$ against $14.60\%$): once the system is well conditioned, the bottleneck of reconstruction accuracy is no longer the quantum optimization but the choice of spatial regularizer. However, for Ablation 1 without pre-encoding, the optimizer requires a median of 500 iterations. \textbf{(Writing needs improvement...)}}
% and meets the convergence criterion on 43\% of the restarts

\begin{figure}[!ht]
	\begin{center}
		\includegraphics[width=\textwidth]{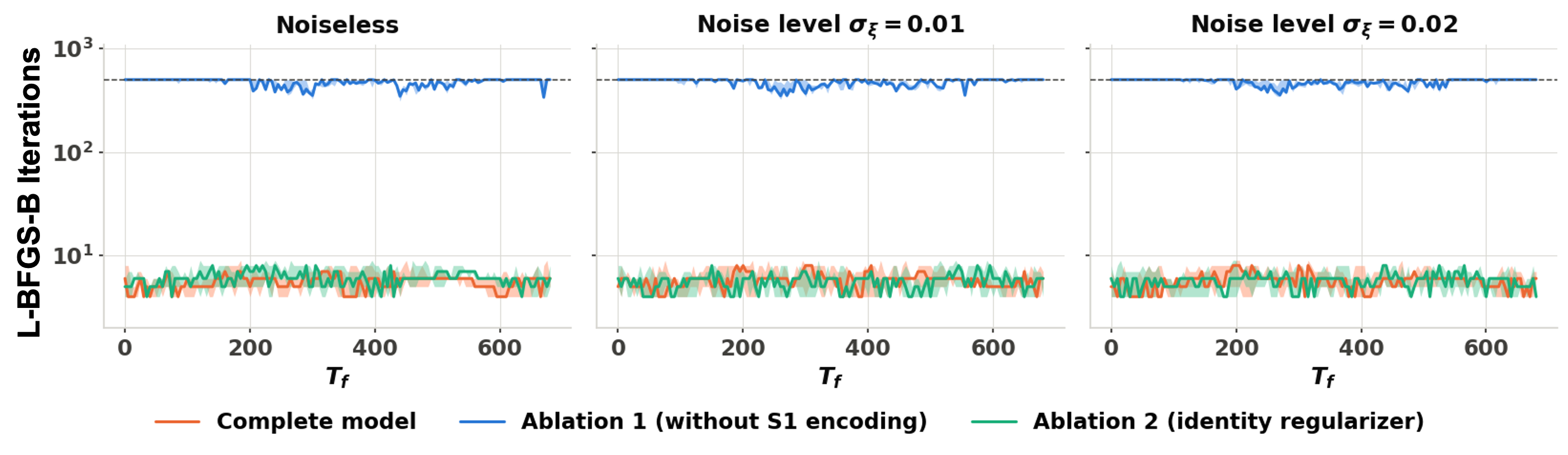}
		\caption{Number of L-BFGS-B iterations per frame. Lines denote the median and shaded bands the range over three random restarts. The dashed line marks the 500-iteration budget.}
		\label{Fig:effort}
	\end{center}
\end{figure}

Fig.~\ref{Fig:effort} reports the optimization effort of the VQLS in terms of the number of L-BFGS-B iterations per frame. For the complete model and Ablation 2, the optimizer converges within a median of six iterations at every noise level. This rapid convergence is a direct consequence of the pre-encoding. 
% With $\kappa_2(\mathbf{A}_\text{q}) = 1.00$, the singular values of $\mathbf{A}_\text{q}$ are nearly identical, so that the cost landscape is well shaped (requirement (ii) in Section~\ref{s:s1}) and the quasi-Newton iteration requires only a few steps to reach the minimum. 
In contrast, Ablation 1 reaches the 500-iteration budget on the majority of the frames. Combined with the solver gap in Fig.~\ref{Fig:gap}, this indicates that the ill-conditioning of the raw normal equations prevents the optimizer from attaining an accurate solution even at a substantially higher computational effort.

%{\color{orange}Fig.~\ref{Fig:time} decomposes the wall-clock time per frame into the four stages of the pipeline. For the complete model, the noise-free frame requires $0.035$\,s for frame preparation, $2.10$\,s to construct the pre-encoding, $51.0$\,s for the VQLS solve, and $1.8 \times 10^{-5}$\,s to decode the solution to the 256 heart nodes. 
% Ablation 2 requires $4.15$\,s and $55.4$\,s for the pre-encoding and the solve, respectively. Therefore, the quantum solve accounts for $92$--$96\%$ of the pipeline time, while the pre-encoding costs only a few seconds. This small encoding overhead yields a large reduction in quantum cost: removing two system qubits and two ancilla qubits reduces the simulation time of a single cost evaluation from $38.5$\,s to $1.09$\,s in our implementation, and it lowers the subnormalization from $256$ to $64$ (Table~\ref{Tab:resources}). Accordingly, the VQLS solve of Ablation 1 requires approximately 23\,h per frame, roughly three orders of magnitude longer than that of the complete model. {\color{red}\textbf{(Same here...)}}}

\begin{figure}[!ht]
	\begin{center}
		\includegraphics[width=\textwidth]{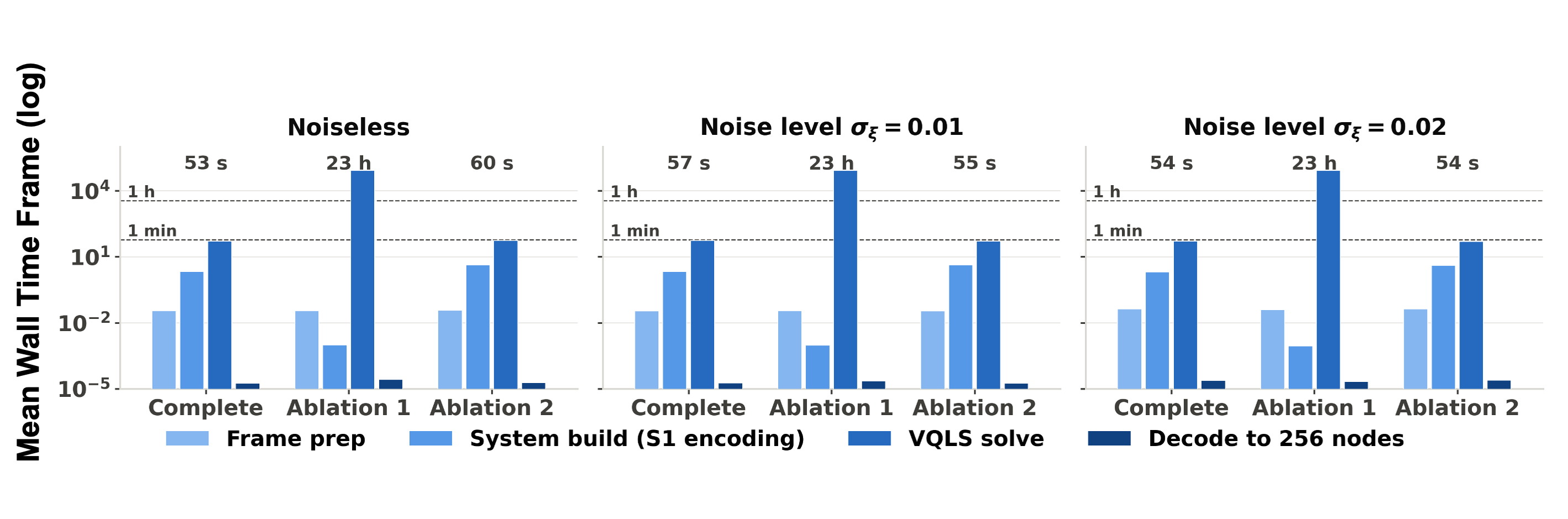}
		\caption{Mean wall-clock time per frame for each pipeline stage (log scale) over all frames.}
		\label{Fig:time}
	\end{center}
\end{figure}

Fig.~\ref{Fig:time} decomposes the mean wall-clock time per frame into the four stages of the pipeline, i.e., frame preparation, construction of the pre-encoded system (S1), the VQLS solve (S2), and decoding to the full heart nodes (S3). For the complete model on noise-free data, these stages require $0.035$\,s, $2.10$\,s, $51.0$\,s, and $1.8 \times 10^{-5}$\,s, respectively. The VQLS solve dominates the runtime, accounting for $92$--$96\%$ of the pipeline time, whereas the pre-encoding adds only a few seconds. Note that this small classical overhead substantially reduces computational cost of the subsequent VQLS solver. Specifically, compared with Ablation 1, the pre-encoding reduces the circuit size by two system qubits and two ancilla qubits, which reduces the simulation time of a single cost evaluation from $38.5$\,s to $1.09$\,s in our implementation and lowers the FABLE subnormalization factor from $256$ to $64$ (Table~\ref{Tab:resources}). Together with the larger number of optimizer iterations in Fig.~\ref{Fig:effort}, the VQLS solve of Ablation 1 requires approximately $23$\,h per frame, which is roughly three orders of magnitude longer than that of the complete model.

  \section{Conclusions} \label{s:conclusions}

This paper presents a hybrid quantum--classical framework for inverse ECG reconstruction, which, to the best of our knowledge, is the first to apply a quantum linear solver to ECG imaging.
By integrating Laplacian-regularized pre-encoding and a coherent VQLS, our framework addresses the fundamental limitations of applying quantum linear solvers to dense and severely ill-conditioned inverse problems.
Additionally, we mathematically demonstrate that the conditioning transformations preserve the solution of the original system, and that the decoding does not amplify errors in the quantum solution.
We validated the framework by reconstructing HSPs from BSPs in a realistic heart--torso geometry. Numerical experiments demonstrate accurate reconstruction and close agreement between the VQLS and classical solutions. Ablation studies further demonstrate that pre-encoding reduces quantum resource requirements and improves optimization convergence, while Laplacian regularization in pre-encoding enhances reconstruction accuracy. Our framework provides a promising foundation for noninvasive cardiac imaging and has broad potential applicability to other ill-posed inverse problems in scientific, engineering, and biomedical systems as quantum hardware continues to advance.

  \section*{Data Availability Statement}
    The mesh and geometry data supporting the findings of this study are openly available in the PhysioNet/Computing in Cardiology Challenge 2007 at \url{https://physionet.org/content/challenge-2007/1.0.0/}, reference number ~\cite{goldberger2000physiobank}. 
    
    %The code for this paper can be accessed through this link: \href{https://github.com/CiciZhang-star/-Sum-2024_Geometry-aware-Active-Learning-of-Spatiotemporal-Dynamic-System.git}{GitHub Repository}

% \section*{Code Availability}
%   {\color{red}The code for this paper can be accessed through this link: \href{https://liveutk-my.sharepoint.com/:f:/g/personal/xzhan145_vols_utk_edu/IgA1VZad3ZhBQ4odVpT-0cM-AeV7PEaEN2stIJspWuurWqM?e=U0GvFW}{Repository}}

  \section*{Acknowledgement}
 % \spacingset{1}
This research work was supported by the National Heart, Lung, And Blood Institute of the National Institutes of Health under Award Number R01HL172292. The content is solely the responsibility of the authors and does not necessarily represent the official views of the National Institutes of Health. This research used resources of the Oak Ridge Leadership Computing Facility at Oak Ridge National Laboratory, which is supported by the Advanced Scientific Computing Research programs in the Office of Science of the U.S. Department of Energy under Contract No. DE-AC05-00OR22725. The funders played no role in study design, data collection, analysis and interpretation of data, or the writing of this manuscript. %{\color{red}\textbf{(Required by npj Quantum Information: The role of the funding body in the design of the study and collection, analysis, and interpretation of data and in writing the manuscript should also be declared. If the funder has played no role in the research, this should be stated. )}}

\section*{Author Contributions}
XZZ contributed to conceptualization, methodology, software, analysis, and writing. CL contributed to conceptualization, methodology, supervision, and writing. BY contributed to conceptualization, methodology, supervision, funding acquisition, and writing. All authors read and approved the final manuscript.

\section*{Competing Interests}
All authors declare no financial or non-financial competing interests.

  {
    \bibliographystyle{apalike}
    \spacingset{1}
    \bibliography{Ref}
  }

% ==========================================
% TRANSITION TO SUPPLEMENTARY (as in Template/Ref_Main.tex)
% ==========================================
\clearpage % Force a new page
\typeout{SUPPLEMENT-START-PAGE=\the\value{page}}% used to split the PDF for submission

% Reset all counters
\setcounter{section}{0}
\setcounter{subsection}{0}
\setcounter{equation}{0}
\setcounter{figure}{0}
\setcounter{table}{0}
\setcounter{proposition}{0}
\setcounter{footnote}{0}
\setcounter{page}{1}
\hypersetup{pageanchor=false}% avoid duplicate page anchors after the page reset

% Add the "S" prefix to all labels (and unique hyperlink targets)
\renewcommand{\thesection}{S\arabic{section}}
\renewcommand{\theequation}{S\arabic{equation}}
\renewcommand{\thefigure}{S\arabic{figure}}
\renewcommand{\thetable}{S\arabic{table}}
\renewcommand{\theproposition}{S\arabic{proposition}}
\renewcommand{\theHsection}{S\arabic{section}}
\renewcommand{\theHequation}{S\arabic{equation}}
\renewcommand{\theHfigure}{S\arabic{figure}}
\renewcommand{\theHtable}{S\arabic{table}}
\renewcommand{\theHproposition}{S\arabic{proposition}}

% Smaller section headings for the supplementary material
\makeatletter
    \renewcommand\section{\@startsection {section}{1}{\z@}%
                                       {-3.5ex \@plus -1ex \@minus -.2ex}%
                                       {2.3ex \@plus.2ex}%
                                       {\normalfont\fontfamily{phv}\fontsize{14}{17}\bfseries}}
    \renewcommand\subsection{\@startsection{subsection}{2}{\z@}%
                                         {-3.25ex\@plus -1ex \@minus -.2ex}%
                                         {1.5ex \@plus .2ex}%
                                         {\normalfont\fontfamily{phv}\fontsize{12}{15}\bfseries}}
\makeatother

% Supplementary Title
\begin{center}
  {\LARGE \textbf{Supplementary Material}} \vspace{1em} \\
  {\large Xizhuo (Cici) Zhang, Chao Lu, and Bing Yao$^*$\footnote{Xizhuo (Cici) Zhang and Bing Yao ($^*$corresponding author: byao3@utk.edu) are with the Department of Industrial and Systems Engineering, The University of Tennessee, Knoxville, TN 37996 USA. Chao Lu is with the National Center for Computational Sciences, Oak Ridge National Laboratory, Oak Ridge, TN 37830 USA.}}
\end{center}
% \vspace{2em}

% ==========================================
% IMPORT SUPPLEMENTARY CONTENT
% ==========================================
\appendix
\numberwithin{equation}{section}

This supplementary document provides additional details that support the main manuscript titled: \textit{``Inverse ECG Reconstruction via a Preconditioned Variational Quantum Linear Solver''}. 

% Equations, sections, figures, and tables numbered without the prefix ``S'' refer to the main manuscript.

\section{Theoretical Properties of the Proposed Framework}
\label{app:theory}

This appendix establishes the properties of the pre-encoding, the coherent VQLS, and the decoding stage that are invoked in Section~\ref{s:method}. Throughout, $\norm{\cdot}$ denotes the Euclidean norm for vectors and the induced spectral norm for matrices, $\mathbf{M}^{+}$ denotes the Moore-Penrose pseudoinverse of a matrix $\mathbf{M}$, and the remaining notation follows Section~\ref{s:method}.

\subsection{Forward-Error Certificate}
\label{app:certificate}

%{\color{orange}(Old version)Let $\mathbf{R} \in \mathbb{R}^{m \times n}$ have full column rank with singular values $\omega_1 \ge \dots \ge \omega_n > 0$. Its pseudoinverse is then $\mathbf{R}^{+} = (\mathbf{R}^\top\mathbf{R})^{-1}\mathbf{R}^\top$, which has two properties used below: (P1) $\mathbf{R}^{+}\mathbf{R} = \mathbf{I}_n$, and (P2) $\norm{\mathbf{R}^{+}} = 1/\omega_n$, so that $\kappa_2(\mathbf{R}) = \norm{\mathbf{R}}\norm{\mathbf{R}^{+}} = \omega_1/\omega_n$. Note that $\mathbf{R}\mathbf{R}^{+} \neq \mathbf{I}_m$ in general, since $\mathbf{R}\mathbf{R}^{+}$ is the orthogonal projector onto the column space of $\mathbf{R}$. The proof below therefore only applies the exact round trip $\mathbf{R}^{+}\mathbf{R}$ to vectors in $\mathbb{R}^{n}$.}

%{\color{blue}The forward-error certificate in Section~\ref{s:coarse} relates a candidate's data residual to its error relative to the true HSP. It supplies the admissibility test for noise-free regularization-parameter selection. Here, we state the rank and noise assumptions explicitly, and explain why the same bound can be too conservative for selecting a parameter under measurement noise.}

\begin{proposition}[Forward-error certificate]
\label{prop:certificate}
Let $\mathbf{R} \in \mathbb{R}^{m \times n}$ have full column rank, and let $\tilde{\boldsymbol{y}} = \mathbf{R}\boldsymbol{x}^\star + \boldsymbol{\xi}$ with $\tilde{\boldsymbol{y}} \neq \boldsymbol{0}$ and $\norm{\boldsymbol{\xi}} \le \eta\norm{\tilde{\boldsymbol{y}}}$ for some $\eta \in [0, 1)$. Then $\boldsymbol{x}^\star \neq \boldsymbol{0}$, and every $\boldsymbol{x} \in \mathbb{R}^{n}$ satisfies
\begin{equation}
   RE= \frac{\norm{\boldsymbol{x} - \boldsymbol{x}^\star}}{\norm{\boldsymbol{x}^\star}} \le \kappa_2(\mathbf{R}) \, \frac{\rho(\boldsymbol{x}) + \eta}{1 - \eta}, 
    \qquad
    \rho(\boldsymbol{x}) = \frac{\norm{\mathbf{R}\boldsymbol{x} - \tilde{\boldsymbol{y}}}}{\norm{\tilde{\boldsymbol{y}}}}.
    \label{eq:app_certificate}
\end{equation}
\end{proposition}

\begin{proof}

Let $\omega_1 \ge \dots \ge \omega_n>0$ be the singular values of $\mathbf{R}$. The full column rank assumption yields $\mathbf{R}^{+}\mathbf{R}=\mathbf{I}_n$, $\norm{\mathbf{R}^{+}}=1/\omega_n$, and $\norm{\mathbf{R}}=\omega_1$. Thus, the reconstruction error can be recovered from the residual and the noise through
\begin{equation}
\begin{aligned}
    \boldsymbol{x} - \boldsymbol{x}^\star
    =\mathbf{R}^{+}\mathbf{R}(\boldsymbol{x} - \boldsymbol{x}^\star) =\mathbf{R}^{+}\bigl[(\mathbf{R}\boldsymbol{x}-\tilde{\boldsymbol{y}})+\boldsymbol{\xi}\bigr],
\end{aligned}
    \label{eq:app_error_identity}
\end{equation}
which is used to derive an upper bound on the numerator of $RE$:
\begin{equation}
\begin{aligned}
    \norm{\boldsymbol{x} - \boldsymbol{x}^\star}
    \le \norm{\mathbf{R}^{+}}\bigl(\norm{\mathbf{R}\boldsymbol{x}-\tilde{\boldsymbol{y}}}+\norm{\boldsymbol{\xi}}\bigr) \le \frac{\norm{\tilde{\boldsymbol{y}}}}{\omega_n}\bigl(\rho(\boldsymbol{x})+\eta\bigr).
\end{aligned}
    \label{eq:app_numerator}
\end{equation}
Additionally, the reverse triangle inequality gives
$\norm{\mathbf{R}\boldsymbol{x}^\star}
=\norm{\tilde{\boldsymbol{y}}-\boldsymbol{\xi}}
\ge (1-\eta)\norm{\tilde{\boldsymbol{y}}}$, combining which with $\norm{\mathbf{R}\boldsymbol{x}^\star}\le\omega_1\norm{\boldsymbol{x}^\star}$ gives the lower bound on the denominator of $RE$
\begin{equation}
    \norm{\boldsymbol{x}^\star}\ge
    \frac{(1-\eta)\norm{\tilde{\boldsymbol{y}}}}{\omega_1}>0.
    \label{eq:app_denominator}
\end{equation}
%The strict positivity follows from $\tilde{\boldsymbol{y}}\neq\boldsymbol{0}$ and $\eta<1$. 
Dividing the upper bound in Eq.~\eqref{eq:app_numerator} by the positive lower bound in Eq.~\eqref{eq:app_denominator}, and applying the fact that $\omega_1/\omega_n=\kappa_2(\mathbf{R})$, proves Eq.~\eqref{eq:app_certificate}.
\end{proof}

\textbf{Implication for parameter selection.} The certificate in Eq.~\eqref{eq:certificate} separates into a residual-dependent term and a noise-dependent floor:
\begin{equation}
\begin{aligned}
    \beta(\boldsymbol{x},\eta)
    &=\underbrace{\kappa_2(\mathbf{R})\frac{\rho(\boldsymbol{x})}{1-\eta}}_{\text{depends on the candidate}}
    \quad+\underbrace{\kappa_2(\mathbf{R})\frac{\eta}{1-\eta}}_{\text{independent of the candidate}}.
\end{aligned}
    \label{eq:app_certificate_split}
\end{equation}
 When $\eta=0$ (noise-free data), the admissibility condition
$\beta(\boldsymbol{x}_\text{c}(\lambda),0)\le\tau$ is equivalent to
$\rho(\boldsymbol{x}_\text{c}(\lambda))\le\tau/\kappa_2(\mathbf{R})$.
Because this residual is nondecreasing with $\lambda$,
Eq.~\eqref{eq:lambda_cert} selects the largest regularization weight
satisfying this constraint, giving the strongest regularization
compatible with the prescribed error certificate. Under noisy conditions, no choice of the candidate or of $\lambda$ can lower the certificate below $\kappa_2(\mathbf{R})\eta/(1-\eta)$. If
\[
    \eta>\frac{\tau}{\kappa_2(\mathbf{R})+\tau},
\]
this floor already exceeds $\tau$, and the certificate admits no candidate. %For illustration, $\kappa_2(\mathbf{R})=10^5$ and $\tau=0.10$ give a threshold of approximately $10^{-6}$. The bound remains valid under its assumptions, but it can become uninformative for parameter selection. 
This limitation motivates the discrepancy rule in Eq.~\eqref{eq:lambda_morozov}, whose numerical implementation is specified in Appendix~\ref{app:lambda_search}.

\subsection{Solution Invariance and Conditioning of the Encoded System}
\label{app:invariance}

%{\color{blue}The following proposition separates two claims used in Section~\ref{s:krylov}. Invertible left transformations preserve the compressed-system solution, whereas the exact unit-condition-number conclusion additionally requires the tridiagonal structure that limits the number of nonunit singular values.}
\begin{proposition}[Solution invariance and conditioning]
\label{prop:invariance}
Assume $\mathbf{R}$ has full column rank, $\mathbf{B} \in \mathbb{R}^{n \times N}$ has orthonormal columns, $\mathbf{Q} \in \mathbb{R}^{N \times N}$ is orthogonal, and let $\mathbf{T}$, $\boldsymbol{h}$, $\mathbf{P}$, $\mathbf{A}_0$, $\boldsymbol{b}_0$, $\mathbf{F}_\text{tail}$, $\mathbf{A}_\text{q}$, and $\boldsymbol{b}_\text{q}$ be constructed as in Section~\ref{s:krylov}. Then:
\begin{enumerate}[(a)]
    \item $\mathbf{T}$ and $\mathbf{P}$ are symmetric positive definite, and $\mathbf{A}_0$, $\mathbf{F}_\text{tail}$, and $\mathbf{A}_\text{q}$ are invertible.
    \item The systems $\mathbf{T}\boldsymbol{z} = \boldsymbol{h}$, $\mathbf{A}_0\boldsymbol{z} = \boldsymbol{b}_0$, and $\mathbf{A}_\text{q}\boldsymbol{z} = \boldsymbol{b}_\text{q}$ have the same unique solution $\boldsymbol{z}^\star$, and $\boldsymbol{\Phi}\boldsymbol{z}^\star = \arg\min_{\boldsymbol{x} \in \operatorname{range}(\mathbf{B})} \norm{\mathbf{R}\boldsymbol{x} - \tilde{\boldsymbol{y}}}$.
    \item The singular values of $\mathbf{A}_\text{q}$ are $f_j\gamma_j$, $j = 1, \dots, N$, where $f_j = 1$ for $j \notin \mathcal{J}$ ($\mathcal{J}$ is defined in Eq.~(\ref{def-J})). As such , all singular values of $\mathbf{A}_\text{q}$ lie in $[\gamma_{N-r}, \gamma_{r+1}]$, and $\kappa_2(\mathbf{A}_\text{q}) = \gamma_{r+1}/\gamma_{N-r}$.
\end{enumerate}
\end{proposition}

\begin{proof}
 \textbf{(a) Positive definiteness and invertibility.}
For $\boldsymbol{u}\neq\boldsymbol{0}$, orthonormality of $\mathbf{B}$ ensures that $\mathbf{B}\boldsymbol{u}\neq\boldsymbol{0}$, and full column rank of $\mathbf{R}$ further guarantees that $\mathbf{R}\mathbf{B}\boldsymbol{u}\neq\boldsymbol{0}$. Hence, $\mathbf{G}=(\mathbf{R}\mathbf{B})^\top(\mathbf{R}\mathbf{B})$ is symmetric positive definite because
\[
    \boldsymbol{u}^{\top}\mathbf{G}\boldsymbol{u}
    =\norm{\mathbf{R}\mathbf{B}\boldsymbol{u}}^{2}>0.
\]
Similarly, because $\mathbf{Q}$ is orthogonal, $\mathbf{Q}\boldsymbol{z}\neq\boldsymbol{0}$ for $\boldsymbol{z}\neq\boldsymbol{0}$, and
\[
    \boldsymbol{z}^{\top}\mathbf{T}\boldsymbol{z}
    =(\mathbf{Q}\boldsymbol{z})^{\top}\mathbf{G}(\mathbf{Q}\boldsymbol{z})>0
\]
establishing that $\mathbf{T}$ is symmetric positive definite. As such, each diagonal block $\mathbf{T}_{jj}$ is positive definite and admits a Cholesky factorization, which ensures that $\mathbf{P}$ is symmetric positive definite. Furthermore, it follows directly that $\mathbf{A}_0=\mathbf{P}^{-1}\mathbf{T}$ is invertible and positive definite, and thus all singular values of $\mathbf{A}_0$ are positive, i.e., $\gamma_N>0$. Additionally, the tail factors $f_j$, $j\in\mathcal{J}$, defined in Section~\ref{s:krylov} as singular-value ratios of $\mathbf{A}_0$, are finite and positive, while for $j\notin\mathcal{J}$ we have $f_j=1$.
%{\color{orange}It follows that $\mathbf{A}_0=\mathbf{P}^{-1}\mathbf{T}$ is invertible, so $\gamma_N>0$. \color{red}{(This sentence suddenly appears without any connection with the previous writing...->Revised)} All tail factors are positive, and the central factors equal one.}{\color{blue}It remains to show that $\mathbf{A}_0$, $\mathbf{F}_\text{tail}$, and $\mathbf{A}_\text{q}$ are invertible. Since symmetric positive definite matrices have positive eigenvalues, $\mathbf{T}$ and $\mathbf{P}$ are invertible, and so is $\mathbf{A}_0=\mathbf{P}^{-1}\mathbf{T}$. Thus all singular values of $\mathbf{A}_0$ are positive, i.e., $\gamma_N>0$. Therefore, the tail factors $f_j$, $j\in\mathcal{J}$, defined in Section~\ref{s:krylov} as singular-value ratios, are finite and positive, while for $j\notin\mathcal{J}$ we have $f_j=1$.}
Completing the left singular vectors to the full orthogonal matrix $\mathbf{U}_\text{L}$ yields the equivalent representation
\begin{equation}
    \mathbf{F}_\text{tail}
    =\mathbf{U}_\text{L}\operatorname{diag}(f_1,\dots,f_N)\mathbf{U}_\text{L}^{\top}.
    \label{eq:app_tail_diagonal}
\end{equation}
% Hence, $\mathbf{F}_\text{tail}$ and $\mathbf{A}_\text{q}=\mathbf{F}_\text{tail}\mathbf{A}_0$ are invertible. 
Because every $f_j$ is positive, $\mathbf{F}_\text{tail}$ is
invertible. As such,
$\mathbf{A}_\text{q}=\mathbf{F}_\text{tail}\mathbf{A}_0$
is also invertible.
%{\color{orange}This argument uses full column rank of $\mathbf{R}\mathbf{B}$; the stated full-column-rank assumption on $\mathbf{R}$ is a sufficient, stronger condition. {\color{red}\textbf{(Why this sentence suddenly appears at here? -> I forget to delete it. It's my own notes.)}}}

\textbf{(b) Preservation of the solution.} Because $\mathbf{F}_\text{tail}$ and $\mathbf{P}$ are invertible, we have
\begin{equation}
    \mathbf{A}_\text{q}\boldsymbol{z} = \boldsymbol{b}_\text{q}
    \iff \mathbf{A}_0\boldsymbol{z} = \boldsymbol{b}_0
    \iff \mathbf{T}\boldsymbol{z} = \boldsymbol{h}
    \iff \mathbf{G}(\mathbf{Q}\boldsymbol{z}) = \boldsymbol{g},
\end{equation}
where the last equivalence follows from $\mathbf{Q}\mathbf{Q}^\top = \mathbf{I}_N$. The system $\mathbf{G}\boldsymbol{u} = \mathbf{G}(\mathbf{Q}\boldsymbol{z}) = \boldsymbol{g}$ can be solved by the normal equations of $\min_{\boldsymbol{u}} \norm{\mathbf{R}\mathbf{B}\boldsymbol{u} - \tilde{\boldsymbol{y}}}$, whose solution $\boldsymbol{u}^\star$ is unique by (a). Therefore, $\boldsymbol{z}^\star = \mathbf{Q}^\top\boldsymbol{u}^\star$ is the unique solution of all three systems, and $\boldsymbol{\Phi}\boldsymbol{z}^\star = \mathbf{B}\mathbf{Q}\mathbf{Q}^\top\boldsymbol{u}^\star = \mathbf{B}\boldsymbol{u}^\star$ minimizes $\norm{\mathbf{R}\boldsymbol{x} - \tilde{\boldsymbol{y}}}$ over $\boldsymbol{x} = \mathbf{B}\boldsymbol{u} \in \operatorname{range}(\mathbf{B})$.

\textbf{(c) Singular values after tail correction.} Using $\mathbf{U}_\text{L}^\top\mathbf{U}_\text{L} = \mathbf{I}_N$, we have
\begin{equation}
    \mathbf{A}_\text{q} = \mathbf{F}_\text{tail}\mathbf{A}_0 = \mathbf{U}_\text{L}\operatorname{diag}(f_1, \dots, f_N)\mathbf{U}_\text{L}^\top\mathbf{U}_\text{L}\boldsymbol{\Gamma}\mathbf{U}_\text{R}^\top = \mathbf{U}_\text{L}\operatorname{diag}(f_1\gamma_1, \dots, f_N\gamma_N)\mathbf{U}_\text{R}^\top.
    \label{eq: A_q decomp}
\end{equation}
%{\color{orange}Because $\mathbf{U}_\text{L}$ and $\mathbf{U}_\text{R}$ are orthogonal and $f_j\gamma_j > 0$, Eq. (\ref{eq: A_q decomp}) is a singular value decomposition of $\mathbf{A}_\text{q}$ {\color{red}up to the ordering of the singular values \textbf{(?->Revised. Because $\gamma_1 \ge \dots \ge \gamma_N > 0$, these entries are positive and nonincreasing in $j$.)}}, which are $f_j\gamma_j$. By the definition of $f_j$, we have $f_j\gamma_j = \gamma_{r+1}$ for $j \le r$, $f_j\gamma_j = \gamma_j \in [\gamma_{N-r}, \gamma_{r+1}]$ for $r < j \le N - r$, and $f_j\gamma_j = \gamma_{N-r}$ for $j > N - r$. The largest and smallest singular values of $\mathbf{A}_\text{q}$ are thus $\gamma_{r+1}$ and $\gamma_{N-r}$, respectively.Consequently, $\kappa_2(\mathbf{A}_\text{q})=\gamma_{r+1}/\gamma_{N-r}$.}

By the definition of the rescaling factors $f_j$ in
Section~\ref{s:krylov},
\[
    f_j\gamma_j=
    \begin{cases}
        \gamma_{r+1}, & 1\le j\le r,\\
        \gamma_j,     & r<j\le N-r,\\
        \gamma_{N-r}, & N-r<j\le N.
    \end{cases}
\]
Because $\gamma_1\ge\dots\ge\gamma_N>0$, these values are
positive and arranged in nonincreasing order.
Together with the orthogonality of $\mathbf{U}_\text{L}$ and
$\mathbf{U}_\text{R}$, this establishes that
Eq.~\eqref{eq: A_q decomp} is a singular value decomposition
of $\mathbf{A}_\text{q}$.
Its singular values lie in
$[\gamma_{N-r},\gamma_{r+1}]$, with largest and smallest
values $\gamma_{r+1}$ and $\gamma_{N-r}$, respectively.
As such, $\kappa_2(\mathbf{A}_\text{q})=\gamma_{r+1}/\gamma_{N-r}$.
\end{proof}

Proposition~\ref{prop:invariance}(b) shows that the Krylov projection, the block-Jacobi preconditioning, and the spectral tail correction introduce no approximation, so that the approximation of the pre-encoding lies entirely in the restriction to $\operatorname{range}(\mathbf{B})$. %{\color{orange}Note that when {\color{red}$k > N$ \textbf{(Please re-mention the definition of $k$ and $N$ here to improve the clarity->Revised)}}, $\mathbf{Q}$ has orthonormal columns but is not square, and the Krylov projection constitutes an additional Galerkin approximation. This case does not arise in the configuration $s = N - 1$ \textbf{\color{red}(Please re-mention the definition of $s$ here to improve the clarity->Revised)} adopted in this study.}
Recall that the box-aware subspace $\operatorname{range}(\mathbf{B})$ in Eq.~\eqref{eq:basis} is spanned by the canonical basis vectors of the $s$ heart nodes selected by the box-aware score and by the background direction $\boldsymbol{c}_\text{o}$, so that its dimension is $k = s + 1$. In contrast, the encoded system has the dimension $N = 2^{q}$ of the $q$-qubit register. When $k > N$, the Krylov subspace $\mathcal{K}_N(\mathbf{G}, \boldsymbol{g})$ is a proper subspace of $\mathbb{R}^{k}$, so that $\mathbf{Q} \in \mathbb{R}^{k \times N}$ has orthonormal columns but is not square, and the Krylov projection constitutes an additional Galerkin approximation. This case does not arise in this study, because we select $s = N - 1$ support nodes, so that $k = N$ and $\mathbf{Q}$ is a square orthogonal matrix.

\subsection{Isometric Decoder}
\label{app:isometry}

The decoder is designed to transfer coefficient errors to the heart mesh without changing their Euclidean norm, as illustrated in Proposition \ref{prop:isometry}. %This property is independent of the spectral correction, provided the same orthonormal reconstruction basis is retained.

\begin{proposition}[Isometric decoder]
\label{prop:isometry}
Let $\mathbf{B}$ be constructed as in Eq.~\eqref{eq:basis}, $\mathbf{Q} \in \mathbb{R}^{k \times N}$ have orthonormal columns, $\boldsymbol{\Phi} = \mathbf{B}\mathbf{Q}$ be the decoder in Eq.~\eqref{eq:encoded}, and $\mathcal{P}_{[0,1]}$ denote the box projection $\mathbb{R}^{n} \to [0,1]^{n}$ in Eq.~\eqref{eq:box}. Then $\mathbf{B}^\top\mathbf{B} = \mathbf{I}_k$ and $\boldsymbol{\Phi}^\top\boldsymbol{\Phi} = \mathbf{I}_N$. As such, for all $\boldsymbol{z}_1, \boldsymbol{z}_2 \in \mathbb{R}^{N}$,
\begin{equation}
    \norm{\boldsymbol{\Phi}\boldsymbol{z}_1 - \boldsymbol{\Phi}\boldsymbol{z}_2} = \norm{\boldsymbol{z}_1 - \boldsymbol{z}_2}
    \qquad \text{and} \qquad
    \norm{\mathcal{P}_{[0,1]}(\boldsymbol{\Phi}\boldsymbol{z}_1) - \mathcal{P}_{[0,1]}(\boldsymbol{\Phi}\boldsymbol{z}_2)} \le \norm{\boldsymbol{z}_1 - \boldsymbol{z}_2}.
    \label{eq:app_isometry}
\end{equation}
\end{proposition}

% {\color{blue}In the second relation, the box projection $\mathcal{P}_{[0,1]}$ of Eq.~\eqref{eq:box} is applied to both decoded vectors, as it is to every reconstruction in Section~\ref{s:decoding}. This relation thus bounds the discrepancy between two final reconstructions, e.g., $\hat{\boldsymbol{x}}_\text{VQLS} = \mathcal{P}_{[0,1]}(\boldsymbol{\Phi}\hat{\boldsymbol{z}})$ and $\hat{\boldsymbol{x}}_\text{CLS} = \mathcal{P}_{[0,1]}(\boldsymbol{\Phi}\boldsymbol{z}^\star)$, by the discrepancy between their reduced coordinates.}

% \begin{proof}
% The columns of $\mathbf{E}_{\mathcal{S}}$ are distinct canonical basis vectors, so $\mathbf{E}_{\mathcal{S}}^\top\mathbf{E}_{\mathcal{S}} = \mathbf{I}_s$, and $\norm{\boldsymbol{c}_\text{o}} = 1$ by construction. For every $i \in \mathcal{S}$, $\boldsymbol{e}_i^\top(\mathbf{I}_n - \mathbf{E}_{\mathcal{S}}\mathbf{E}_{\mathcal{S}}^\top)\boldsymbol{x}_\text{c} = x_{\text{c},i} - x_{\text{c},i} = 0$, so $\boldsymbol{c}_\text{o}$ is orthogonal to every column of $\mathbf{E}_{\mathcal{S}}$. Hence, $\mathbf{B}^\top\mathbf{B} = \mathbf{I}_k$, and $\boldsymbol{\Phi}^\top\boldsymbol{\Phi} = \mathbf{Q}^\top\mathbf{B}^\top\mathbf{B}\mathbf{Q} = \mathbf{Q}^\top\mathbf{Q} = \mathbf{I}_N$. The first relation in Eq.~\eqref{eq:app_isometry} follows from $\norm{\boldsymbol{\Phi}(\boldsymbol{z}_1 - \boldsymbol{z}_2)}^2 = (\boldsymbol{z}_1 - \boldsymbol{z}_2)^\top\boldsymbol{\Phi}^\top\boldsymbol{\Phi}(\boldsymbol{z}_1 - \boldsymbol{z}_2)$, and the second follows from the first together with Proposition~\ref{prop:box}.
% \end{proof}

\begin{proof}
In the nondegenerate construction, the canonical columns of $\mathbf{E}_{\mathcal{S}}$ satisfy $\mathbf{E}_{\mathcal{S}}^{\top}\mathbf{E}_{\mathcal{S}}=\mathbf{I}_s$. The background vector $\boldsymbol{c}_\text{o}$ has unit norm and vanishes on $\mathcal{S}$, ensuring its orthogonality to every column of $\mathbf{E}_{\mathcal{S}}$. In the fallback construction, the added columns are unused canonical basis vectors and have the same orthonormality property. Hence, in both cases,
\[ \mathbf{B}^{\top}\mathbf{B}=\mathbf{I}_k,
    \qquad
    \boldsymbol{\Phi}^{\top}\boldsymbol{\Phi}
    =\mathbf{Q}^{\top}\mathbf{B}^{\top}\mathbf{B}\mathbf{Q}
    =\mathbf{I}_N.
\]
Therefore, for any $\boldsymbol{z}_1$ and $\boldsymbol{z}_2$,
\[
    \norm{\boldsymbol{\Phi}(\boldsymbol{z}_1-\boldsymbol{z}_2)}^2
    =(\boldsymbol{z}_1-\boldsymbol{z}_2)^{\top}
      \boldsymbol{\Phi}^{\top}\boldsymbol{\Phi}
      (\boldsymbol{z}_1-\boldsymbol{z}_2)
    =\norm{\boldsymbol{z}_1-\boldsymbol{z}_2}^2
\]
which establishes the first relation in Eq. (\ref{eq:app_isometry}). Applying the non-expansiveness of the box projection from Proposition~\ref{prop:box} to both decoded vectors establishes the second equation in Eq. (\ref{eq:app_isometry}).

% {\color{blue}
% The second relation compares two postprocessed reconstructions, not a projected reconstruction with an unprojected one. In particular, it applies to
% $\hat{\boldsymbol{x}}_\text{VQLS}=\mathcal{P}_{[0,1]}(\boldsymbol{\Phi}\hat{\boldsymbol{z}})$ and
% $\hat{\boldsymbol{x}}_\text{CLS}=\mathcal{P}_{[0,1]}(\boldsymbol{\Phi}\boldsymbol{z}^\star)$
% formed from the same encoded system and the same fixed decoder. It does not bound the error caused by measurement noise, by restricting the reconstruction space, or by changing the data-dependent decoder between two data sets.}
\end{proof}

Applied to the scale-restored VQLS solution $\hat{\boldsymbol{z}}$ and the classical solution $\boldsymbol{z}^\star$ of the same encoded system, Proposition~\ref{prop:isometry} shows that the discrepancy between the two reconstructions on the heart mesh never exceeds their discrepancy in the reduced coordinates. %This property depends on the choice of left preconditioning in Eq.~\eqref{eq:precond}. Under the symmetric alternative $\mathbf{L}^{-1}\mathbf{T}\mathbf{L}^{-\top}$, the solution coordinates change to $\mathbf{L}^\top\boldsymbol{z}$, and the decoder becomes $\mathbf{B}\mathbf{Q}\mathbf{L}^{-\top}$. Because $\mathbf{B}\mathbf{Q}$ has orthonormal columns, the singular values of $\mathbf{B}\mathbf{Q}\mathbf{L}^{-\top}$ coincide with those of $\mathbf{L}^{-\top}$, and hence $\kappa_2(\mathbf{B}\mathbf{Q}\mathbf{L}^{-\top}) = \kappa_2(\mathbf{L}) = \kappa_2(\mathbf{P})^{1/2}$. Errors of equal norm in the reduced coordinates would then be mapped to errors on the heart mesh whose norms differ by a factor of up to $\kappa_2(\mathbf{L})$, which is avoided by left preconditioning.

% {\color{blue}
% The left-preconditioned system retains the coordinate vector $\boldsymbol{z}$ and hence the decoder $\mathbf{B}\mathbf{Q}$. In contrast, set $\mathbf{P}=\mathbf{L}\mathbf{L}^{\top}$ and use the symmetric alternative with the changed coordinates $\mathbf{L}^{\top}\boldsymbol{z}$. Its decoder is $\mathbf{B}\mathbf{Q}\mathbf{L}^{-\top}$. Since multiplication by $\mathbf{B}\mathbf{Q}$ preserves Euclidean norms, the singular values of this decoder equal those of $\mathbf{L}^{-\top}$, giving
% \[
%     \kappa_2(\mathbf{B}\mathbf{Q}\mathbf{L}^{-\top})
%     =\kappa_2(\mathbf{L})=\kappa_2(\mathbf{P})^{1/2}.
% \]
% This ratio describes direction-dependent stretching, not an absolute amplification factor for every error. The largest absolute error gain is $\norm{\mathbf{L}^{-\top}}$, whereas left preconditioning keeps the decoder isometric and preserves every coefficient-error norm.}

\subsection{Expressivity of the M\"{o}tt\"{o}nen Ansatz}
\label{app:expressivity}

\begin{proposition}[Expressivity]
\label{prop:expressivity}
The set of trial states generated by the M\"{o}tt\"{o}nen ansatz in Eq.~\eqref{eq:ansatz}, $\{\ket{\boldsymbol{z}(\boldsymbol{w})} : \boldsymbol{w} \in \mathbb{R}^{N} \setminus \{\boldsymbol{0}\}\}$, is equal to the set of all real unit vectors $\mathbb{S}^{N-1} = \{\boldsymbol{\psi} \in \mathbb{R}^{N} : \norm{\boldsymbol{\psi}} = 1\}$.
\end{proposition}

\begin{proof}
%Every trial state has unit norm by Eq.~\eqref{eq:ansatz}, which shows one inclusion. 
For any $\boldsymbol{\psi} \in \mathbb{S}^{N-1}$, choosing $\boldsymbol{w} = \boldsymbol{\psi}$ and \(\|\boldsymbol{\psi}\|_2=1\) yields $\ket{\boldsymbol{z}(\boldsymbol{w})} = \boldsymbol{\psi}$. As such, M\"{o}tt\"{o}nen state preparation synthesizes a circuit that prepares any prescribed amplitude vector exactly \citep{mottonen2004transformation}.
\end{proof}

%{\color{orange}Because $\ket{\boldsymbol{z}(\nu\boldsymbol{w})} = \ket{\boldsymbol{z}(\boldsymbol{w})}$ for every $\nu > 0$, the $N$ parameters carry $N - 1$ effective degrees of freedom, which equals the dimension of $\mathbb{S}^{N-1}$. The ansatz therefore neither misses any real state of the register nor restricts the optimization to a subset of the state space, and the cost landscape seen by the optimizer is that of the cost function in Eq.~\eqref{eq:cost_def} itself. This invariance also matches the invariance of $C(\boldsymbol{w})$ to the norm of $\boldsymbol{w}$.}

Positive rescaling of the parameter vector leaves both the trial state and the cost unchanged, i.e., $\ket{\boldsymbol{z}(\nu\boldsymbol{w})}
=\ket{\boldsymbol{z}(\boldsymbol{w})}$
and $C(\nu\boldsymbol{w})=C(\boldsymbol{w})$ for every $\nu>0$.
Therefore, the parameter norm is redundant, leaving $N-1$ effective degrees of freedom, matching the dimension of $\mathbb{S}^{N-1}$.
Proposition~\ref{prop:expressivity} establishes that this parameterization covers the entire real unit sphere, so the ansatz can represent the normalized real solution of the encoded system exactly.

\subsection{Coherent Evaluation of the Global Cost (Eq.~\eqref{eq:cost})}
\label{app:cost_derivation}

We adopt the convention that the ancilla register comprising $n_\text{anc}$ qubits is listed first, followed by the system register consisting of $q$ qubits. Moreover, we use the shorthand notation $\ket{\boldsymbol{z}}$ to denote $\ket{\boldsymbol{z}(\boldsymbol{w})}$.

\textbf{Circuit states.} %State preparation gives $\ket{\boldsymbol{\varphi}_0} = \ket{0}^{\otimes n_\text{anc}} \otimes \ket{\boldsymbol{z}}$. By Eq.~\eqref{eq:block_encoding}, applying the block encoding yields
%\begin{equation}
%    \begin{aligned}
%        \ket{\boldsymbol{\varphi}_1}
%        &= \mathbf{U}_{\mathbf{A}}\ket{\boldsymbol{\varphi}_0}
%        = \ket{0}^{\otimes n_\text{anc}} \otimes \frac{\mathbf{A}_\text{q}\ket{\boldsymbol{z}}}{\alpha}
%        + \sum_{j=1}^{2^{n_\text{anc}}-1} \ket{j} \otimes \ket{\boldsymbol{\zeta}_j},
 %   \end{aligned}
%    \label{eq:app_split}
%\end{equation}
%where $\ket{j}$ are the nonzero computational basis states of the ancilla register and $\ket{\boldsymbol{\zeta}_j}$ are unnormalized system vectors. The circuit output is
%\[
%    \ket{\boldsymbol{\varphi}_2}
%    = (\mathbf{I}_\text{anc} \otimes \mathbf{U}_{\boldsymbol{b}}^\dagger)
%      \ket{\boldsymbol{\varphi}_1}.
%\]
To derive the measurement probabilities used in the cost function, we track the joint state of the ancilla and system registers. After initializing the ancillas to zero and preparing the normalized trial state \(|\boldsymbol z\rangle\), the joint state is $|\boldsymbol\varphi_0\rangle
=
|0\rangle^{\otimes n_{\mathrm{anc}}}\otimes|\boldsymbol z\rangle.$
Applying \(\mathbf U_{\mathbf A}\) and expanding the output in the computational basis of the ancilla register yields
\[
|\boldsymbol\varphi_1\rangle
=
\mathbf U_{\mathbf A}|\boldsymbol\varphi_0\rangle
=
\sum_{j=0}^{2^{n_{\mathrm{anc}}}-1}
|j\rangle_{\mathrm{anc}}\otimes|\boldsymbol\zeta_j\rangle,
\]
where $|\boldsymbol\zeta_j\rangle=(\langle j|_{\mathrm{anc}}\otimes\mathbf I_N)
\mathbf U_{\mathbf A}|\boldsymbol\varphi_0\rangle$
is the unnormalized system vector associated with ancilla basis state \(|j\rangle_{\mathrm{anc}}\). In particular, \(|0\rangle_{\mathrm{anc}}=|0\rangle^{\otimes n_{\mathrm{anc}}}\), and the block-encoding identity in Eq.~\(\eqref{eq:block_encoding}\) ensures that $|\boldsymbol\zeta_0\rangle=
\frac{\mathbf A_{\mathrm q}|\boldsymbol z\rangle}{\alpha}$. Separating this component yields
\[
|\boldsymbol\varphi_1\rangle
=
|0\rangle^{\otimes n_{\mathrm{anc}}}
\otimes\frac{\mathbf A_{\mathrm q}|\boldsymbol z\rangle}{\alpha}
+
\sum_{j=1}^{2^{n_{\mathrm{anc}}}-1}
|j\rangle_{\mathrm{anc}}\otimes|\boldsymbol\zeta_j\rangle.
\]
Thus, the all-zero ancilla component  $|\boldsymbol\zeta_0\rangle=$ contains the desired matrix action, while the remaining terms lie in the orthogonal ancilla subspace. This decomposition describes the coherent output state before measurement. Finally, the circuit applies \(\mathbf U_{\boldsymbol b}^{\dagger}\) to the system register and generates the output:
\[
|\boldsymbol\varphi_2\rangle
=
(\mathbf I_{\mathrm{anc}}\otimes\mathbf U_{\boldsymbol b}^{\dagger})
|\boldsymbol\varphi_1\rangle.
\]
\textbf{Measurement probabilities.} We evaluate two measurement probabilities to quantify the norm of the matrix-transformed trial state and its overlap with the normalized right-hand side. The first corresponds to obtaining zero on all ancilla qubits, irrespective of the system-register outcome. Its projector $
\boldsymbol{\Pi}_{\mathrm{anc}}$ selects the component containing the encoded matrix action. Because \(\mathbf U_{\boldsymbol b}^{\dagger}\) acts only on the system register and preserves its norm, it leaves this probability unchanged. By the Born rule \citep{nielsen2010quantum}, we have
\[
\begin{aligned}
p_{\mathrm{anc}}
=
\langle\boldsymbol\varphi_2|
\boldsymbol{\Pi}_{\mathrm{anc}}
|\boldsymbol\varphi_2\rangle
=
\frac{\|\mathbf U_{\boldsymbol b}^{\dagger}
\mathbf A_{\mathrm q}|\boldsymbol z\rangle\|_2^2}{\alpha^2}
=
\frac{\|\mathbf A_{\mathrm q}|\boldsymbol z\rangle\|_2^2}{\alpha^2}.
\end{aligned}
\]
The second probability corresponds to the joint all-zero outcome on the ancilla and system registers, represented by \(\boldsymbol{\Pi}_{\mathrm{glob}}\). Selecting the all-zero ancilla component isolates the encoded matrix action, while the inverse preparation \(\mathbf U_{\boldsymbol b}^{\dagger}\) converts its overlap with \(|\boldsymbol b\rangle\) into an amplitude on the all-zero system state. Indeed, taking the adjoint of \(\mathbf U_{\boldsymbol b}|0\rangle^{\otimes q}=|\boldsymbol b\rangle\) yields $\langle0|^{\otimes q}\mathbf U_{\boldsymbol b}^{\dagger}
=
\langle\boldsymbol b|$.
Hence, the amplitude of the joint all-zero outcome is
\[
\begin{aligned}
\bigl(\langle0|^{\otimes n_{\mathrm{anc}}}
\otimes\langle0|^{\otimes q}\bigr)
|\boldsymbol\varphi_2\rangle
=
\frac{\langle0|^{\otimes q}\mathbf U_{\boldsymbol b}^{\dagger}
\mathbf A_{\mathrm q}|\boldsymbol z\rangle}{\alpha}
=
\frac{\langle\boldsymbol b|\mathbf A_{\mathrm q}
|\boldsymbol z\rangle}{\alpha}.
\end{aligned}
\]
Taking the squared modulus yields
\[
p_{\mathrm{glob}}
=
\langle\boldsymbol\varphi_2|
\boldsymbol{\Pi}_{\mathrm{glob}}
|\boldsymbol\varphi_2\rangle
=
\frac{|\langle\boldsymbol b|\mathbf A_{\mathrm q}
|\boldsymbol z\rangle|^2}{\alpha^2}.
\]

%The unitary $\mathbf{U}_{\boldsymbol{b}}^\dagger$ acts only on the system register, so it leaves the probability of the all-zero ancilla outcome unchanged. By the Born rule \citep{nielsen2010quantum} and Eq.~\eqref{eq:app_split},
%\begin{equation}
%    p_\text{anc}
%    = \bra{\boldsymbol{\varphi}_2}\boldsymbol{\Pi}_\text{anc}\ket{\boldsymbol{\varphi}_2}    = \frac{\norm{\mathbf{A}_\text{q}\ket{\boldsymbol{z}}}^2}{\alpha^2}.
%\end{equation}
%Because $\mathbf{U}_{\boldsymbol{b}}\ket{0}^{\otimes q} = \ket{\boldsymbol{b}}$, the amplitude of the all-zero outcome on both registers is
%\[
%    \bigl(\bra{0}^{\otimes n_\text{anc}} \otimes \bra{0}^{\otimes q}\bigr)
%    \ket{\boldsymbol{\varphi}_2}
%    = \frac{\bra{\boldsymbol{b}}\mathbf{A}_\text{q}\ket{\boldsymbol{z}}}{\alpha}.
%\]
%Thus,
%\begin{equation}
%    p_\text{glob}
%    = \bra{\boldsymbol{\varphi}_2}\boldsymbol{\Pi}_\text{glob}\ket{\boldsymbol{\varphi}_2}    = \frac{\bigl|\bra{\boldsymbol{b}}\mathbf{A}_\text{q}\ket{\boldsymbol{z}}\bigr|^2}{\alpha^2}.
%\end{equation}

\textbf{Cost evaluation.} The two measurement probabilities determine the cost through their ratio. Proposition~\ref{prop:invariance}(a) ensures that $\mathbf{A}_\text{q}$ is invertible, so $p_\text{anc}>0$. Substituting the probability expressions yields
\[
    1-\frac{p_\text{glob}}{p_\text{anc}}
    = 1-\frac{\bigl|\bra{\boldsymbol{b}}\mathbf{A}_\text{q}\ket{\boldsymbol{z}}\bigr|^2}
                 {\norm{\mathbf{A}_\text{q}\ket{\boldsymbol{z}}}^2},
\]
which is the cost defined in Eq.~\eqref{eq:cost_def}. The ratio $p_\text{glob}/p_\text{anc}$ is the conditional probability of the all-zero system outcome after $\mathbf{U}_{\boldsymbol{b}}^\dagger$, given the all-zero ancilla outcome.
Equivalently, it measures the alignment between $\ket{\boldsymbol{b}}$ and the normalized state proportional to $\mathbf{A}_\text{q}\ket{\boldsymbol{z}}$.

%On quantum hardware, these probabilities are estimated from repeated measurements. For the FABLE normalization $\alpha=2^q c_\text{F}$,
%\[
%    p_\text{anc}
%    = \frac{\norm{\mathbf{A}_\text{q}\ket{\boldsymbol{z}}}^2}{4^q c_\text{F}^2}.
%\]
%A small $p_\text{anc}$ increases the sampling effort needed to estimate the conditional probability accurately. In this study, both probabilities are evaluated directly from the simulated state vector, without measurement sampling.

%\textbf{Scale restoration.} For a real unit vector $\hat{\boldsymbol{\psi}}$, minimizing $\norm{\nu\mathbf{A}_\text{q}\hat{\boldsymbol{\psi}}-\boldsymbol{b}_\text{q}}^2$ over $\nu\in\mathbb{R}$ gives a strictly convex quadratic because $\mathbf{A}_\text{q}$ is invertible. Setting its derivative to zero yields
%\[
 %   \nu^\star
%    = \frac{\langle\mathbf{A}_\text{q}\hat{\boldsymbol{\psi}},\boldsymbol{b}_\text{q}\rangle}
           %{\norm{\mathbf{A}_\text{q}\hat{\boldsymbol{\psi}}}^2},
%\]
%as in Eq.~\eqref{eq:scale}. If $\hat{\boldsymbol{\psi}}=\pm\ket{\boldsymbol{z}^\star}$, then $\nu^\star\hat{\boldsymbol{\psi}}=\boldsymbol{z}^\star$.

\subsection{Non-Expansiveness of the Physical Prior}
\label{app:box}

\begin{proposition}[Non-expansive prior]
\label{prop:box}
For all $\boldsymbol{v}_1, \boldsymbol{v}_2 \in \mathbb{R}^{n}$, $\norm{\mathcal{P}_{[0,1]}(\boldsymbol{v}_1) - \mathcal{P}_{[0,1]}(\boldsymbol{v}_2)} \le \norm{\boldsymbol{v}_1 - \boldsymbol{v}_2}$. In particular, every $\boldsymbol{x}^\star \in [0,1]^{n}$ satisfies $\norm{\mathcal{P}_{[0,1]}(\boldsymbol{v}) - \boldsymbol{x}^\star} \le \norm{\boldsymbol{v} - \boldsymbol{x}^\star}$ for all $\boldsymbol{v} \in \mathbb{R}^{n}$.
\end{proposition}

\begin{proof}
The scalar map $\pi(\chi) = \min(\max(\chi, 0), 1)$ is continuous, nondecreasing, and piecewise linear with slopes in $\{0, 1\}$. Hence, $0 \le \pi(\chi_2) - \pi(\chi_1) \le \chi_2 - \chi_1$ for all $\chi_1 \le \chi_2$, i.e., $|\pi(\chi_1) - \pi(\chi_2)| \le |\chi_1 - \chi_2|$ for all $\chi_1, \chi_2 \in \mathbb{R}$. Applying this inequality to each component, squaring, and summing over $i = 1, \dots, n$ yields the first claim. The second claim follows by setting $\boldsymbol{v}_2 = \boldsymbol{x}^\star$ and noting that $\mathcal{P}_{[0,1]}(\boldsymbol{x}^\star) = \boldsymbol{x}^\star$.
\end{proof}

Note that the projection generally increases the data residual, i.e., $\norm{\mathbf{R}\hat{\boldsymbol{x}} - \tilde{\boldsymbol{y}}} \ge \norm{\mathbf{R}\hat{\boldsymbol{x}}_\text{raw} - \tilde{\boldsymbol{y}}}$ may hold, because $\hat{\boldsymbol{x}}$ no longer coincides with the least-squares fit. The prior thus trades data fidelity for physical admissibility without increasing the error with respect to any HSP in the admissible box.

\section{Implementation Details}
\label{app:implementation}

\subsection{Search for the Regularization Weight}
\label{app:lambda_search}

To select the strongest admissible regularization, we locate the upper boundary of the admissible weights within the prescribed search range. For the residual-based criteria in Eqs.~\(\eqref{eq:lambda_cert}\) and \(\eqref{eq:lambda_morozov}\), the residual \(\|\mathbf R\boldsymbol x_{\mathrm c}(\lambda)-\tilde{\boldsymbol y}\|_2\) is nondecreasing with \(\lambda\). Hence, once a weight is admissible, every smaller positive weight is also admissible.
The search begins at a prescribed upper bound \(\lambda_{\max}\), which is returned immediately if it satisfies the criterion. Otherwise, the trial weight is reduced by a factor of ten until an admissible value is found or the lower search limit \(10^{-12}\) is reached. The first admissible value, denoted by \(\lambda_{+}\), and the preceding inadmissible value, denoted by \(\lambda_{-}\), define a bracket with \(\lambda_{+}<\lambda_{-}\). If no admissible positive weight is found within the search range, the procedure returns \(\lambda=0\) as a fallback and records the unsuccessful search. Once a bracket is established, it is refined through a fixed number of bisection steps in \(\log\lambda\). At each step, the trial value is the geometric midpoint, $\lambda_{\mathrm{trial}}=\sqrt{\lambda_{+}\lambda_{-}}$.
If this value satisfies the criterion, it replaces \(\lambda_{+}\); otherwise, it replaces \(\lambda_{-}\). This update preserves an admissible lower endpoint and an inadmissible upper endpoint while halving the bracket width on the logarithmic scale, which is appropriate for weights spanning several orders of magnitude. This procedure eventually returns the final admissible endpoint \(\lambda_{+}\), the largest admissible weight evaluated during the search.

\subsection{Block Partition Search}
\label{app:partition}

%The block sizes of the preconditioner $\mathbf{P}$ in Eq.~\eqref{eq:precond} are selected by an exhaustive search over the ordered compositions of $N$ into contiguous blocks whose sizes lie between $3$ and a prescribed maximum $b_\text{max}$. The number of blocks ranges from $\lceil N/b_\text{max} \rceil$ to $\lceil N/b_\text{max} \rceil + 1$. For each candidate partition, $\kappa_2(\mathbf{P}^{-1}\mathbf{T})$ is evaluated, and the minimizer is retained. The search reads $\mathbf{T}$ only and never the reference HSP. The choice of partition matters because partitions that follow the coupling structure of $\mathbf{T}$ leave the preconditioned matrix considerably closer to the identity than uniform partitions with the same number of blocks. The Cholesky factorization of each selected block additionally verifies its positive definiteness.

The block partition of the preconditioner \(\mathbf P\) in Eq.~\(\eqref{eq:precond}\) is selected to minimize the condition number of the left-preconditioned system. We exhaustively evaluate all admissible ordered partitions of the index set \(\{1,\dots,N\}\) into contiguous blocks. Each block contains between \(3\) and \(b_{\max}\) indices, and the number of blocks is restricted to \(\lceil N/b_{\max}\rceil\) or \(\lceil N/b_{\max}\rceil+1\), subject to these size constraints. For each candidate partition, we construct \(\mathbf P\), evaluate \(\kappa_2(\mathbf P^{-1}\mathbf T)\), and retain a partition attaining the smallest value. This partition determines which couplings in \(\mathbf T\) are incorporated into the diagonal blocks and which remain between blocks. Allowing the nonuniform block sizes enables the preconditioner to account for the coupling structure of \(\mathbf T\), potentially improving conditioning relative to a uniform partition. 

\subsection{Evaluation of the All-Zero Projector}
\label{app:projector}

%The global projector $\boldsymbol{\Pi}_\text{glob}$ on all $2q + 1$ qubits is implemented as the product of single-qubit projectors, $\boldsymbol{\Pi}_\text{glob} = \bigotimes_{j} \ket{0}\bra{0}_j$. The two representations are mathematically identical. A single multi-qubit projector, however, is represented under adjoint differentiation by a dense Hermitian matrix of size $2^{2q+1} \times 2^{2q+1}$, whereas the product form requires memory that grows only linearly with the number of qubits.

To reduce the memory required to represent the measurement observables during adjoint differentiation, we express the global all-zero projector as a tensor product of single-qubit projectors:
$\boldsymbol{\Pi}_{\mathrm{glob}}=\bigotimes_{j=1}^{2q+1}\bigl(|0\rangle\langle0|\bigr)_j$. This factorized expression is mathematically identical to the projector onto the joint all-zero state. Explicitly storing it as a dense matrix would require \(2^{2q+1}\times2^{2q+1}\) entries, whereas storing the individual factors requires memory linear in the number of qubits. Therefore, the factorized representation avoids this dense observable allocation, provided that the simulation backend preserves and processes the tensor-product structure.
The same consideration applies to $\boldsymbol{\Pi}_\text{anc}$ when the number of ancilla qubits is large.

\subsection{Configuration Summary}
\label{app:config}

Table~\ref{Tab:config} summarizes the configuration of the complete model used in the experimental study.

\begin{table}[!ht]
    \centering
    \caption{Configuration of the proposed framework.}
    \label{Tab:config}
    \small
    \begin{tabular}{@{}lll@{}}
        \toprule
        Setting & Value & Description \\
        \midrule
        $N = 2^{q}$ & $64$ ($q = 6$) & Encoded dimension and number of system qubits \\
        $s$ & $N - 1 = 63$ & Number of selected support nodes in pre-encoding \\
        $\tau$ & $0.10$ & Certified error level for noise-free data, Eq.~\eqref{eq:lambda_cert} \\
        $\lambda_\text{max}$, bisection steps (noise-free) & $18.0067$, $4$ & Search for Eq.~\eqref{eq:lambda_cert} \\
        $\lambda_\text{max}$, bisection steps (noisy) & $10^{4}$, $6$ & Search for Eq.~\eqref{eq:lambda_morozov} \\
        Block sizes & $3$ to $b_\text{max} = 16$ & Bounds of the partition search \\
        Preconditioning & Left & Keeps the decoder an isometry \\
        $r$ & $8$ & Tail rank per side ($16$ corrected singular values) \\
        Ansatz & M\"{o}tt\"{o}nen & $N$ real parameters \\
        Block encoding & FABLE & $n_\text{anc} = q + 1$ ancilla qubits \\
        Optimizer & L-BFGS-B & $500$ iterations, convergence threshold $10^{-10}$ on $C(\boldsymbol{w})$ \\
        Restarts & $3$ & Seeds $2025$, $2026$, and $2027$; unit-norm initialization \\
        Physical prior & $[0, 1]$ & Euclidean projection, Eq.~\eqref{eq:box} \\
        \bottomrule
    \end{tabular}
\end{table}

\section{Summary of Notations}
\label{app:notations}

Table~\ref{Tab:notations} summarizes the notation used throughout this paper. Bold upright letters denote matrices, including quantum gates and operators; bold italic letters denote vectors; and italic letters denote scalars. 
% Subscripts and superscripts set in upright text are abbreviations, whereas those that refer to a variable follow the convention of that variable.

\begin{longtable}{@{}p{0.34\linewidth}p{0.62\linewidth}@{}}
\caption{Summary of notations.}
\label{Tab:notations} \\
\toprule
\textbf{Symbol} & \textbf{Description} \\
\midrule
\endfirsthead

\multicolumn{2}{c}{{\bfseries \tablename\ \thetable{} (continued)}} \\
\toprule
\textbf{Symbol} & \textbf{Description} \\
\midrule
\endhead

\midrule \multicolumn{2}{r}{{Continued on next page}} \\
\endfoot

\bottomrule
\endlastfoot

%=========================================================
% Operators and functions
%=========================================================
\multicolumn{2}{@{}l}{\textbf{Operators and Functions}}\\
\addlinespace[0.25em]
$\norm{\cdot}$ & Euclidean norm of a vector; induced spectral norm of a matrix \\
$(\cdot)^\top$, $(\cdot)^{-1}$, $(\cdot)^{+}$ & Transpose, inverse, and Moore-Penrose pseudoinverse \\
$(\cdot)^\dagger$ & Conjugate transpose \\
$\kappa_2(\cdot)$ & Spectral condition number \\
$\operatorname{diag}(\cdot)$, $\operatorname{blockdiag}(\cdot)$ & Diagonal and block-diagonal matrix formation \\
$\operatorname{span}(\cdot)$, $\operatorname{range}(\cdot)$ & Linear span; column space of a matrix \\
$\otimes$ & Tensor (Kronecker) product \\
$\ket{\cdot}$, $\bra{\cdot}$ & Quantum state vector and its conjugate transpose \\
$\langle \cdot, \cdot \rangle$ & Inner product \\
$\mathcal{P}_{[0,1]}(\cdot)$ & Euclidean projection onto the box $[0,1]^{n}$ \\
$\pi(\chi)$ & Scalar clipping map $\min(\max(\chi, 0), 1)$ \\
$\beta(\boldsymbol{x}, \eta)$ & Forward-error certificate of a candidate HSP \\
$\rho(\boldsymbol{x})$ & Relative forward residual of a candidate HSP \\
$C(\boldsymbol{w})$ & Normalized global VQLS cost function \\
$d_\text{tr}(\cdot, \cdot)$ & Trace distance between two pure states \\
$\mathcal{N}(\cdot, \cdot)$ & Gaussian distribution \\
\addlinespace[0.35em]

%=========================================================
% Sets and spaces
%=========================================================
\hline \multicolumn{2}{@{}l}{\textbf{Sets and Spaces}}\\
\addlinespace[0.25em]
$\mathbb{R}^{n}$, $\mathbb{R}^{m \times n}$ & Real vector and matrix spaces \\
$\mathcal{S}$ & Active support of the selected heart nodes, $|\mathcal{S}| = s$ \\
$\mathcal{J}$ & Indices of the $r$ largest and the $r$ smallest singular values of $\mathbf{A}_0$ \\
$\mathcal{K}_N(\mathbf{G}, \boldsymbol{g})$ & Order-$N$ Krylov subspace generated by $\mathbf{G}$ and $\boldsymbol{g}$ \\
$\mathbb{S}^{N-1}$ & Set of real unit vectors in $\mathbb{R}^{N}$ \\
\addlinespace[0.35em]

%=========================================================
% Dimensions and indices
%=========================================================
\hline \multicolumn{2}{@{}l}{\textbf{Dimensions and Indices}}\\
\addlinespace[0.25em]
$n$ & Number of heart-surface nodes \\
$m$ & Number of torso electrodes \\
$q$ & Number of system qubits \\
$N = 2^{q}$ & Dimension of the encoded system \\
$s$ & Support size \\
$k = s + 1$ & Subspace dimension \\
$r$ & Tail rank per side of the spectral tail correction \\
$n_\text{anc} = q + 1$ & Number of ancilla qubits of the FABLE block encoding \\
$n_\text{b}$ & Number of blocks of the block-Jacobi preconditioner \\
$b_\text{max}$ & Maximum block size in the partition search \\
$t$ & Time instance \\
$i$, $j$ & Generic indices \\
\addlinespace[0.35em]

%=========================================================
% Scalars
%=========================================================
\hline \multicolumn{2}{@{}l}{\textbf{Scalars}}\\
\addlinespace[0.25em]
$\lambda$, $\lambda^\star$ & Regularization weight and its selected value \\
$\lambda_\text{max}$ & Starting weight of the regularization search \\
$\lambda_{+}$, $\lambda_{-}$ & First admissible and last inadmissible trial weights \\
$\tau$ & Prescribed level of the certified relative error \\
$\eta$ & Noise level relative to the measured BSP \\
$\eta_\text{c}$ & Noise level relative to the noise-free BSP \\
$\sigma_{\boldsymbol{\xi}}$ & Dimensionless noise level \\
$\sigma$ & Reference standard deviation of the BSP recording \\
$\omega_1$, \dots, $\omega_n$ & Singular values of $\mathbf{R}$ (or of a generic $\mathbf{M}$) in descending order \\
$\gamma_1$, \dots, $\gamma_N$ & Singular values of $\mathbf{A}_0$ in descending order \\
$f_j$ & Rescaling factor of the $j$-th singular value in the spectral tail correction \\
$x_{\text{c},i}$ & $i$-th entry of the coarse estimate \\
$p_i$ & Box-projected score of node $i$ \\
$v_i$ & $i$-th entry of a generic vector $\boldsymbol{v}$ \\
$w_i$ & $i$-th trainable parameter of the M\"{o}tt\"{o}nen ansatz \\
$\alpha$ & Subnormalization factor of the block encoding \\
$c_\text{F}$ & Rescaling factor of the FABLE encoding \\
$\bar{a}_{ij}$ & Rescaled $(i,j)$ entry of $\mathbf{A}_\text{q}$ \\
$\theta$, $\theta_{ij}$ & Rotation angle; FABLE rotation angle of entry $(i,j)$ \\
$p_\text{anc}$ & Probability that all ancilla qubits are measured in state $\ket{0}$ \\
$p_\text{glob}$ & Probability that all qubits are measured in state $\ket{0}$ \\
$\nu$, $\nu^\star$ & Scale of the recovered solution direction and its optimal value \\
$\varepsilon$ & Perturbation size in the sharpness example of the certificate \\
$\chi$, $\chi_1$, $\chi_2$ & Real arguments of the clipping map \\
$\epsilon_\text{mach}$ & Machine precision \\
$\mathrm{i}$ & Imaginary unit \\
$\text{RE}$ & Relative error of a reconstructed HSP \\
$F$ & Fidelity between the CLS and VQLS reconstructions \\
$a_\text{AP}$, $c_\text{AP}$, $e_{0,\text{AP}}$ & Parameters of the Aliev-Panfilov model \\
$\mu_{1,\text{AP}}$, $\mu_{2,\text{AP}}$ & Parameters of the Aliev-Panfilov model \\
$\delta_\text{AP}$ & Diffusion coefficient of the Aliev-Panfilov model \\
$\Delta t$ & Time step of the cardiac simulation \\
\addlinespace[0.35em]

%=========================================================
% Vectors
%=========================================================
\hline \multicolumn{2}{@{}l}{\textbf{Vectors}}\\
\addlinespace[0.25em]
$\boldsymbol{x}(t)$ & HSP at time $t$ \\
$\boldsymbol{x}^\star$ & True HSP of a frame \\
$\boldsymbol{x}$ & Candidate HSP in the certificate and the regularized problems \\
$\boldsymbol{y}(t)$ & Noise-free BSP at time $t$ \\
$\tilde{\boldsymbol{y}}(t)$ & Measured BSP at time $t$ \\
$\boldsymbol{\xi}(t)$ & Measurement noise at time $t$ \\
$\boldsymbol{x}_\text{c}(\lambda)$, $\boldsymbol{x}_\text{c}$ & Coarse estimate and its value at $\lambda^\star$ \\
$\hat{\boldsymbol{x}}_\text{raw}$ & Decoded HSP before the physical prior \\
$\hat{\boldsymbol{x}}$ & Reconstructed HSP \\
$\hat{\boldsymbol{x}}_\text{CLS}$, $\hat{\boldsymbol{x}}_\text{VQLS}$ & Reconstructed HSPs obtained with the CLS and the VQLS \\
$\boldsymbol{e}_i$ & $i$-th canonical basis vector \\
$\boldsymbol{c}_\text{o}$ & Background direction of the subspace \\
$\boldsymbol{u}$, $\boldsymbol{u}^\star$ & Subspace coordinates; solution of the compressed system \\
$\boldsymbol{g}$ & Right-hand side of the compressed system \\
$\boldsymbol{q}_j$ & $j$-th Lanczos basis vector \\
$\boldsymbol{h}$ & Right-hand side in Krylov coordinates \\
$\boldsymbol{b}_0$ & Right-hand side after block-Jacobi preconditioning \\
$\boldsymbol{b}_\text{q}$ & Right-hand side of the encoded system \\
$\boldsymbol{b}$ & Right-hand side of the system without pre-encoding (Ablation 1) \\
$\boldsymbol{f}_{\mathcal{J}}$ & Rescaling factors indexed by $\mathcal{J}$ \\
$\boldsymbol{z}$, $\boldsymbol{z}^\star$ & Reduced solution; exact solution of the encoded system \\
$\hat{\boldsymbol{z}}$ & Scale-restored VQLS solution \\
$\boldsymbol{z}_1$, $\boldsymbol{z}_2$ & Generic vectors in the reduced coordinates \\
$\boldsymbol{w}$, $\boldsymbol{w}^{(0)}$, $\hat{\boldsymbol{w}}$ & Trainable parameters; initial and optimized values \\
$\hat{\boldsymbol{\psi}}$ & Optimized unit-norm solution direction \\
$\boldsymbol{\psi}$, $\boldsymbol{\psi}_1$, $\boldsymbol{\psi}_2$ & Generic real unit vectors (pure states) \\
$\boldsymbol{v}$, $\boldsymbol{v}_1$, $\boldsymbol{v}_2$ & Generic vectors in $\mathbb{R}^{n}$ \\
$\boldsymbol{v}^{\mathbf{R}}_1$, $\boldsymbol{v}^{\mathbf{R}}_n$ & Right singular vectors of $\mathbf{R}$ associated with $\omega_1$ and $\omega_n$ \\
$\boldsymbol{0}$, $\boldsymbol{1}$ & Vectors of zeros and ones \\
\addlinespace[0.35em]

%=========================================================
% Quantum states
%=========================================================
\hline \multicolumn{2}{@{}l}{\textbf{Quantum States}}\\
\addlinespace[0.25em]
$\ket{\boldsymbol{z}(\boldsymbol{w})}$ & Trial solution state \\
$\ket{\boldsymbol{z}^\star}$ & Normalized solution of the encoded system \\
$\ket{\boldsymbol{b}}$ & Normalized right-hand side $\boldsymbol{b}_\text{q}/\norm{\boldsymbol{b}_\text{q}}$ \\
$\ket{0}^{\otimes q}$, $\ket{i}$ & All-zero state of the system register; $i$-th computational basis state \\
$\ket{\cdot}_\text{f}$ & State of the flag qubit \\
$\ket{\boldsymbol{\varphi}_0}$, $\ket{\boldsymbol{\varphi}_1}$, $\ket{\boldsymbol{\varphi}_2}$ & Intermediate states of the coherent VQLS circuit \\
$\ket{\boldsymbol{\zeta}_j}$ & Unnormalized system-register components of the unsuccessful branches \\
\addlinespace[0.35em]

%=========================================================
% Matrices
%=========================================================
\hline \multicolumn{2}{@{}l}{\textbf{Matrices}}\\
\addlinespace[0.25em]
$\mathbf{R}$ & Transfer matrix from the HSP to the BSP \\
$\mathbf{M}$ & Generic full-column-rank matrix \\
$\mathbf{D}$ & Surface Laplacian of the heart mesh \\
$\mathbf{I}_n$ & Identity matrix of size $n$ (similarly $\mathbf{I}_m$, $\mathbf{I}_s$, $\mathbf{I}_k$, $\mathbf{I}_N$) \\
$\mathbf{E}_{\mathcal{S}}$ & Selection matrix of the support \\
$\mathbf{B}$ & Orthonormal subspace basis \\
$\mathbf{G}$ & Gram matrix of the compressed system \\
$\mathbf{Q}$ & Orthonormal Lanczos basis of the Krylov subspace \\
$\mathbf{T}$ & Compressed system matrix in Krylov coordinates \\
$\mathbf{T}_{jj}$ & $j$-th diagonal block of $\mathbf{T}$ \\
$\mathbf{P}$ & Block-Jacobi preconditioner \\
$\mathbf{L}_j$, $\mathbf{L}$ & Cholesky factor of $\mathbf{T}_{jj}$; block-diagonal collection of the factors \\
$\mathbf{A}_0$ & Preconditioned system matrix \\
$\mathbf{U}_\text{L}$, $\mathbf{U}_\text{R}$ & Left and right singular vector matrices of $\mathbf{A}_0$ \\
$\boldsymbol{\Gamma}$ & Diagonal matrix of the singular values of $\mathbf{A}_0$ \\
$\mathbf{U}_{\text{L},\mathcal{J}}$ & Columns of $\mathbf{U}_\text{L}$ indexed by $\mathcal{J}$ \\
$\mathbf{F}_\text{tail}$ & Spectral tail correction \\
$\mathbf{A}_\text{q}$ & Encoded system matrix delivered to the VQLS \\
$\mathbf{A}$ & System matrix without pre-encoding (Ablation 1) \\
$\boldsymbol{\Phi}$ & Isometric decoder \\
\addlinespace[0.35em]

%=========================================================
% Quantum operators
%=========================================================
\hline \multicolumn{2}{@{}l}{\textbf{Quantum Gates and Operators}}\\
\addlinespace[0.25em]
$\mathbf{V}(\boldsymbol{w})$ & M\"{o}tt\"{o}nen state preparation \\
$\mathbf{U}_{\mathbf{A}}$ & FABLE block encoding of $\mathbf{A}_\text{q}$ \\
$\mathbf{O}_{\mathbf{A}}$ & FABLE oracle \\
$\mathbf{U}_{\boldsymbol{b}}$ & State preparation of $\ket{\boldsymbol{b}}$ \\
$\mathbf{U}_\text{circ}(\boldsymbol{w})$ & Complete coherent VQLS circuit \\
$\mathbf{R}_\text{Y}(\theta)$, $\mathbf{R}_\text{Z}(\theta)$ & Single-qubit rotations about the Y and Z axes \\
$\mathbf{H}$ & Hadamard gate \\
$\mathbf{SWAP}$ & Qubit-wise swap of the index and system registers \\
$\boldsymbol{\Pi}_\text{anc}$, $\boldsymbol{\Pi}_\text{glob}$ & Projectors onto the all-zero state of the ancilla qubits and of all qubits \\
$\mathbf{I}_\text{anc}$, $\mathbf{I}_\text{f}$ & Identity on the ancilla qubits and on the flag qubit \\
\end{longtable}

\end{document}